%% file: DetectionDefenses.tex
\documentclass[10pt,twocolumn,letterpaper]{article}

\usepackage[pagenumbers]{wacv} 

\usepackage{graphicx}
\usepackage{amsmath}
\usepackage{amssymb}
\usepackage{booktabs}

\usepackage{dsfont}
\usepackage{tikz}
\usepackage{algpseudocode}
\usepackage{algorithm}
\usepackage{multicol}
\usepackage{multirow}
\usepackage{pifont}
\usepackage{colortbl}
\usepackage{csquotes}

\usepackage{caption}
\usepackage{subcaption}

\usepackage[pagebackref,breaklinks,colorlinks]{hyperref}

\usepackage[capitalize]{cleveref}
\crefname{section}{Sec.}{Secs.}
\Crefname{section}{Section}{Sections}
\Crefname{table}{Table}{Tables}
\crefname{table}{Tab.}{Tabs.}
\crefformat{footnote}{#2\footnotemark[#1]#3}

\newcommand{\darkleftbox}[3]{%
\ifx&#3&%
\includegraphics[#1]{#2}%
\else%
\begin{tikzpicture}\node[inner sep=0pt] (p00) {\includegraphics[#1]{#2}};\node[rectangle, inner sep=1pt, above right, fill=black, fill opacity=0.6, text opacity=1.] at (p00.south west) {\footnotesize \textcolor{white}{#3}}; \end{tikzpicture}%
\fi%
}

\newcommand{\darkrightbox}[3]{%
\ifx&#3&%
\includegraphics[#1]{#2}%
\else%
\begin{tikzpicture}\node[inner sep=0pt] (p00) {\includegraphics[#1]{#2}};\node[rectangle, inner sep=1pt, below left, fill=black, fill opacity=0.6, text opacity=1.] at (p00.north east) {\footnotesize \textcolor{white}{#3}}; \end{tikzpicture}%
\fi%
}

\newcommand{\darkleftrightbox}[4]{%
\ifx&#3&%
\includegraphics[#1]{#2}%
\else%
\begin{tikzpicture}\node[inner sep=0pt] (p00) {\includegraphics[#1]{#2}};%
\node[rectangle, inner sep=1pt, above right, fill=black, fill opacity=0.6, text opacity=1.] at (p00.south west) {\footnotesize \textcolor{white}{#3}};%
\node[rectangle, inner sep=1pt, below left, fill=black, fill opacity=0.6, text opacity=1.] at (p00.north east) {\footnotesize \textcolor{white}{#4}};%
\end{tikzpicture}%
\fi%
}

\newcommand{\lightlefttopbox}[2]{%
\begin{tikzpicture}\node[inner sep=0pt] (p00) {#1};\node[rectangle, inner sep=1pt, below right, fill=white, fill opacity=0.6, text opacity=1.] at (p00.north west) {\footnotesize \textcolor{black}{#2}}; \end{tikzpicture}%
}

\newcommand{\leftrotbox}[3]{%
\ifx&#3&%
\includegraphics[#1]{#2}%
\else%
\begin{tikzpicture}\node[inner sep=0pt] (p00) {\includegraphics[#1]{#2}};\node[rectangle, inner sep=1pt, left, fill=white, fill opacity=0.6, text opacity=1.] at (p00.west) [xshift=-.1cm] {\rotatebox{90}{\footnotesize #3}}; \end{tikzpicture}%
\fi%
}

\newcommand{\leftrotfbox}[3]{%
\ifx&#3&%
\includegraphics[#1]{#2}%
\else%
\begin{tikzpicture}\node[inner sep=0pt] (p00) {\fbox{\includegraphics[#1]{#2}}};\node[rectangle, inner sep=1pt, left, fill=white, fill opacity=0.6, text opacity=1.] at (p00.west) [xshift=-.1cm] {\rotatebox{90}{\footnotesize \phantom{j}#3\phantom{j}}}; \end{tikzpicture}%
\fi%
}

\newcommand{\leftrotboxwithfbox}[3]{%
\ifx&#3&%
\fbox{\includegraphics[#1]{#2}}
\else%
\begin{tikzpicture}%
\node[inner sep=0pt] (p00) {\fbox{\includegraphics[#1]{#2}}};%
\node[rectangle, inner sep=1pt, above right, fill=white, fill opacity=0.6, text opacity=1.] at (p00.south west) {\footnotesize \textcolor{black}{#3}};
\end{tikzpicture}%
\fi%
}

\newcommand{\leftrotboxwithfboxupper}[4]{%
\ifx&#3&%
\fbox{\includegraphics[#1]{#2}}
\else%
\begin{tikzpicture}%
\node[inner sep=0pt] (p00) {\fbox{\includegraphics[#1]{#2}}};%
\node[rectangle, inner sep=1pt, above right, fill=white, fill opacity=0.6, text opacity=1.] at (p00.south west) {\footnotesize \textcolor{black}{#3}};
\node[rectangle, inner sep=1pt, below, fill=white, fill opacity=0.6, text opacity=1.] at (p00.north) {\footnotesize \textcolor{black}{#4}};
\end{tikzpicture}%
\fi%
}

\newcommand{\leftrotboxwithfboxupperone}[3]{%
\ifx&#3&%
\fbox{\includegraphics[#1]{#2}}
\else%
\begin{tikzpicture}%
\node[inner sep=0pt] (p00) {\fbox{\includegraphics[#1]{#2}}};%
\node[rectangle, inner sep=1pt, below, fill=white, fill opacity=0.6, text opacity=1.] at (p00.north) {\footnotesize \textcolor{black}{#3}};
\end{tikzpicture}%
\fi%
}

\newcommand{\darkuptwobox}[4]{%
\ifx&#3&%
\includegraphics[#1]{#2}%
\else%
\begin{tikzpicture}\node[inner sep=0pt] (p00) {\includegraphics[#1]{#2}};%
\node[rectangle, inner sep=1pt, below right, fill=black, fill opacity=0.6, text opacity=1.] at (p00.north west) {\footnotesize \textcolor{white}{#3}};%
\node[rectangle, inner sep=1pt, below left, fill=black, fill opacity=0.6, text opacity=1.] at (p00.north east) {\footnotesize \textcolor{white}{#4}};%
\end{tikzpicture}%
\fi%
}

\newcommand{\inpicbox}[6]{%
\begin{tikzpicture}
\node[inner sep=0pt, anchor=south west] (p00) {\includegraphics[#1]{#2}};
\draw[draw=red!70!black,line width=.5pt] (#3,#4) rectangle (#3+#5,#4+#6) ; 
\end{tikzpicture}%
}

\newcommand{\circlebox}[4]{%
\begin{tikzpicture}
\node[inner sep=0pt, anchor=south west] (p00) {#1};
\node[circle,draw=blue!70!green,line width=.5pt, minimum size =#2cm] (c) at (#3,#4){}; 
\end{tikzpicture}%
}

\newcommand{\smallcirclebox}[4]{
\begin{tikzpicture}
\node[inner sep=0pt, anchor=south west] (p00) {#1};
\node[circle,draw=blue!70!green,line width=.2pt, scale=0.5, minimum size =#2cm] (c) at (#3,#4){}; 
\end{tikzpicture}%
}

\newcommand{\R}{\mathds{R}}
\newcommand{\flow}{f}

\def\wacvPaperID{00196} 
\def\confName{WACV}
\def\confYear{2024}

\def\wacvAcceptance{IEEE/CVF Winter Conference on Applications of Computer Vision (WACV)}
\def\copyright{\linespread{1}IEEE 2023. Personal use of this material is permitted. Permission from IEEE must be obtained for all other uses, in any current or future media, including reprinting/republishing this material for advertising or promotional purposes, creating new collective works, for resale or redistribution to servers or lists, or reuse of any copyrighted component of this work in other works.}

\begin{document}


\title{Detection Defenses: An Empty Promise\\against Adversarial Patch Attacks on Optical Flow}

\author{\renewcommand{\thefootnote}{\arabic{footnote}}Erik Scheurer\renewcommand{\thefootnote}{\fnsymbol{footnote}}\footnotemark[1] \renewcommand{\thefootnote}{\arabic{footnote}}\footnotemark[1]  \hspace*{.75cm} Jenny Schmalfuss\renewcommand{\thefootnote}{\fnsymbol{footnote}}\footnotemark[1] \renewcommand{\thefootnote}{\arabic{footnote}}\footnotemark[2] \hspace*{.75cm} Alexander Lis \hspace*{.75cm} Andrés Bruhn\footnotemark[2]\\
Institute for Visualization and Interactive Systems, University of Stuttgart\\
{\renewcommand{\thefootnote}{\arabic{footnote}}\tt\small first.last@\{\footnotemark[1]\hspace{.5em}simtech,\footnotemark[2]\hspace{.5em}vis\}.uni-stuttgart.de}
}
\maketitle


\begin{abstract}
Adversarial patches undermine the reliability of optical flow predictions when placed in arbitrary scene locations.
Therefore, they pose a realistic threat to real-world motion detection and its downstream applications.
Potential remedies are defense strategies that detect and remove adversarial patches, but their influence on the underlying motion prediction has not been investigated.
In this paper, we thoroughly examine the currently available detect-and-remove defenses ILP and LGS for a wide selection of state-of-the-art optical flow methods, and illuminate their side effects on the quality and robustness of the final flow predictions.
In particular, we implement defense-aware attacks to investigate whether current defenses are able to withstand attacks that take the defense mechanism into account.
Our experiments yield two surprising results: Detect-and-remove defenses do not only lower the optical flow quality on benign scenes, in doing so, they also harm the robustness under patch attacks for all tested optical flow methods except FlowNetC.
As currently employed detect-and-remove defenses fail to deliver the promised adversarial robustness for optical flow, they evoke a false sense of security.
The code is available at \url{https://github.com/cv-stuttgart/DetectionDefenses}
\end{abstract}
\thispagestyle{empty}

\begin{figure}
   \setlength{\tabcolsep}{.2pt}
   \setlength{\fboxsep}{0pt}
   \setlength{\fboxrule}{.1pt}
   \centering
   \begin{tabular}{c@{\ }c@{\ }c}

      \multicolumn{3}{c}{\hspace*{.1281\linewidth}\small Vanilla patch attack}                                                \\
      \includegraphics[height=.1281\linewidth]{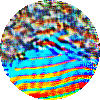}                &
      \darkleftbox{height=.1281\linewidth}{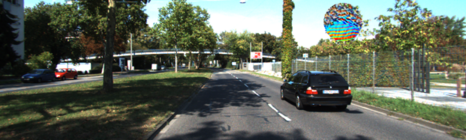}{Defense: None} &
      \fbox{\includegraphics[height=.1281\linewidth]{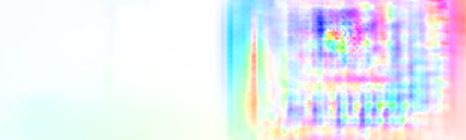}}      \\
      \\[-.5em]

      \includegraphics[height=.1281\linewidth]{graphics/intro/PStd.png}                &
      \darkleftbox{height=.1281\linewidth}{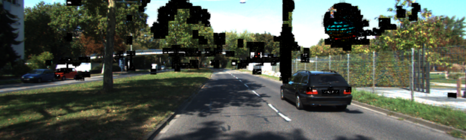}{Defense: LGS}         &
      \fbox{\includegraphics[height=.1281\linewidth]{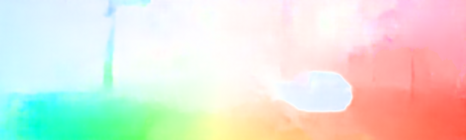}}       \\[-.1em]

      \includegraphics[height=.1281\linewidth]{graphics/intro/PStd.png}                &
      \darkleftbox{height=.1281\linewidth}{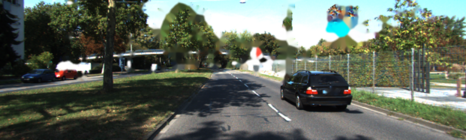}{Defense: ILP}         &
      \fbox{\includegraphics[height=.1281\linewidth]{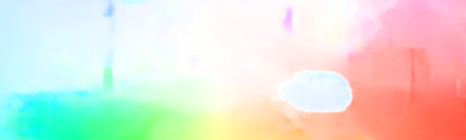}}       \\[.1em]

      \multicolumn{3}{c}{\hspace*{.1281\linewidth}\small Defense-aware patch attack}                                                \\
      \includegraphics[height=.1281\linewidth]{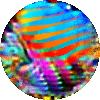}                &
      \darkleftbox{height=.1281\linewidth}{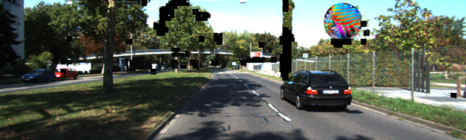}{Defense: LGS}         &
      \fbox{\includegraphics[height=.1281\linewidth]{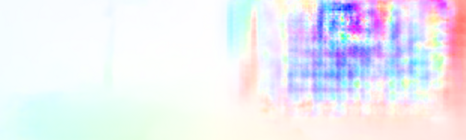}}   \\[-.1em]

      \includegraphics[height=.1281\linewidth]{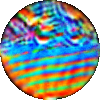}                &
      \darkleftbox{height=.1281\linewidth}{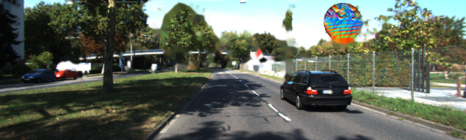}{Defense: ILP}         &
      \fbox{\includegraphics[height=.1281\linewidth]{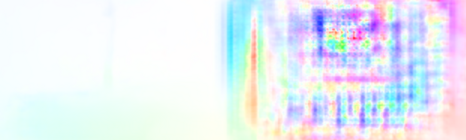}}   \\
   \end{tabular}
   \caption{Standard patch attack~\cite{Ranjan2019} (vanilla) and defense-aware attacks on FlowNetC's~\cite{FlowNetC2015} optical flow prediction. Left: Adversarial patch. Middle: Attacked image with applied defense (LGS or ILP, if any). Right: Optical flow. While both LGS and ILP defenses can defend against the vanilla patch attack~\cite{Ranjan2019,Anand2020} (top) neither defense withstands defense-aware patch attacks (bottom).}
   \label{fig:intro}
\end{figure}


\section{Introduction}

\renewcommand{\thefootnote}{\fnsymbol{footnote}}
\footnotetext[1]{Equal contribution.}
\renewcommand{\thefootnote}{\arabic{footnote}}
Adversarial attacks have an enormous potential to mislead optical flow methods into predicting the wrong apparent 2D motion from image sequences.
Among them, adversarial patches~\cite{Ranjan2019,Wortman2021HiddenPatchAttacks,Yamanaka2021SimultaneousAttackCnn} are the most safety-critical 
as they distort the optical flow predictions largely independent of their location and orientation, \cf \cref{fig:intro}, and~are printable for effective physical-world attacks~\cite{Ranjan2019,Yamanaka2020AdversarialPatchAttacks}.
On top of that, embedding the distorted optical flow into high-level recognition methods, \eg for flow-based action recognition~\cite{Carreira2017QuoVadisAction,ilic2022appearance}, often corrupts the downstream application~\cite{Inkawhich2018AdversarialAttacksOptical,Zhang2022AdversariallyRobustVideo}.

To protect methods against the negative effects of adversarial patches, a straightforward defense concept is to first detect the patch and then render it harmless, \eg by masking the former patch area~\cite{Naseer2019lgs,Hayes2018CVPRWorkshops}.
However, for classification, it was soon discovered that early defenses do not withstand attacks that take the defense mechanism into account~\cite{Chiang2020Certified,Athalye18}.
Such broken defenses are useless for practical applications because attackers aware of the method design (including potential defense mechanisms) can easily overcome them~\cite{Tramer2020,Athalye18,Carlini2017AdversarialExamplesAre}.
Among the broken defenses~\cite{Chiang2020Certified} for patch attacks on classification is Local Gradient Smoothing (LGS)~\cite{Naseer2019lgs}, which also has been considered to defend optical-flow based action recognition pipelines~\cite{Anand2020}.
Because LGS simply blackens the detected adversarial patch, Inpainting with Laplacian Prior (ILP)~\cite{Anand2020}  was proposed.
ILP inpaints the patch region using neighborhood information to improve the classification accuracy for patch-attacked action recognition pipelines.
However, the evaluation of LGS and ILP on the optical-flow component for action recognition in~\cite{Anand2020} has two major problems:
First, it is unclear whether LGS or ILP withstand defense-aware attacks in the context of optical flow -- given the results for LGS in classification~\cite{Chiang2020Certified}, this is unlikely.
And second, an evaluation of how these defenses affect the quality and robustness of optical flow methods is missing, which significantly impacts their practical applicability for all optical-flow-based problems.
This work addresses both aspects by providing the first comprehensive analysis of detection defenses against patch attacks proposed in the context of optical flow.

\medskip
\noindent
\textbf{Contributions.} We make four contributions.
(i)~We develop \emph{defense-aware patch attacks} on the ILP and LGS defense for optical flow estimation, by making the defenses differentiable (replacing gradient-free operations) and by avoiding patch detection by the defense (with tailored loss terms).
(ii)~Moreover, we \emph{investigate the effectiveness} of ILP and LGS on a large set of optical flow methods.
Surprisingly, the defenses not only lower the quality of benign (unattacked) predictions but also decrease the robustness for standard (vanilla) and defense-aware attacks -- leaving no advantage of defended methods over undefended ones.
(iii)~Then, we find these \emph{significant defense shortcomings to be caused} by the delocalized destruction of image information for benign scenes, 
which currently prevents viable detection defenses for optical flow estimation.
(iv)~Finally, we formulate evaluation advice for defenses on pixel-wise prediction tasks like optical flow, to help avoid common evaluation mistakes in future defense proposals.


\section{Related work}

\noindent
\textbf{Adversarial (patch) attacks on optical flow.}
Optical flow methods take a pair of input frames $I_1, I_2\in\R^{M\!\times\!N\!\times\! 3}$ to predict the 2D vectors that describe the apparent motion, or optical flow $\flow\in\R^{M\!\times\!N\!\times\! 2}$ from $I_1$ to $I_2$.
Adversarial attacks then modify the input frames to \emph{corrupt the optical flow} prediction.
The first adversarial attacks on optical flow go back to Ranjan \etal~\cite{Ranjan2019} who considered patches~\cite{firstPatch2017}.
Since then, patch attacks were extended to include transparencies~\cite{Wortman2021HiddenPatchAttacks}, simultaneously harm depth estimators~\cite{Yamanaka2021SimultaneousAttackCnn} or were used to attack flow-based action recognition~\cite{Inkawhich2018AdversarialAttacksOptical}.
Meanwhile, image-wide attacks on optical flow range from global \cite{Schrodi2022,Schmalfuss2022,Agnihotri2023CospgdUnifiedWhite} over semantically constrained attacks~\cite{Koren2021ConsistentSemanticAttacks} to adversarial weather~\cite{schmalfuss2023distracting,Schmalfuss2022AttackingMotionEstimation}.
Here, we investigate adversarial patches for being a threat in the physical world.

\medskip
\noindent
\textbf{Adversarial defenses and their evaluation.}
Adversarial defense mechanisms are designed to protect methods against the perturbing effects of adversarial attacks.
Typical defense strategies are adversarial training as a form of data augmentation \cite{szegedy2014intriguing,ifgsm2017,Gittings_2020_vaxanet,rao2020adversarial}, upstream strategies that filter perturbations from the inputs \cite{Naseer2019lgs,Hayes2018CVPRWorkshops,Liu2022SegmentCompleteDefending} and certified defenses that come with robustness guarantees \cite{brendel2018approximating,raghunathan2018certified,Chiang2020Certified,Levine2020,Xiang2021PatchguardProvablyRobust}.
However, many early defenses based on filtering operations were found to be ineffective if the attacker takes the defense mechanism into account~\cite{Tramer2020,Carlini2017AdversarialExamplesAre,Athalye18}.
Therefore, a defense's effectiveness has to be shown under defense-aware adversarial attacks to justify its merit.
In the process, one has to adequately (\ie effec\-tively) include the defense into the defense-aware attack: Prior work~\cite{Tramer2020,Athalye18} demonstrated  in the context of \emph{classification} that by neglecting this fact, many defenses appear unjustifiedly strong despite being evaluated with defense-aware attacks.
Hence in this work, following the evaluation guidelines from~\cite{Tramer2020,Athalye18}, we design the first defense-aware attacks on optical flow.

\medskip
\noindent
\textbf{Defenses against adversarial patch attacks.}
Very few defenses are specialized to optical flow~\cite{Anand2020,Zhang2022AdversariallyRobustVideo}. Hence, we first discuss general adversarial patch defenses related to \emph{classification}.
Certifiable defenses against patch attacks on classification are provably robust~\cite{Chiang2020Certified,Xiang2021PatchguardProvablyRobust,Levine2020}, but often lead to smaller robustness improvements.
Adversarial training~\cite{rao2020adversarial} and architectural modification~\cite{mu2021defending} have been also shown to improve the robustness against patch attacks.
A last class of patch defenses aims to detect the patch in order to remove it~\cite{Naseer2019lgs,Hayes2018CVPRWorkshops,Liu2022SegmentCompleteDefending,McCoyd2020MinorityReportsDefense}.
Among them, digital watermarking uses saliency maps~\cite{Hayes2018CVPRWorkshops}, while Local Gradient Smoothing (LGS) detects anomalies in the input gradients~\cite{Naseer2019lgs}.
Both use non-differentiable operations that hinder the backpropagation to train adversarial patches, but if these operations are replaced by differentiable ones~\cite{Athalye18}, both defenses are ineffective against defense-aware attacks~\cite{Chiang2020Certified}.

In the context of \emph{optical flow}, LGS has been applied to defend optical-flow-based action recognition~\cite{Anand2020}.
An optical-flow-specific improvement is Inpainting with Laplacian Prior (ILP)~\cite{Anand2020}, which yields visually pleasing defended images.
Also for action recognition, \cite{Zhang2022AdversariallyRobustVideo} proposed an optical-flow defense based on self-supervised counter-perturbations against noise-like perturbations.
Since we focus on defenses against \emph{patch attacks} for optical flow, this leaves LGS and ILP as potential methods.
However, for such defenses, no analysis with defense-aware attacks has been performed, and neither have optical flow methods been considered independent of action recognition.


\section{Defending optical flow with LGS and ILP}

We begin by providing technical details for the LGS~\cite{Naseer2019lgs} and ILP~\cite{Anand2020} defenses.
Both defenses detect the adversarial patch based on large image gradients and then replace these regions to remove the adversarial patch.

\medskip
\noindent
\textbf{Patch detection.}
To detect the patch, the image is split into overlapping blocks $B =K\!\times\!K$ of size $K$ and overlap $O$.
Then, a subset of blocks containing potential adversarial modifications is selected.
As adversarial patches often have large color changes (\cf \cref{fig:intro}), the gradient magnitude is accumulated for each block to identify blocks with the largest gradients.
The gradient magnitude computation differs for ILP and LGS:
While LGS considers first derivatives of the input image $I\in\R^{M\!\times\!N\!\times\! 3}$, ILP uses second derivatives, resulting in the gradient fields $G\in\R^{M\!\times\!N}$:
\begin{align}
      G_\text{LGS} & = ||\nabla I||,   \\
      G_\text{ILP} & = ||\Delta I||.
\end{align}
Normalizing gradients per image yields scale invariance:
\begin{equation}
\bar{G}_{i,j}  =  \frac{G_{i,j} - \min_{i,j\in M\times N}G_{i,j}}{\max_{i,j\in M\times N}G_{i,j} - \min_{i,j\in M\times N}G_{i,j}}.
\end{equation}
Based on $\bar{G}$, adversarially modified pixels are marked:
Per pixel $(i,j)$, we denote all enclosing blocks by $B_{(i,j)}$, and let these blocks vote whether the sum of their gradients exceeds a threshold $t\in [0,1]$ ($t$ is relative to the distribution of $\bar{G}$).
If at least one block has large gradients, the respective pixel is marked as adversarial in a binary mask $M\in \R^{M\!\times\!N}$:
\begin{equation}
\label{eq:thresholding}
      M_{i,j}    = \begin{cases}
                      1 & \text{if} \ \ \exists \ B\! \in\! B_{(i,j)}: \sum_{k,l \in B} \bar{G}_{k,l} > t,\\
                      0 & \text{else}.
                   \end{cases}
\end{equation}
\begin{figure}
   \centering
   \setlength{\tabcolsep}{0.3pt}
   \setlength{\fboxsep}{0pt}
   \setlength{\fboxrule}{.1pt}
   \begin{tabular}{cc}
      \darkleftrightbox{width=.48\linewidth}{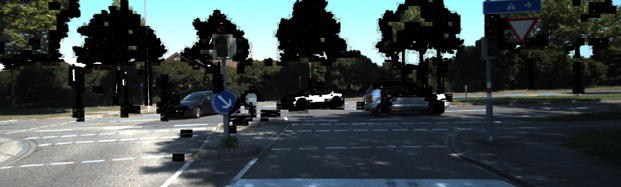}{Defense: LGS}{Attack: None} &
      \darkrightbox{width=.48\linewidth}{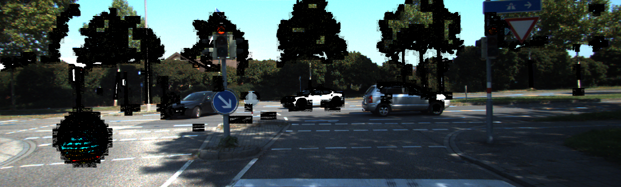}{Attack: Vanilla}\\[-.3em]
      \darkleftbox{width=.48\linewidth}{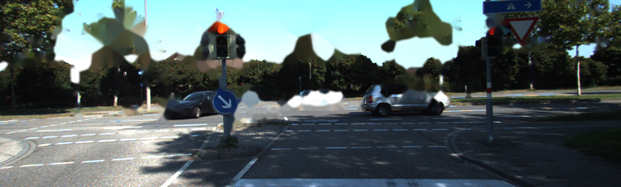}{Defense: ILP} &
      \includegraphics[width=.48\linewidth]{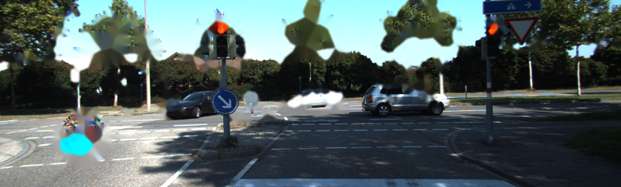}\\
   \end{tabular}
   \caption{ILP (top) and LGS (bottom) defenses on unattacked (left) and attacked (right) images of the KITTI 2015 dataset \cite{kitti2015}. Defenses degrade the visual quality, but LGS more than ILP.}
   \label{fig:defense-comparison}
\end{figure}
Through this procedure, a block with large gradients causes all contained pixels to be marked as adversarial.
To remove incorrectly marked (non-adversarial) pixels, ILP performs a reevaluation of candidates in $M$.
After scaling with $s_\text{ILP}$, their gradients must exceed a threshold $t_\text{ILP}$ to yield ILP's final mask $M_\text{ILP}$, with $\odot$ as pixel-wise multiplication:
\begin{equation}
\label{eq:ILPthresh}
    M_\text{ILP} =  M\odot \operatorname{tr}(s_\text{ILP} \cdot \bar{G}_\text{ILP} > t_\text{ILP}).
\end{equation}

\noindent
\textbf{Patch removal.}
Next, the defenses replace these potentially adversarial pixels from $M$ for LGS and from $M_\text{ILP}$ for ILP.
LGS reduces the gradients in the detected area, which results in the modified image
\begin{equation}
\label{eq:ILGSclip}
   I_\text{LGS} = (1- {\operatorfont clip}_{[0,1]}(b_\text{LGS} \cdot \bar{G}\odot M)) \odot I,
\end{equation}
where $b_\text{LGS}$ is a smoothing parameter.
If $b_\text{LGS}$ is large, this darkens the selected adversarial pixels.
ILP instead inpaints the selected pixels with Telea's algorithm \cite{Telea2004} with radius $r_\text{Telea}$ for more pleasing visual results.
This smoothes colors from the edges of the selected areas into their center:
\begin{equation}
\label{eq:ILPTelea}
   I_\text{ILP} = {\operatorfont Telea}(M_\text{ILP}, I, r_\text{Talea}),
\end{equation}

\cref{fig:intro} and \cref{fig:defense-comparison} show LGS- and ILP-defended images.
Our final hyperparameters for ILP and LGS are $K\!=\!16$, $O\!=\!8$, $t\!=\!0.15$, $t_\text{ILP}\!=\!0.5$, $s_\text{ILP}\!=\!15$, $b_\text{LGS}\!=\!15$ and $r_\text{Telea}\!=\!5$;
The supplement provides details on their selection.


\section{Defense-aware patch attacks for optical flow}
\label{sec:modifiedPatchAttack}

LGS and ILP defenses were only used to attack optical flow predictions for action recognition in a black-box way so far~\cite{Anand2020}, meaning adversarial patches were trained without knowledge about the defense.
According to best-practice for defense evaluation~\cite{Tramer2020,Carlini2017AdversarialExamplesAre,Athalye18}, the \emph{defended} model must be evaluated under \emph{defense-aware} attacks to show that it indeed offers protection.
In the following, we develop white box patch attacks on the ILP and LGS defenses for optical flow.
Our defense-aware attacks expand on Chiang \etal \cite{Chiang2020Certified} who successfully attacked LGS for classification, but neither considered ILP nor the optical flow problem.

\medskip
\noindent
\textbf{Gradient computations through the defense mechanism.}
The defensive properties of LGS and ILP are based on \emph{shattered gradients}, \ie the use of mathematical operations with nonexistent gradients that prevent adversarial optimization~\cite{Papernot2017Practical,Athalye18}.
To still optimize adversarial patches in a defense-aware manner, the Backward Pass Differential Approximation (BPDA)~\cite{Athalye18} replaces these operations with differentiable approximations during backpropagation.
The forward pass is executed normally.
Within LGS and ILP, the problematic operations are the block-wise filtering steps (LGS and ILP), thresholding (ILP) and clipping operations (LGS), and the inpainting step (ILP).
Below, we describe how they are approximated to enable backpropagation.

In the block-wise filtering step for LGS and ILP, \cf \cref{eq:thresholding}, the gradients do not exist for the conditional selection.
To bypass them with BPDA~\cite{Athalye18}, the filtering is replaced with the differentiable identity function, resulting in $\nabla M = 1$.
The thresholding in ILP's filtering has a similar problem, \cf \cref{eq:ILPthresh}, hence we also replace it with an identity function in the backward pass.
For the clipping in LGS's smoothing, \cf \cref{eq:ILGSclip}, the true gradient is one when the argument is in $[0,1]$ and otherwise undefined.
In practice, we find this operation responsible for most gradient shattering: Whenever the smoothing darkens values below zero, the clipping then sets them to zero, losing the gradient.
Therefore, in the backpropagation, we approximate gradients with the identity if the value to clip is in $[0,1]$ and with zero otherwise.
As the ILP inpainting is very time-consuming, \cf \cref{eq:ILPTelea}, we treat it as being gradient-free.
Similar to the clipping approximation, we bypass it with an identity operation for non-inpainted pixels and a zero-gradient for inpainted ones.
To overcome optimization problems for zero-gradients in the clipping- and inpainting approximations, we introduce additional loss terms to improve the patch in areas with no gradient information.

\medskip
\noindent
\textbf{Defense-aware loss functions.}
Optimizing defense-aware adversarial patches requires a loss function that encourages patches with a perturbing effect on the optical flow output.
As baseline loss that defines the overall goal for the patch attack, we use the Average Cosine Similarity (ACS) which was used to train adversarial patches on optical flow in~\cite{Ranjan2019}.
It encourages adversarial patches to invert the original optical flow prediction $\flow$ to yield the adversarial flow $\check{\flow}$:
\begin{equation}
   \label{eq:acs}
   \mathcal{L}^\text{ACS}(\flow, \check{\flow}) = \frac{1}{NM}\sum_{i,j\in M\!\times\!N} \frac{\langle\flow_{i,j} , \check{\flow}_{i,j}\rangle}{\|\flow_{i,j}\|_2 \|\check{\flow}_{i,j}\|_2}.
\end{equation}
Besides the ACS, another loss term is required to overcome the zero-gradients of BPDA.
While the differentiable LGS and ILP approximations allow optimizing adversarial patches, these patches may still be detected by the defenses and hence be stopped from perturbing the flow.
Therefore, we use loss terms to penalize large gradient magnitudes in the patches.
To optimize defense-aware patches $P$, we therefore penalize first-order gradients $\|\nabla P\|$ for LGS and second-order gradients $\|\Delta P\|$ for ILP-awareness:
\begin{align}
   \label{eq:lossLGS}
   \mathcal{L}^\text{LGS}(\flow,\check{\flow},P) & = \mathcal{L}^\text{ACS}(\flow,\check{\flow}) + \alpha\|\nabla P\|,    \\
   \label{eq:lossILP}
   \mathcal{L}^\text{ILP}(\flow,\check{\flow},P) & = \mathcal{L}^\text{ACS}(\flow,\check{\flow}) + \alpha\|\Delta P\|.
\end{align}
The parameter $\alpha$ balances the loss terms and is set to $\alpha=1\text{e}{-8}$.
With small gradient magnitudes, the patches are likely below the filtering threshold as it is relative to the remaining image gradients.
This way, they evade the defenses and affect the optical flow output as in \cref{fig:intro}.

In the ACS implementation, we exclude the patch area from the computation.
This measures to which extent the patch modifies the optical flow outside its direct area, \ie it assesses the de-localized impact per patch.
This is because one may take two points of view on the role of the patch:
In the first view, the patch is an image part, with a zero ground-truth flow at the patch area.
In the second view, the patch is an attack part, and defenses should mitigate its effect and restore the ground truth flow in its area.
To refrain from assuming a \enquote{correct} optical flow for the patch, we exclude the patch region from our loss. 

\medskip
\noindent
\textbf{Defense-aware patch optimization.}
\label{sec:optimization}
We test two different methods for optimization: The Iterative Fast Gradient Sign Method (I-FGSM) \cite{ifgsm2017} and Stochastic Gradient Descent (SGD).
To ensure a valid color range of the patch $P$ after optimization, we consider clipping the values to their valid range in $[0,1]$ after each update~\cite{Ranjan2019} and a change of variables (CoV) via tanh to optimize the values in $[-\infty,\infty]$ before transforming them back into the valid range~\cite{Schmalfuss2022,Carlini2017cov}.

\section{Metrics for defended quality and robustness}
\label{sec:metrics}

Including a defense to protect an existing method against attacks effectively creates a new method that consists of the original method plus defense D.
Hence, we have to evaluate the quality and robustness of this new method instead of the original defense-free approach's metrics~\cite{Tramer2020,Carlini2017AdversarialExamplesAre}.
\begin{figure*}
   \centering
   \begin{tabular}{ccccccc}
      \hspace{.4cm}\footnotesize FlowNetC \cite{FlowNetC2015} & \footnotesize FNCR \cite{Schrodi2022} & \footnotesize SpyNet \cite{SPyNet2017} & \footnotesize PWCNet \cite{PWCNet2018} & \footnotesize RAFT \cite{RAFT2020} & \footnotesize GMA \cite{GMA2021} & \footnotesize FlowFormer \cite{FlowFormer2022} \\
      \leftrotbox{width=.22\columnwidth}{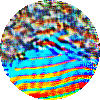}{Vanilla}
      & \includegraphics[width=.22\columnwidth]{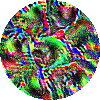}
      & \includegraphics[width=.22\columnwidth]{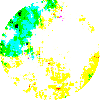}
      & \includegraphics[width=.22\columnwidth]{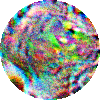}
      & \includegraphics[width=.22\columnwidth]{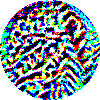}
      & \includegraphics[width=.22\columnwidth]{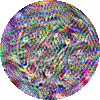}
      & \includegraphics[width=.22\columnwidth]{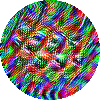}
      \\
      \leftrotbox{width=.22\columnwidth}{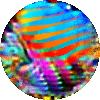}{LGS-aware}
      & \includegraphics[width=.22\columnwidth]{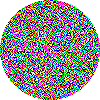}
      & \includegraphics[width=.22\columnwidth]{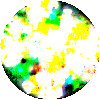}
      & \includegraphics[width=.22\columnwidth]{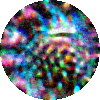}
      & \includegraphics[width=.22\columnwidth]{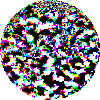}
      & \includegraphics[width=.22\columnwidth]{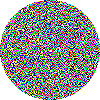}
      & \includegraphics[width=.22\columnwidth]{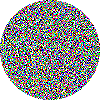}
      \\
      \leftrotbox{width=.22\columnwidth}{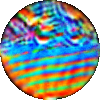}{ILP-aware}
      & \includegraphics[width=.22\columnwidth]{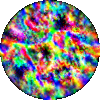}
      & \includegraphics[width=.22\columnwidth]{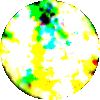}
      & \includegraphics[width=.22\columnwidth]{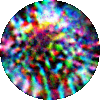}
      & \includegraphics[width=.22\columnwidth]{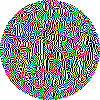}
      & \includegraphics[width=.22\columnwidth]{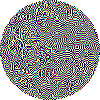}
      & \includegraphics[width=.22\columnwidth]{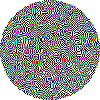}
      \\
   \end{tabular}
   \caption{Overview of adversarial patches with size 100 for vanilla, ILP- and LGS-aware patch attacks against all tested networks.}
   \label{fig:patch_overview}
\end{figure*}

\medskip
\noindent
\textbf{Quality.}
To evaluate the quality of defended optical flow methods, one typically measures the average endpoint error (EPE) between the ground-truth flow $\flow^*$ and the predicted flow $\flow$, where low errors indicate high quality:
\begin{equation}
\text{EPE}(\flow^*, \flow) = \frac{1}{MN} \sum_{i \in M\!\times\! N} \|\flow^*_{i} - \flow_i\|_2 \, .
\end{equation}

\noindent
\textbf{Robustness.}
To evaluate robustness, we use the methodology from~\cite{Schmalfuss2022} and measure the distance between the benign and the attacked flow prediction.
This quantifies how much an attack changes a method's output and is motivated by the Lipschitz continuity of functions.
Due to the previously discussed two views on the adversarial patch, we evaluate the EPE for all pixels \emph{outside} $P$, quantifying the patch's negative effect outside its immediate area.
We denote the benign flow prediction of a method defended with D as $\flow_\text{D}$ and its prediction under attack by $A$ as $\flow^\text{A}_\text{D}$.
Then, the robustness is
\begin{equation}
\text{EPE}_{P}(\flow_\text{D}, \flow_\text{D}^\text{A}) = \frac{1}{MN\!-\!P}\!\!\! \sum_{i \in M\!\times\! N \setminus P} \!\!\! \|(\flow_\text{D})_i - (\flow_\text{D}^\text{A})_{i}\|_2
\end{equation}
with low values for robust methods, as attacked predictions outside the patch should coincide with the unattacked ones.

\medskip
\noindent
\textbf{Quality and robustness for pipelines.}
When a method is defended with a defense D and attacked with an attack that is defense-aware towards D, we call the setup a \emph{full pipeline} with defense D.
Its quality is $\text{Q}_\text{D} = \text{EPE}(\flow^*, \flow_\text{D})$ and the resulting pipeline robustness is $R^\text{D}_\text{D} = \text{EPE}_{P}(\flow_\text{D}, \flow_\text{D}^\text{D})$.


\section{Experiments}
\label{sec:experiments}

We now assess how defenses against patch attacks impact the quality and robustness of optical flow methods.
We begin by evaluating the quality of defended methods and then separately assess their robustness against patch attacks.
Afterward, we jointly analyze both, quality and robustness, to find them being negatively impacted by defenses against patch attacks.
Finally, we explore the reasons for their poor performance.
All attacks and defenses are implemented with PyTorch~\cite{Pytorch}, 
and available at 
\url{https://github.com/cv-stuttgart/DetectionDefenses}.

\medskip
\noindent
\textbf{Optical flow methods and attack setups.}
As optical flow methods, we select FlowNetC (FNC)~\cite{FlowNetC2015}, SpyNet \cite{SPyNet2017} and PWCNet (PWC)~\cite{PWCNet2018} as milestone architectures in flow estimation, RAFT \cite{RAFT2020}, GMA \cite{GMA2021} and FlowFormer (FF) \cite{FlowFormer2022} as current state-of-the-art, and FlowNetCRobust (FNCR)~\cite{Schrodi2022} as it improves FlowNetC's patch robustness.

We generate effective adversarial patches by optimizing learning rates, box constraints and optimizer choice for each optical flow network.
As optimizers, we consider I-FGSM~\cite{ifgsm2017} and SGD, learning rates from 0.1-100 (optimizer-dependent), box constraints via change of variables (CoV) or clipping.
Following the protocol for adversarial patches from~\cite{Ranjan2019}, we optimize patches of size 100 using KITTI Raw \cite{kittiRaw} and evaluate on KITTI train \cite{kitti2015}.
Evaluations using Sintel~\cite{Butler2012NaturalisticOpenSource}, Driving~\cite{Mayer2016LargeDatasetTrain}, HD1K~\cite{Kondermann2016HciBenchmarkSuite} and Spring~\cite{Mehl2023SpringHighResolution} are shown in the supplement.
For defense-aware patches, we choose $\alpha=1\text{e}{-8}$ for the loss function in \cref{eq:lossLGS} and \cref{eq:lossILP}.
Per flow method, we generate vanilla adversarial patches (without defense awareness, as in~\cite{Ranjan2019}) and defense-aware patches for LGS~\cite{Naseer2019lgs} or ILP~\cite{Anand2020}.
We train 4 patches per parameter combination, and average the evaluated metrics.
Among the parameters, we select the strongest adversarial configuration for the worst robustness.

The full evaluation of the best parameters per flow method and defense strategy is in the supplement.
\cref{fig:patch_overview} visualizes the most effective patches.
Both defenses detect high gradient magnitudes, causing smooth defense-aware patches with small derivatives.
Interestingly, defense-aware patches for methods like RAFT, GMA or FlowFormer contain high-frequent noise.
This calls for detection rather than evasion by ILP or LGS, which we explore in \cref{sec:manualpatch}.

\subsection{Quality of defended optical flow methods}

\begin{table}
   \caption{Quality $\text{Q}_\text{D}=\text{EPE}(\flow^*, \flow_\text{D})$ for optical flow pipelines with defense D on the KITTI train dataset \cite{kitti2015}; Best quality is \textbf{bold}. All defenses lead to a worse quality on unattacked frames.}
   \label{tab:accur}
   \small
   \centering
   \begin{tabular}{l@{\ \ }l|r@{\ \ \ }r@{\ \ \ }r@{\ \ \ }r@{\ \ \ }r@{\ \ \ }r@{\ \ \ }r}
      \toprule
      \multicolumn{2}{l}{\rotatebox{0}{Defense}}
      & \rotatebox{60}{FNC} & \rotatebox{60}{FNCR} & \rotatebox{60}{SpyNet} & \rotatebox{60}{PWC} & \rotatebox{60}{RAFT} & \rotatebox{60}{GMA} & \rotatebox{60}{FF}  \\
      \midrule
      None   & $\text{Q}$             & \textbf{15.42}                                        & \textbf{11.10}                                   & \textbf{24.03}                                    & \textbf{13.26}                                    & \textbf{0.63}                                 & \textbf{0.61}                               & \textbf{0.62}                                             \\
      LGS    & $\text{Q}_\text{LGS}$  & 16.70                                        & 13.13                                   & 25.15                                    & 14.61                                    & 1.42                                 & 1.55                               & 1.42                                             \\
      ILP    & $\text{Q}_\text{ILP}$  & 16.46                                        & 12.77                                   & 24.74                                    & 14.52                                    & 1.36                                 & 1.39                               & 1.30                                             \\
      \bottomrule
   \end{tabular}
\end{table}
\begin{table*}
   \caption{Robustness scores for all combinations of defended methods and defense-aware attacks on optical flow methods on KITTI train~\cite{kitti2015}. For a given defense D and attack A, the robustness is defined as $R_\text{D}^\text{A} = \text{EPE}_{P}(\flow_\text{D}, \flow_\text{D}^\text{A})$. Per attack, the robustness values of the best defense are \textbf{bold}. Per defense, the robustness values for the attack it is most vulnerable to are \underline{underlined}.
   Full pipelines are highlighted in gray, and provide the corresponding robustness values to the quality scores from \cref{tab:accur}.
   }
   \label{tab:allresults}
   \small
   \centering
   \begin{tabular}{lll@{\quad}|@{\quad}rrrrrrr}
      \toprule
      Attack type                    & \multicolumn{2}{l}{Defense}                        & \multicolumn{1}{c}{\rotatebox{0}{FNC}} & \multicolumn{1}{c}{\rotatebox{0}{FNCR}}  & \multicolumn{1}{c}{\rotatebox{0}{SpyNet}} & \multicolumn{1}{c}{\rotatebox{0}{PWC}} & \multicolumn{1}{c}{\rotatebox{0}{RAFT}} & \multicolumn{1}{c}{\rotatebox{0}{GMA}} & \multicolumn{1}{c}{\rotatebox{0}{FF}} \\
      \midrule
      \multirow{3}{*}{Vanilla}       & \cellcolor{gray!20}None & \cellcolor{gray!20}$R^\text{Van}$                                        & \cellcolor{gray!20}\underline{73.74}                                        & \cellcolor{gray!20}\underline{\textbf{1.78}}                                    & \cellcolor{gray!20}\underline{\textbf{1.48}}                                     & \cellcolor{gray!20}\underline{\textbf{2.17}}                                     & \cellcolor{gray!20}\underline{\textbf{0.33}}                                 & \cellcolor{gray!20}\underline{\textbf{0.56}}                               & \cellcolor{gray!20}\underline{\textbf{0.57}}                                             \\
                                     & LGS  & $R^\text{Van}_\text{LGS}$                             & \textbf{3.75}                                         & {2.97}                                    & 3.97                                     & 3.34                                     & 1.45                                 & 1.31                               & 1.30                                             \\
                                     & ILP  & $R^\text{Van}_\text{ILP}$                             & 4.66                                         & 3.11                                    & 3.34                                     & 3.29                                     & 1.43                                 & 0.99                               & 1.41                                             \\
      \cmidrule(l{0em}r{0em}){2-10}
      \multirow{3}{*}{+LGS (LGS-aware)}     & None & $R^\text{LGS}$                                        & 50.46                                        & \textbf{0.46}                                    & \textbf{1.40}                                     & \textbf{2.10}                                     & \textbf{0.25}                                 & \textbf{0.27}                               & \textbf{0.44}                                             \\
                                     & \cellcolor{gray!20}LGS  & \cellcolor{gray!20}$R^\text{LGS}_\text{LGS}$                             & \cellcolor{gray!20}\underline{23.36}                                        & \cellcolor{gray!20}\underline{3.27}                                    & \cellcolor{gray!20}\underline{4.05}                                     & \cellcolor{gray!20}4.13                                     & \cellcolor{gray!20}1.46                                 & \cellcolor{gray!20}\underline{1.60}                               & \cellcolor{gray!20}1.67                                             \\
                                     & ILP  & $R^\text{LGS}_\text{ILP}$                             & \textbf{23.04}                                        & \underline{4.21}                                    & 3.32                                     & 3.35                                     & 1.48                                 & \underline{1.54}                               & 1.80                                             \\
      \cmidrule(l{0em}r{0em}){2-10}
      \multirow{3}{*}{+ILP (ILP-aware)}     & None & $R^\text{ILP}$                                        & 56.56                                        & \textbf{1.02}                                    & \textbf{1.45}                                     & \textbf{2.16}                                     & \textbf{0.20}                                 & \textbf{0.26}                               & \textbf{0.45}                                             \\
                                     & LGS  & $R^\text{ILP}_\text{LGS}$                             & \textbf{10.99}                                        & 3.68                                    & 3.03                                     & \underline{4.06}                                     & \underline{1.47}                                 & 1.57                               & \underline{1.68}                                             \\
                                     & \cellcolor{gray!20}ILP  & \cellcolor{gray!20}$R^\text{ILP}_\text{ILP}$                             & \cellcolor{gray!20}\underline{55.26}                                        & \cellcolor{gray!20}3.25                                    & \cellcolor{gray!20}\underline{3.36}                                     & \cellcolor{gray!20}\underline{4.25}                                     & \cellcolor{gray!20}\underline{1.49}                                 & \cellcolor{gray!20}1.51                               & \cellcolor{gray!20}\underline{1.81}                                             \\
      \bottomrule
   \end{tabular}
\end{table*}

To begin our investigation of defenses D, we assess the quality of defended and undefended optical flow methods on unattacked input frames.
\cref{tab:accur} lists the endpoint errors $\text{EPE}(\flow^*, \flow_\text{D})$ on KITTI train, where the ground truth flow is available.
Across all optical flow methods, we find the lowest errors when no defense is applied;
Both ILP and LGS lead to larger errors.
But for accurate methods, the errors rise more than for less accurate ones, \ie by 156\% for GMA and 4\% for SpyNet, using LGS vs.\ no defense.
On average, ILP increases the error less than LGS, \ie by 129\% instead of\ 156\% for GMA, compared to no defense.
ILP performs better due to its more sophisticated image restoration, which adds pixel-wise filtering with inpainting rather than smoothing, \cf \cref{fig:defense-comparison}.
In the figure, applying ILP and LGS to un\-attacked images visually degrades them, leading to worse predicted flows.
Overall, detect-and-remove defenses lower the accuracy on unattacked frames, as they strongly affect the image quality, which harms the flow quality.

\subsection{Robustness under defense-aware patch attacks}

To study the robustness of defended flow methods under defense-aware attacks, we measure $R^\text{A}_\text{D} = \text{EPE}_{P}(\flow_\text{D}, \flow_\text{D}^\text{A})$ for all combinations of defenses D (None, LGS and ILP) and attacks A (vanilla, LGS- and ILP-aware).
\cref{tab:allresults} gives the full results.
In the analysis process, we (i) evaluate whether defense-aware attacks bypass the defenses, and then (ii) identify the most effective defense for each attack.

\medskip
\noindent
\textbf{Most effective attack per defense.}
First, we evaluate if our defense-aware attacks evade the defenses.
In practice, this corresponds to choosing a defense to observe how the defended model fares against different attacks.
For a fixed defense D in \cref{tab:allresults} we \underline{underline} the worst robustness, \ie the most effective attack.
Hence we compare $R_\text{D}^\text{Van}$, $R_\text{D}^\text{LGS}$ and $R_\text{D}^\text{ILP}$ (\eg the 2nd line in each block for D=LGS).

Without defenses every method is most vulnerable towards the vanilla attack: $R^\text{Van} \geq R^\text{LGS}, R^\text{ILP}$.
This is plausible, as LGS- and ILP-aware attacks impose additional constraints on the patches, which impairs their effectiveness for an undefended model.
For defended models, the corresponding defense-aware attacks are often most effective, \eg $R_\text{LGS}^\text{LGS}$ is largest for the LGS-defended FlowNetC, FlowNetCRobust, SpyNet and GMA.
This confirms that our adaptive attacks are truly defense-aware, as they are most effective on the defended models, \ie $R_\text{D}^\text{D} \geq R_\text{A}^\text{D}$ for the majority of models.
Still, in some cases, an ILP-aware attack performs better on an LGS-defended model and vice versa.
This indicates transferable patches for LGS and ILP, as the differences are small in these cases, \eg LGS-defended RAFT scores $R_\text{LGS}^\text{LGS}= 1.46$ and $R_\text{ILP}^\text{LGS}= 1.47$.

Overall, our defense-aware attacks are most effective \wrt the respective defended models, which validates their design and implementation.
Likewise, the defense-aware attacks are less effective on other defenses, as inappropriate constraints hinder the patch's effectiveness.
Nonetheless, we find that LGS- and ILP-aware patches are transferable.

\medskip
\noindent
\textbf{Most effective defense per attack.}
Next, we analyze the most effective defense for a given attack;
or in other words, which defense withstands most attacks.
Per fixed attack~A in \cref{tab:allresults}, we \textbf{boldface} the best robustness per network, 
comparing $R^\text{A}$, $R_\text{LGS}^\text{A}$ and $R_\text{ILP}^\text{A}$ with differing defenses.

\begin{figure}
    \centering
    \setlength{\tabcolsep}{0.3pt}
    \setlength{\fboxsep}{0pt}
    \setlength{\fboxrule}{.1pt}
    \begin{tabular}{c c}
        \leftrotboxwithfboxupper{width=.495\linewidth}{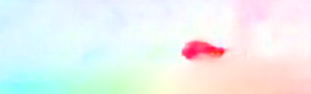}{FlowNetC}{$f$ unattacked}   & \circlebox{\leftrotboxwithfboxupperone{width=.495\linewidth}{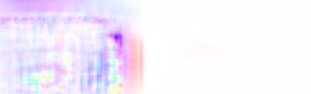}{$f^\text{Van}_\text{None}$ vanilla-attacked}}{.34}{.62}{.321}       \\[-.3em]
        \leftrotboxwithfbox{width=.495\linewidth}{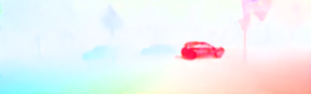}{PWCNet}       & \circlebox{\fbox{\includegraphics[width=.495\linewidth]{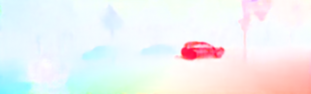}}}{.34}{.62}{.321}         \\[-.3em]
        \leftrotboxwithfbox{width=.495\linewidth}{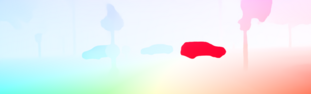}{RAFT}           & \circlebox{\fbox{\includegraphics[width=.495\linewidth]{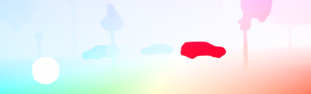}}}{.34}{.62}{.321}           \\[-.3em]
        \leftrotboxwithfbox{width=.495\linewidth}{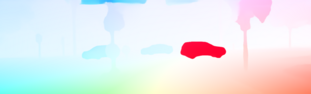}{FlowFormer}       & \circlebox{\fbox{\includegraphics[width=.495\linewidth]{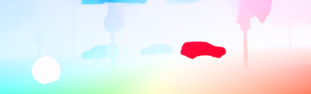}}}{.34}{.62}{.321}     
    \end{tabular}
    \caption{Optical flow estimations for selected methods on a KITTI frame that is unattacked (left) and attacked with the vanilla patch attack (right). Blue circles indicate the patch location. An overview of all optical flow methods is in the supplement.}
    \label{fig:vanilla_attack}
\end{figure}

Focusing first on the vanilla attack (\cref{tab:allresults} block~1), the undefended robustness $R^\text{Van}$ strongly differs.
FlowNetC is particularly vulnerable~\cite{Ranjan2019,Anand2020,Schrodi2022,Schmalfuss2022} due to a limited field of vision that was expanded in FlowNetCRobust~\cite{Schrodi2022} and improved its robustness by 97\%, making it comparable to SpyNet or PWCNet.
Most robust against vanilla patch attacks are the state-of-the-art methods RAFT, GMA and FlowFormer.
Their robustness appears linked to their quality, as they detect the static patches in \cref{fig:vanilla_attack}, correctly estimating the zero motion.
This retains correct flow predictions around the patch and results in low robustness scores.

For methods that are robust against vanilla attacks without defense, \ie all except FlowNetC, defending harms their robustness scores: $R^\text{A} < R^\text{A}_\text{LGS}, R^\text{A}_\text{ILP}$, independent of the attack A (vanilla, LGS- or ILP-aware).
This renders the defenses ineffective, as improving robustness against attacks is their sole purpose.
In contrast, defending FlowNetC improves its robustness against vanilla attacks from 73.74 to 3.75 with LGS and 4.66 with ILP.
For action recognition, a similar improvement was seen with FlowNetC~\cite{Anand2020}, but compared to other methods, FlowNetC is not robust even when defended and therefore should not be used.

All in all, the reported robustness enhancement through ILP and LGS in~\cite{Anand2020} can not be confirmed for our large test body of optical flow methods.
Instead, we find that LGS and ILP defenses harm the robustness of competitive optical flow methods for all tested patch attacks.

\subsection{Quality and robustness for defended methods}
\begin{figure}
   \centering
   \includegraphics[width=\linewidth]{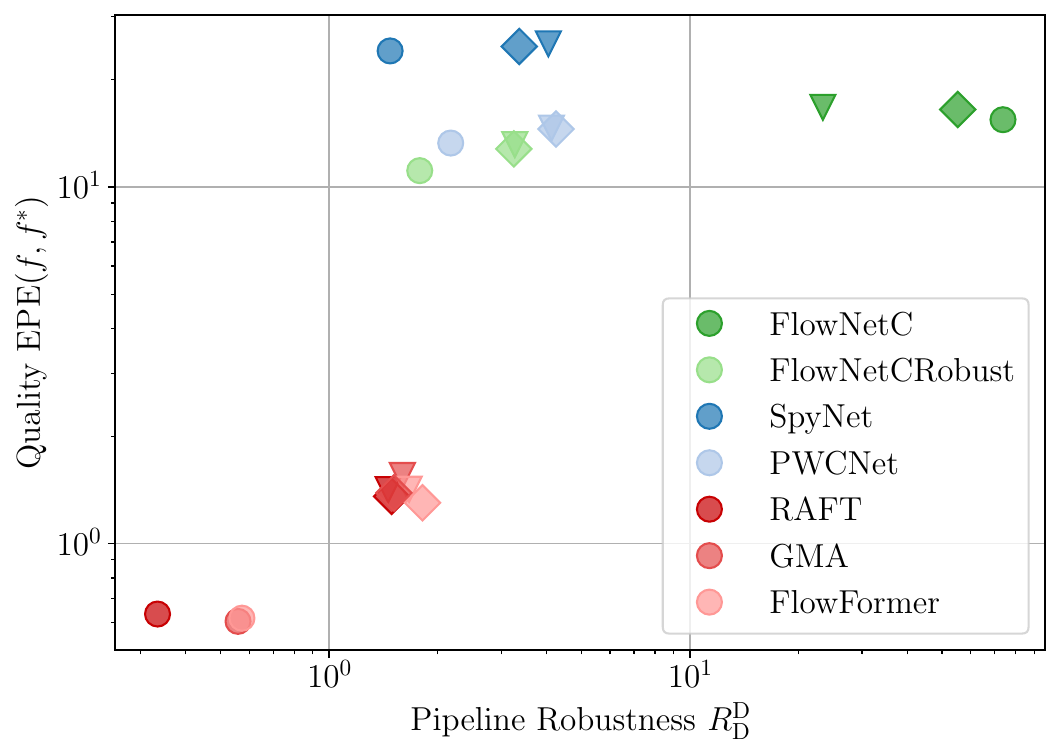}
   \caption{
   Quality vs. robustness of flow networks on KITTI train in a double logarithmic plot.
   An ideal method would be in the origin. Undefended networks are circles~$\bigcirc$, networks defended with LGS are triangles~$\triangledown$ and networks defended with ILP are diamonds~$\diamondsuit$.
   ILP and LGS deteriorate quality and robustness. 
   }
   \label{fig:acc-rob-nodef}
\end{figure}
After separately considering quality and robustness of defended methods under adversarial patch attacks, we now jointly analyze both aspects.
Perfect defenses decrease the vulnerability to adversarial attacks without negatively impacting the quality.
As models, we consider all tested flow methods with no defense, LGS or ILP.
Their quality is taken from \cref{tab:accur}.
Their corresponding pipeline robustness, \ie defended model's robustness under the respective defense-aware attacks, which is highlighted in gray in \cref{tab:allresults}.

For all optical flow methods, \cref{fig:acc-rob-nodef} visualizes the qual\-ity-robustness pairs per defense, \eg $\text{Q}_\text{LGS}$ with $R^\text{LGS}_\text{LGS}$.
An ideal method with low scores for quality and robustness would be positioned at the origin.
An improvement in robustness moves the defended point to the left, ideally without decreasing quality.
For all methods except FlowNetC, the undefended standard model~($\bigcirc$) is closest to the origin and therefore offers the best robustness \emph{and} the best quality, without any trade-off.
Using LGS~($\triangledown$) or ILP~($\diamondsuit$) defenses worsen both metrics to a similar extent.
The only outlier is FlowNetC, where both defenses improve the robustness while keeping the quality nearly constant, with larger improvements for LGS than for ILP.
Overall, our investigation shows that almost all optical flow methods are harmed by the detect-and-remove defenses ILP and LGS, as they worsen method quality and robustness alike.

\subsection{Flaws explained: Manual patch attack}
\label{sec:manualpatch}

From \cref{tab:allresults} we saw significant robustness reductions for high-quality methods like RAFT, GMA or FlowFormer when defended with ILP or LGS.
Yet, the reductions are caused by high-frequent defense-aware patches, \cf \cref{fig:patch_overview}, which seems to contradict the optimization for smoothness to evade detection by the ILP and LGS gradient filtering.

\begin{figure}
   \centering
   \setlength{\tabcolsep}{0.3pt}
   \setlength{\fboxsep}{0pt}
   \setlength{\fboxrule}{.1pt}
   \begin{tabular}{c c}
      \darkuptwobox{width=.495\linewidth}{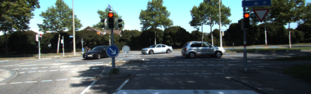}{Attack: None}{Defense: None}               &
      \lightlefttopbox{\fbox{\inpicbox{width=.495\linewidth}{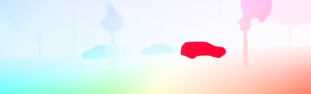}{2.3}{0.4}{1.8}{0.8}}}{$\flow$} \\[-.3em]
      \darkuptwobox{width=.495\linewidth}{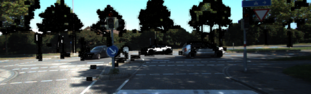}{Attack: None}{Defense: LGS}                &
      \lightlefttopbox{\fbox{\inpicbox{width=.495\linewidth}{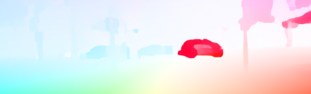}{2.3}{0.4}{1.8}{0.8}}}{$\flow_\text{LGS}$}                           \\[-.3em]
      \circlebox{\darkuptwobox{width=.495\linewidth}{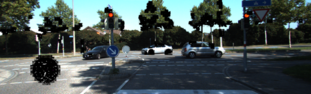}{Attack: LGS}{Defense: LGS}}{.34}{.618}{.321}            &
      \lightlefttopbox{\circlebox{\fbox{\inpicbox{width=.495\linewidth}{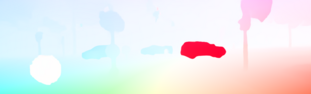}{2.3}{0.4}{1.8}{0.8}}}{.34}{.618}{.321}}{$\flow_\text{LGS}^\text{LGS}$}                       \\[-.3em]
      \circlebox{\darkuptwobox{width=.495\linewidth}{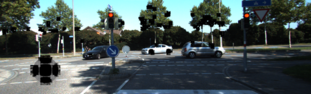}{Attack: Manual}{Defense: LGS}}{.34}{.618}{.321} &
      \lightlefttopbox{\circlebox{\fbox{\inpicbox{width=.495\linewidth}{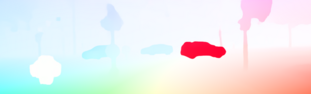}{2.3}{0.4}{1.8}{0.8}}}{.34}{.618}{.321}}{$\flow_\text{LGS}^\text{Man}$}                   \\[-.3em]
   \end{tabular}
   \caption{Effect of the LGS defenses on KITTI~\cite{kitti2015} frames (left) and the resulting optical flow prediction with RAFT~\cite{RAFT2020} (right). Black areas in the input frame are filtered by the LGS defense. Blue circles mark the area of the adversarial patch, the red boxes highlight an area with prominent differences in the flow predictions. Note that the robustness calculation omits the blue circle.}
   \label{fig:defense_effect}
\end{figure}

To understand this behavior, we compare the flows entering into the robustness calculation -- the unattacked flow $\flow_\text{D}$ of the defended method and the flow $\flow_\text{D}^\text{A}$ after applying a defense-aware attack to the defended model.
\cref{fig:defense_effect} shows RAFT's original prediction (unattacked, no defense, Row\ 1) together with flows for the LGS-defended version.
Comparing defended and undefended flows, \eg the car in the red box, the flow from the \emph{unattacked} LGS-defended RAFT ($\flow_\text{LGS}$, Row\ 2) is very erroneous compared to the \emph{attacked} LGS-defended flow ($\flow_\text{LGS}^\text{LGS}$, Row\ 3).
In other words, the gradient filtering of the defense destroys important visual information throughout the image, which yields low-quality optical flow predictions in the absence of adversarial attacks.
If the alterations in an unattacked image are scattered throughout its domain, a patch attack can maximize flow changes by aggregating alterations in a single location, \ie the patch itself.
Incidentally, this \emph{improves} the optical flow prediction in large areas (Rows 1, 3; Red box).

\begin{figure}
   \centering
   \setlength{\tabcolsep}{0.3pt}
   \setlength{\fboxsep}{0pt}
   \setlength{\fboxrule}{.1pt}
   \begin{tabular}{c@{\quad}c@{\quad}c@{\quad}}
      \small Vanilla                                                                                                                                   & \small LGS-aware & \small Manual \\
      \includegraphics[width=.3\linewidth]{graphics/evals/FlowFormer_none_evals_0000_ifgsm_0.01_1e-08_False_neg_flow_acs_none.png} &
      \includegraphics[width=.3\linewidth]{graphics/evals/FlowFormer_lgs_evals_0000_ifgsm_1.0_1e-08_False_neg_flow_acs_lgs.png}    &
      \includegraphics[width=.3\linewidth]{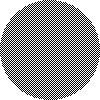}                                                                     \\
   \end{tabular}
   \caption{Visual comparison of patches obtained for vanilla, LGS-aware and manual patch attack on FlowFormer \cite{FlowFormer2022}. The manual patch imitates high derivatives in the LGS-aware adversarial patch.}
   \label{fig:manualpatch}
\end{figure}

\begin{table}
   \caption{Robustness $R_\text{D}^\text{Man} = \text{EPE}_{P}(\flow_\text{D}, \flow_\text{D}^\text{Man})$ against a manual patch attack (Man) of optical flow methods with different defenses D. The best robustness is \textbf{bold}.}
   \label{tab:manualpatch}
   \small
   \centering
\begin{tabular}{l@{\ \ }l|r@{\ \ \ }r@{\ \ \ }r@{\ \ \ }r@{\ \ \ }r@{\ \ \ }r@{\ \ \ }r}
      \toprule
      \multicolumn{2}{l}{Defense}                        & \rotatebox{60}{FNC} & \rotatebox{60}{FNCR}  & \rotatebox{60}{SpyNet} & \rotatebox{60}{PWC} & \rotatebox{60}{RAFT} & \rotatebox{60}{GMA} & \rotatebox{60}{FF} \\
      \midrule
      None & $R^\text{Man}$                                              & \textbf{1.19}                                         & \textbf{0.51}                                    & \textbf{1.15}                                     & \textbf{0.90}                                     & \textbf{0.19}                                 & \textbf{0.25}                               & \textbf{0.44}                                             \\
      LGS  & $R^\text{Man}_\text{LGS}$                                   & 3.69                                         & 3.26                                    & 3.86                                     & 3.40                                     & 1.45                                 & 1.56                               & 1.61                                             \\
      ILP  & $R^\text{Man}_\text{ILP}$                                   & 3.84                                         & 3.35                                    & 3.18                                     & 3.49                                     & 1.53                                 & 1.57                               & 1.86                                             \\
      \bottomrule
   \end{tabular}
   \vspace*{-.45\baselineskip}
\end{table}

Therefore, we hypothesize that the bad robustness scores of high-quality methods are driven by large distortions in un\-attacked frames caused by the defense.
To test this, we design a manual patch (see \cref{fig:manualpatch}) consisting of a checkerboard pattern to maximize first- and second derivatives.
With the manual patch, we then attack defended and undefended optical flow methods.
Their robustness in \cref{tab:manualpatch} confirms our hypothesis.
For undefended methods, the patch hardly affects the optical flow prediction.
For defended methods, however, the robustness significantly degrades, even though the patch impacts the defense, not the network.
Also, we obtain similar results to those of defense-aware attacks in \cref{tab:allresults}, where robustness degrades most significantly for accurate methods, \eg RAFT, GMA and FlowFormer.
However, because FlowNetC's robustness $\text{EPE}_{P}(f_\text{D}, f_\text{D}^\text{A})$ is driven by the attacked flow $\flow_\text{D}^\text{A}$ rather than the unattacked $\flow_\text{D}$, \cf \cref{fig:vanilla_attack}, defenses can succeed as every attacked-flow improvement directly serves the robustness.

In summary, the sub-par quality and robustness of defended high-quality methods are a direct consequence of the defense itself, which causes visual distortions in unattacked frames.
These distortions not only reduce the quality of benign frames but also are the empirically-confirmed cause for the low robustness scores of high-quality methods.


\section{Discussion}

We take a moment to condense the findings from analyzing defended optical flow methods into actionable evaluation advice, to encourage meaningful defense evaluations in the future.
While evaluation advice has been formulated before and should be adhered~\cite{Athalye18,madry2018pgd,Carlini2017AdversarialExamplesAre,Tramer2020}, we refresh some points, reinforce their importance and add discussions specific to pixel-wise prediction tasks like optical flow.

\medskip
\noindent
\textbf{Quality changes with defense.}
Defending a method creates a modified method and thus modifies its quality characteristics.
Therefore, the quality of the defended method $Q_\text{D}$ should be explicitly reported.
Particularly for pixel-wise prediction tasks, subtle changes in the inputs can cause significant output changes over large areas, making it indispensable to \emph{reevaluate the defended quality}.

\medskip
\noindent
\textbf{Use defense-aware attacks.}
Every defense proposal must be evaluated with a sufficiently \emph{strong adaptive attack}~\cite{Tramer2020,Athalye18,Carlini2017AdversarialExamplesAre} and report the pipeline robustness $R^\text{D}_\text{D}$.
Showing that it withstands adversarial samples for the original method is \emph{not enough}.
While many defense-circumvention strategies for classification~\cite{Tramer2020} may apply, individual tuning to pixel-wise prediction is needed for strong adaptive attacks. 

\medskip
\noindent
\textbf{Components matter.}
Defenses for specific components of complex methods should be evaluated on the \emph{specific part}, not only on the full method.
\emph{E.g.}\ when defending optical flow for action recognition~\cite{Anand2020,Zhang2022AdversariallyRobustVideo}, defended quality $Q_\text{D}$ and defense-aware robustness $R^\text{D}_\text{D}$ should be reported for the flow component.
Otherwise, the defense effectiveness and method sensitivity towards the component are entangled.


\section{Limitations}
This work solely focuses on detect-and-remove defenses for optical flow estimation.
Hence, it covers neither defenses for problems unrelated to optical flow, nor optical flow defenses against non-patch attacks (of which none were published so far).
Our findings that detection defenses do not protect against patch attacks could transfer to future defenses based on the gradient magnitude, as we found ILP-aware patches to transfer to LGS-defended methods and vice versa.
Nonetheless, defending optical flow may be possible with more specialized techniques.


\section{Conclusion}
\label{sec:conclusion}
We investigated detect-and-remove defenses against adversarial patch attacks on optical flow methods.
To this end, we designed defense-aware patches that avoid detection by LGS and ILP defenses, allowing us to break both defenses on a large variety of optical flow methods.
On top of that, we found that both defenses reduce the optical flow quality and even failed to increase the robustness against standard (\ie, not defense-aware) attacks.
We could attribute this discouraging performance to the severe image quality degradation resulting from pixel replacements in the defenses.
As image quality is crucial for pixel-wise motion estimation, this illustrates that defenses for classification methods, like LGS, do not automatically protect optical flow.
Consequently, flow pipelines' robustness and quality must be thoroughly investigated for every defense, to promote trust instead of making empty promises.

\medskip
\noindent
\textbf{Acknowledgments.}
We thank Filip Ilic for helpful discussions.
The International Max Planck Research School for Intelligent Systems supports JS.
Funded by the Deutsche Forschungsgemeinschaft (DFG, German Research Foundation) -- Project-ID 251654672 -- TRR 161 (B04).

{\small
\bibliographystyle{ieee_fullname}
\bibliography{egbib}
}

\cleardoublepage

\appendix
\setcounter{table}{0}
\renewcommand{\thetable}{A\arabic{table}}
\setcounter{figure}{0}
\renewcommand{\thefigure}{A\arabic{figure}}
\setcounter{footnote}{0}

\section{Supplementary Material}
\label{sec:supplementary_material}
In our evaluations, we consider the optical flow methods FlowNetC (FNC)~\cite{FlowNetC2015}, FlowNetCRobust (FNCR)~\cite{Schrodi2022}, PWCNet (PWC)~\cite{PWCNet2018}, SpyNet\footnote{Implementation from 
\url{github.com/sniklaus/pytorch-spynet}.}~\cite{SPyNet2017}, RAFT~\cite{RAFT2020}, GMA~\cite{GMA2021} and FlowFormer (FF)~\cite{FlowFormer2022}.

\subsection{Defense hyperparameter evaluation}

For the LGS~\cite{Naseer2019lgs} and ILP~\cite{Anand2020} defenses, we identify those hyperparameters that lead to the most effective defense against the vanilla patch attack~\cite{Ranjan2019} on FlowNetC~\cite{FlowNetC2015}.
The hyperparameters under consideration are the block size $K$, block overlap $O$ and the block filtering threshold $t$, which are used in ILP and LGS.
For LGS, we further consider the smoothing parameter $b_\text{LGS}$.
For ILP, we consider the scaling $s_\text{ILP}$, inpainting radius $r_\text{Talea}$ and the threshold $t_\text{ILP}$.
Out of those, we directly set $r_\text{Talea}=5$ and $t_\text{ILP}=0.5$, which are the values from Anand \etal \cite{Anand2020} that also produced good results for our experiments.
For the other parameters, we perform a parameter study that jointly evaluates the parameter pairs $K$ vs.\ $O$ (for LGS and ILP), $t$ vs.\ $b_\text{LGS}$ (for LGS) and $t$ vs.\ $s_\text{ILP}$ (for ILP).

\begin{figure}
\centering
\begin{tabular}{@{}c@{\ }c@{\ \ \ }c@{\ \ \ }c@{}}
\toprule
& & \small$\text{EPE}(\flow,\flow_\text{D}^\text{Van})$ ($\downarrow$) & \small$\text{EPE}(\flow,\flow_\text{D})$ ($\downarrow$)
\\
\midrule
\\[-.8em]
\rotatebox{90}{\hspace*{1.cm}\footnotesize LGS}
& \rotatebox{90}{\hspace*{.8cm}\footnotesize $K$ vs.\ $O$}
& \includegraphics[width=0.44\linewidth]{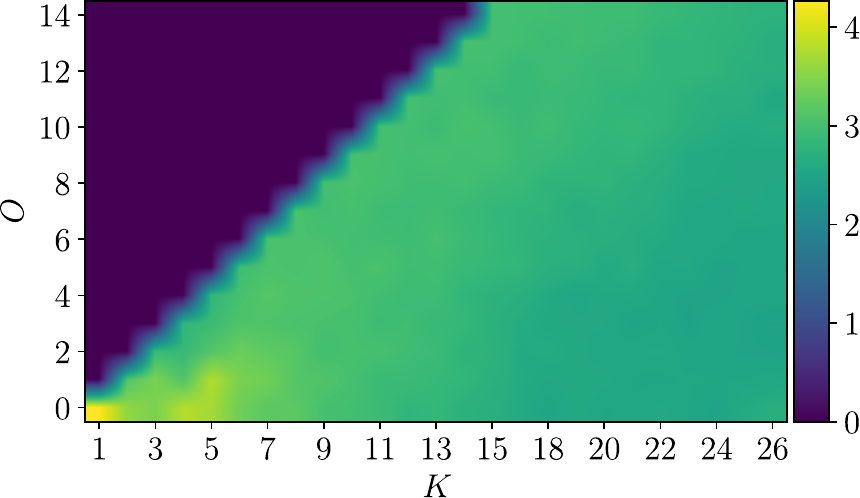}
& \includegraphics[width=0.44\linewidth]{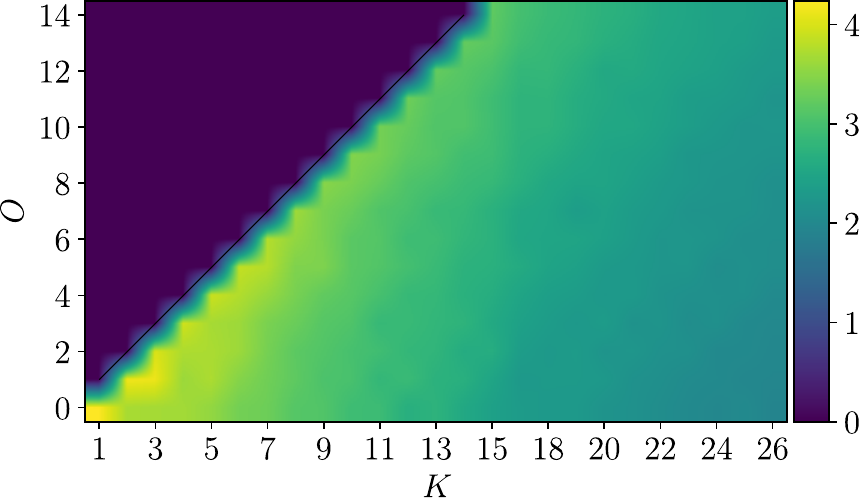}
\\
\rotatebox{90}{\hspace*{1.5cm}\footnotesize LGS}
& \rotatebox{90}{\hspace*{1.2cm}\footnotesize $b_\text{LGS}$ vs.\ $t$}
& \includegraphics[width=0.44\linewidth]{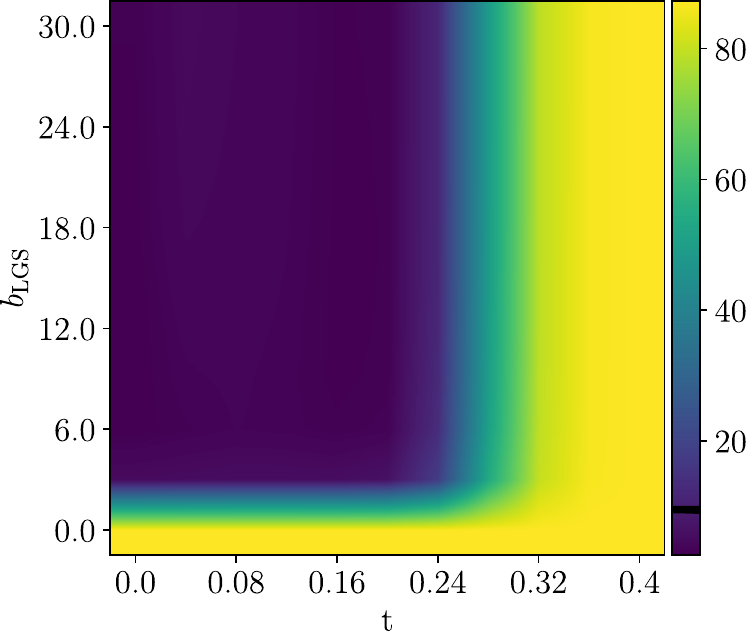}
& \includegraphics[width=0.44\linewidth]{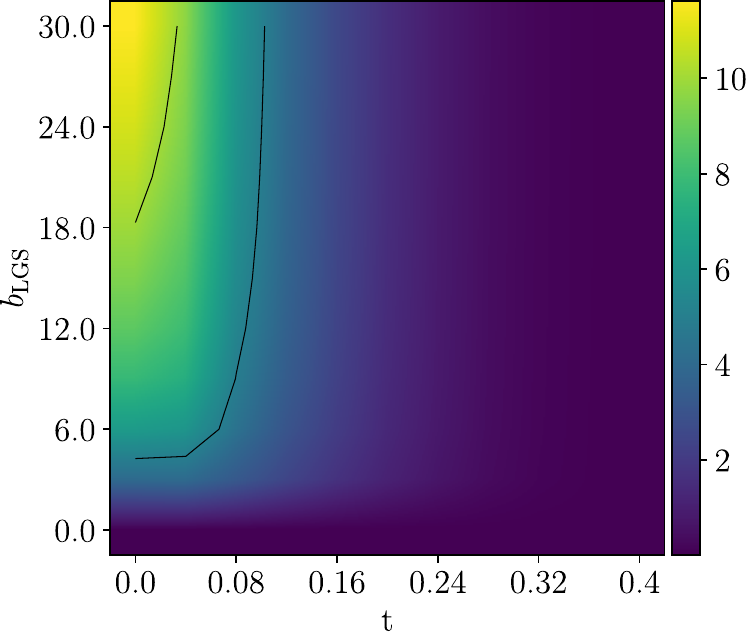}
\\
\midrule
\\[-.8em]
\rotatebox{90}{\hspace*{1.cm}\footnotesize ILP}
& \rotatebox{90}{\hspace*{.8cm}\footnotesize $K$ vs.\ $O$}
& \includegraphics[width=0.44\linewidth]{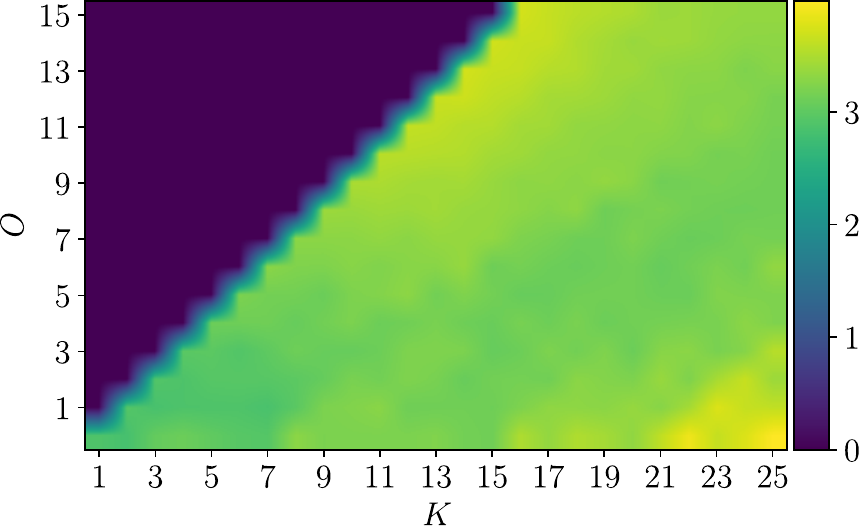}
& \includegraphics[width=0.44\linewidth]{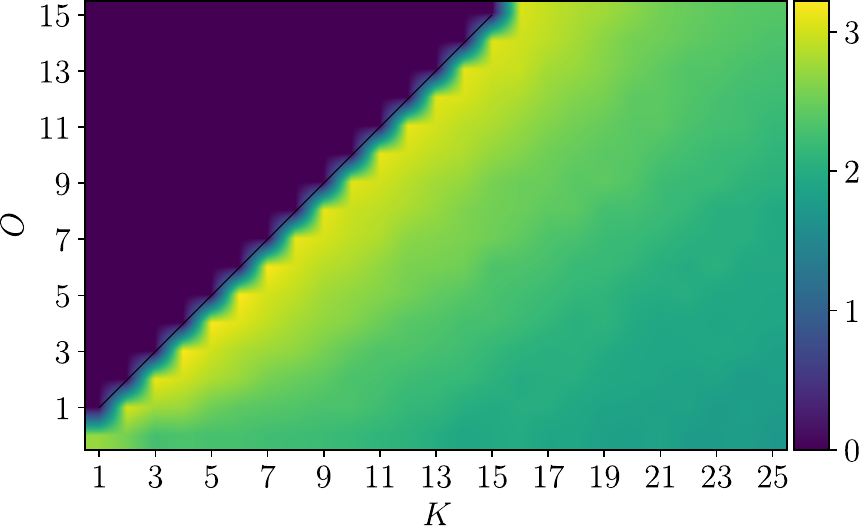}
\\
\rotatebox{90}{\hspace*{1.5cm}\footnotesize ILP}
& \rotatebox{90}{\hspace*{1.2cm}\footnotesize $s_\text{ILP}$ vs.\ $t$}
& \includegraphics[width=0.44\linewidth]{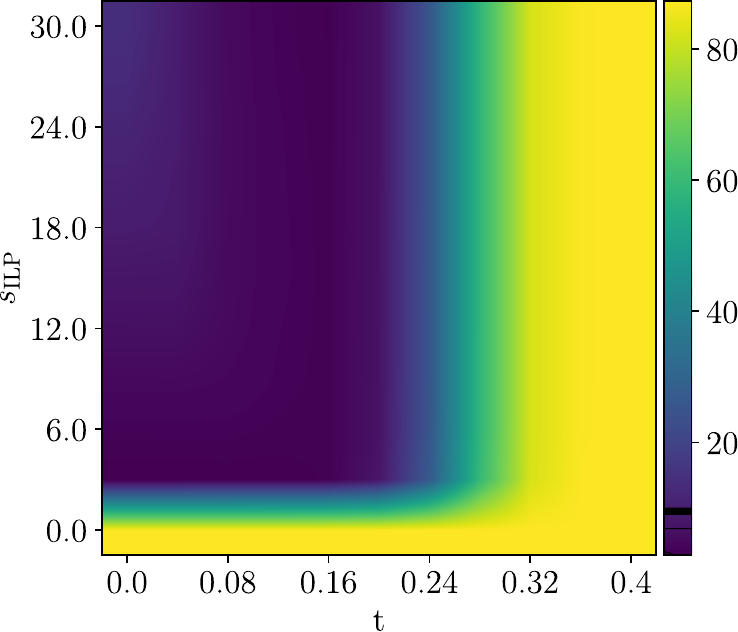}
& \includegraphics[width=0.44\linewidth]{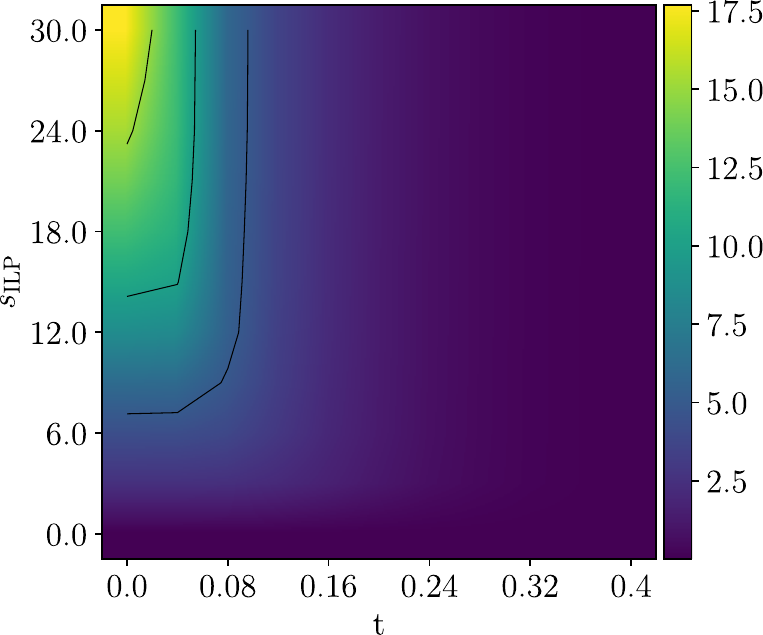}
\\
\bottomrule
\end{tabular}
   \caption{LGS and ILP hyperparameter study based on FlowNetC. Good parameters should balance the robustness against the vanilla attack $\text{EPE}(\flow,\flow_\text{D}^\text{Van})$ (dark color = good robustness) and small flow perturbations through the defense on unattacked frames $\text{EPE}(\flow,\flow_\text{D})$ (dark color = small perturbation). We select $K=16$, $O=8$, $t=0.15$ , $b_\text{LGS}=15$ and $s_\text{ILP}=15$ as best parameters}
   \label{fig:defenseParams}
\end{figure}
Per parameter combination, we evaluate the robustness of the defended FlowNetC against the vanilla attack via $\text{EPE}(\flow,\flow_\text{D}^\text{Van})$ (\cref{fig:defenseParams} left, small values indicate good robustness) and also quantify how much the defense changes the flow prediction for unattacked frames via $\text{EPE}(\flow,\flow_\text{D})$ (\cref{fig:defenseParams} right, small values indicate that defense does not change the flow prediction on benign samples).
\cref{fig:defenseParams} shows the plots for both metrics and all parameter pairs on LGS and ILP.
To select the parameters, for each parameter pair we pick values that lead to small values in both metrics (dark colors in plots for $\text{EPE}(\flow,\flow_\text{D}^\text{Van})$ and $\text{EPE}(\flow,\flow_\text{D})$), because then the defense protects against vanilla attacks but at the same time does not change the flow prediction on unattacked samples.
Thus, we select the parameters $K=16$, $O=8$, $s_\text{ILP}=15$, $b_\text{LGS}=15$ and $t=0.15$, which offer the best trade-off between the two metrics.
Note that for the $K$ vs.\ $O$ plots (\cref{fig:defenseParams}, Row 1 and 3), the dark area with low values in the upper left corner is unfeasible, because the block overlaps $O$ can not be larger than the blocksize $K$.
Overall, our optimized parameters differ slightly from the literature values:
For LGS, the original publication~\cite{Naseer2019lgs} used  $K=15$, $O=5$ and $b_\text{LGS}=2.3$ (for classification), while the original values for ILP from~\cite{Anand2020} are $t=0.25$, and $s_\text{ILP}=10$ (for optical-flow-based action recognition).

\subsection{Defense-aware attack setup and parameters}
\label{sec:patch_training_details}

Next, we evaluate the best combination of optimizers, learning rates (LR), and box constraints to optimize defense-aware patches.
As optimizers, we consider I-FGSM~\cite{ifgsm2017} and SGD, as learning rates 1, 0.1, 0.01 for I-FGSM and 10, 100 for SGD, and as box constraints either clipping or a change of variables (CoV).
Due to the algorithmic differences between I-FGSM and SGD, the considered learning rates have distinct magnitudes to achieve comparable results.
For each defense (none, LGS and ILP) we evaluate the pipeline robustness of the defended method under four separately trained defense-aware patches (using four fixed random seeds).
We report the averaged robustness values for our defense-aware patches in \cref{tab:patches-overview_None_ranking} (no defense), \cref{tab:patches-overview_LGS_ranking} (LGS defense) and \cref{tab:patches-overview_ILP_ranking} (ILP defense).
\input{graphics/patches-overview/patches-overview_None_ranking.tex}
\input{graphics/patches-overview/patches-overview_LGS_ranking.tex}
\input{graphics/patches-overview/patches-overview_ILP_ranking.tex}

Each defense-aware patch is trained for 2500 steps.
The patches are randomly placed on the image, randomly rotated in a range of $[-10,10]$ degrees, and randomly scaled in a range of $[0.95,1.05]$.
Batch size is chosen as 1 as the effect of batch size on the patch training is negligible.
Due to its size and the resulting computational cost to evaluate FlowFormer~\cite{FlowFormer2022}, we only train its patches for 1000 iterations and omit the change of variables to reduce the number of test runs, as it performed similarly to RAFT and GMA.
We found patches to be sufficiently converged after the reduced number of iterations.
Please note that across all methods, the choice of box constraint did not significantly influence the effectiveness of the adversarial patches.
The patch optimization for GMA diverged with SGD and learning rate 100 for LGS- and ILP-aware patches.

Based on this extensive parameter evaluation, we select the best optimization parameters for all combinations of optical flow network and defense-aware attack in \cref{tab:patch_params}, which are boldfaced in the detailed evaluations in \cref{tab:patches-overview_None_ranking}, \cref{tab:patches-overview_LGS_ranking} and \cref{tab:patches-overview_ILP_ranking}.
These parameters were used to produce the defense-aware patches for the experimental evaluation in the Main paper.
\begin{table}
   \caption{Optimal parameter setups for defense-aware patch attacks on all optical flow methods with defenses. LR is the learning rate, and Box indicates whether a change of variables or clipping is used during optimization. The settings are a summary of the best results from  \cref{tab:patches-overview_None_ranking}, \cref{tab:patches-overview_LGS_ranking} and \cref{tab:patches-overview_ILP_ranking}.}
   \label{tab:patch_params}
\small
\centering
\resizebox{\linewidth}{!}{
   \begin{tabular}{l l l| c r c}
      \toprule
      Attacked model                       & Defense & Attack & Optimizer & LR   & Constraint       \\
      \midrule
      \multirow{3}{*}{FlowNetC}       & None & Vanilla  & SGD   & 100.00  & CoV \\
                                      & LGS  & +LGS       & IFGSM & 0.01 & CoV \\
                                      & ILP  & +ILP       & IFGSM & 0.01 & CoV \\
      \midrule
      \multirow{3}{*}{FNCR}           & None & Vanilla  & IFGSM & 0.01 & Clip \\
                                      & LGS  & +LGS       & IFGSM & 1.00  & Clip \\
                                      & ILP  & +ILP       & IFGSM & 0.10  & CoV \\
      \midrule
      \multirow{3}{*}{SpyNet}         & None & Vanilla  & IFGSM & 0.10  & CoV \\
                                      & LGS  & +LGS       & IFGSM & 0.10  & CoV \\
                                      & ILP  & +ILP       & IFGSM & 0.10  & CoV \\
      \midrule
      \multirow{3}{*}{PWCNet}         & None & Vanilla  & IFGSM & 0.01 & CoV \\
                                      & LGS  & +LGS       & IFGSM & 0.01 & Clip \\
                                      & ILP  & +ILP       & IFGSM & 0.01 & Clip \\
      \midrule
      \multirow{3}{*}{RAFT}           & None & Vanilla & IFGSM & 1.00  & CoV \\
                                      & LGS  & +LGS      & SGD   & 100.00  & CoV \\
                                      & ILP  & +ILP      & IFGSM & 1.00  & Clip \\
      \midrule
      \multirow{3}{*}{GMA}            & None & Vanilla  & IFGSM & 0.01 & CoV \\
                                      & LGS  & +LGS       & IFGSM & 1.00  & Clip \\
                                      & ILP  & +ILP       & IFGSM & 1.00  & Clip \\
      \midrule
      \multirow{3}{*}{FlowFormer}     & None & Vanilla  & IFGSM & 0.01 & Clip \\
                                      & LGS  & +LGS       & IFGSM & 1.00  & Clip \\
                                      & ILP  & +ILP       & IFGSM & 1.00  & Clip \\
      \bottomrule
   \end{tabular}
   }
\end{table}

Additionally, we show the best (out of four) defense-aware patches for no defense, LGS-defense and ILP-defense in \cref{tab:patches-overview_None}, \cref{tab:patches-overview_LGS} and \cref{tab:patches-overview_ILP}, respectively.
The best patch is selected based on the worst robustness score of the defended method after training.
\input{graphics/patches-overview/patches-overview_None.tex}
\input{graphics/patches-overview/patches-overview_LGS.tex}
\input{graphics/patches-overview/patches-overview_ILP.tex}

\subsection{Additional flow visualizations for vanilla attack}

\input{graphics/vanilla_effects/effects.tex}
Here, we complement the limited selection of methods whose optical flow was visualized for unattacked and (vanilla) attacked frames in Main \cref{fig:vanilla_attack}.
Unattacked and vanilla-attacked flow visualizations on \emph{all} tested optical flow methods for the previous KITTI scene are in \cref{supp:fig:vanilla-rob1} and for an additional KITTI sample in \cref{supp:fig:vanilla-rob2}.
For a lean representation, only a single frame of the attacked image pair is shown on the right.

In both figures, RAFT~\cite{RAFT2020}, GMA~\cite{GMA2021} and FlowFormer~\cite{FlowFormer2022} are able to recognize the patch as a static object in the scene and therefore predict its output flow as zero.
The less accurate methods SpyNet~\cite{SPyNet2017}, PWCNet~\cite{PWCNet2018} and FlowNetCRobust~\cite{Schrodi2022} also recognize the zero flow, but their flow predictions are overall less precise and the patch bleeds into the surrounding area.
The outlier is FlowNetC~\cite{FlowNetC2015}, where the entire flow prediction is deteriorated by the patch.
In both visualizations, almost all optical flow methods are hardly affected by the patch, as they correctly recognize it as an object and accurately predict its zero motion.

\begin{table*}
\centering
   \caption{Quality $Q_\text{D}^\text{A}=\text{EPE}(\flow^*,\flow_\text{D}^\text{A})$, \ie the distance between ground truth flow and optical flow predictions that are defended with D and attacked with A. The upper block shows the quality scores from Main \cref{tab:accur} (for comparison), and the lower blocks contain the quality scores for full pipelines and manual patch attacks on networks with varying defenses. Per method, we mark the best defended quality for unattacked networks and full pipelines \textbf{bold} (includes the first two blocks, up to double line), and \underline{underline} the best quality if the manual patch is also included -- the undefended baselines that are marked in gray are excluded from both rankings.}
   \label{tab:defendedDistanceToGroundTruth}
   \begin{tabular}{lll|ccccccc}
      \toprule
u      Attack type                & \multicolumn{2}{l}{Defense} & FNC                            & FNCR                       & PWC                    & SpyNet                    & RAFT                      & GMA                       & FF                                           \\
      \midrule
      \multirow{3}{*}{No Attack} & \cellcolor{gray!20}None     & \cellcolor{gray!20}$Q$  & \cellcolor{gray!20}15.42   & \cellcolor{gray!20}11.10  & \cellcolor{gray!20}13.26  & \cellcolor{gray!20}24.03  & \cellcolor{gray!20}0.63   & \cellcolor{gray!20}0.61  & \cellcolor{gray!20}0.62   \\[3pt]
                                 & LGS                         & $Q_\text{LGS}$                      & 16.70                      & 13.13                     & 14.61                     & 25.15                     & 1.42                      & 1.55                     & 1.42                      \\[3pt]
                                 & ILP                         & $Q_\text{ILP}$                      & \textbf{16.46} & 12.77                     & \textbf{14.52}            & \textbf{24.74}            & 1.36                      & 1.39                     & 1.30                      \\[3pt]
      \cmidrule(l{0em}r{0em}){1-10}
      Vanilla                    & \cellcolor{gray!20}None                        & \cellcolor{gray!20}$Q^\text{Van}$                      & \cellcolor{gray!20}84.48                      & \cellcolor{gray!20}12.64                     & \cellcolor{gray!20}15.27                     & \cellcolor{gray!20}25.11                     & \cellcolor{gray!20}0.80                      & \cellcolor{gray!20}0.91                     & \cellcolor{gray!20}0.78                      \\
      +LGS (LGS-aware)           & LGS                         & $Q^\text{LGS}_\text{LGS}$           & 34.41                      & \textbf{11.68}            & 15.43                     & 25.28                     & 0.94                      & 0.90                     & 0.83                      \\
      +ILP (ILP-aware)           & ILP                         & $Q^\text{ILP}_\text{ILP}$           & 65.02                      & 12.31                     & 15.65                     & 25.29                     & \underline{\textbf{0.68}} & \textbf{0.70}            & \underline{\textbf{0.68}} \\
      \cmidrule(l{0em}r{0em}){1-10}                                                                                                                                                                                                                                                                      \\[-15pt]
      \cmidrule(l{0em}r{0em}){1-10}
      \multirow{3}{*}{Manual}    & None \cellcolor{gray!20}    & $Q^\text{Man}$  \cellcolor{gray!20} & \cellcolor{gray!20} 16.21  & \cellcolor{gray!20} 11.38 & \cellcolor{gray!20} 13.87 & \cellcolor{gray!20} 24.79 & \cellcolor{gray!20} 0.71  & \cellcolor{gray!20} 0.70 & \cellcolor{gray!20} 0.70  \\
                                 & LGS                         & $Q^\text{Man}_\text{LGS}$           & 16.66                      & 11.69                     & 14.24                     & 24.87                     & 0.92                      & 0.91                     & 0.88                      \\
                                 & ILP                         & $Q^\text{Man}_\text{ILP}$           & \underline{16.16}                      & \underline{11.52}         & \underline{13.97}         & \underline{24.69}         & 0.70                      & \underline{0.68}         & 0.72                      \\

      \bottomrule
   \end{tabular}
\end{table*}

\subsection{Manual patch attack: Defended quality}
\label{sec:defended_distance_to_ground_truth}

In the manual patch analysis in \cref{sec:manualpatch}, the Main paper visually argued that our high-frequent, manual patch attacks qualitatively improve the optical flow predictions of LGS- and ILP-defended methods, as a result of the significant quality degradation of defenses on unattacked images.
Here, we provide the corresponding quality scores over the whole set of KITTI frames.
To this end, we quantify the quality $Q_\text{D}^\text{A}=\text{EPE}(\flow^*,\flow_\text{D}^\text{A})$, \ie the distance between ground truth flow $\flow^*$ and optical flow predictions $\flow_\text{D}^\text{A}$ of methods that are defended with D and attacked with A.

\begin{figure*}
    \centering
    \setlength{\tabcolsep}{0.3pt}
    \setlength{\fboxsep}{0pt}
    \setlength{\fboxrule}{.1pt}
        \begin{tabular}{cc@{\quad}cc@{\quad}cc}
            \toprule
            \multicolumn{2}{c}{No defense}                                              & \multicolumn{2}{c}{LGS defense}                                                                                 & \multicolumn{2}{c}{ILP defense}                                                                                                                                                                                                                                                                                                                                                               \\
            \midrule
            \leftrotbox{width=0.15\textwidth}{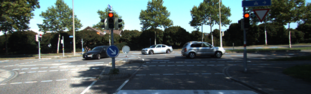}{FNC}    & \fbox{\includegraphics[width=0.15\textwidth]{graphics/vanilla_effects/0_FlowNetC_F_None_None.png}}       & \includegraphics[width=0.15\textwidth]{graphics/vanilla_effects/0_I1_None_LGS.png} & \fbox{\includegraphics[width=0.15\textwidth]{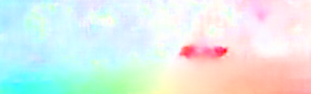}}       & \includegraphics[width=0.15\textwidth]{graphics/vanilla_effects/0_I1_None_ILP.png} & \fbox{\includegraphics[width=0.15\textwidth]{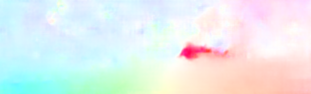}}       \\[-.3em]
            \leftrotbox{width=0.15\textwidth}{graphics/vanilla_effects/0_I1.png}{FNCF}   & \fbox{\includegraphics[width=0.15\textwidth]{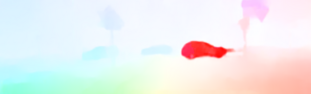}} & \includegraphics[width=0.15\textwidth]{graphics/vanilla_effects/0_I1_None_LGS.png} & \fbox{\includegraphics[width=0.15\textwidth]{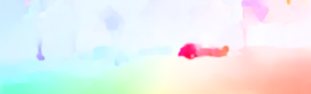}} & \includegraphics[width=0.15\textwidth]{graphics/vanilla_effects/0_I1_None_ILP.png} & \fbox{\includegraphics[width=0.15\textwidth]{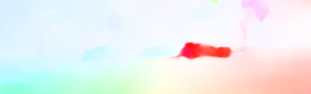}} \\[-.3em]
            \leftrotbox{width=0.15\textwidth}{graphics/vanilla_effects/0_I1.png}{PWC}    & \fbox{\includegraphics[width=0.15\textwidth]{graphics/vanilla_effects/0_PWCNet_F_None_None.png}}         & \includegraphics[width=0.15\textwidth]{graphics/vanilla_effects/0_I1_None_LGS.png} & \fbox{\includegraphics[width=0.15\textwidth]{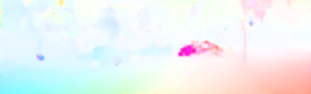}}         & \includegraphics[width=0.15\textwidth]{graphics/vanilla_effects/0_I1_None_ILP.png} & \fbox{\includegraphics[width=0.15\textwidth]{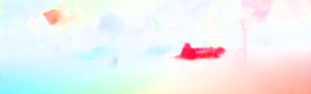}}         \\[-.3em]
            \leftrotbox{width=0.15\textwidth}{graphics/vanilla_effects/0_I1.png}{SpyNet} & \fbox{\includegraphics[width=0.15\textwidth]{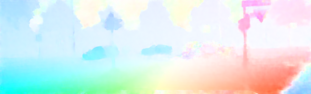}}         & \includegraphics[width=0.15\textwidth]{graphics/vanilla_effects/0_I1_None_LGS.png} & \fbox{\includegraphics[width=0.15\textwidth]{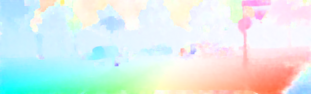}}         & \includegraphics[width=0.15\textwidth]{graphics/vanilla_effects/0_I1_None_ILP.png} & \fbox{\includegraphics[width=0.15\textwidth]{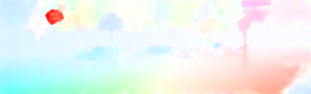}}         \\[-.3em]
            \leftrotbox{width=0.15\textwidth}{graphics/vanilla_effects/0_I1.png}{RAFT}   & \fbox{\includegraphics[width=0.15\textwidth]{graphics/vanilla_effects/0_RAFT_F_None_None.png}}           & \includegraphics[width=0.15\textwidth]{graphics/vanilla_effects/0_I1_None_LGS.png} & \fbox{\includegraphics[width=0.15\textwidth]{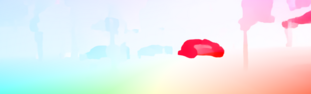}}           & \includegraphics[width=0.15\textwidth]{graphics/vanilla_effects/0_I1_None_ILP.png} & \fbox{\includegraphics[width=0.15\textwidth]{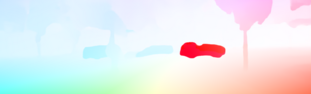}}           \\[-.3em]
            \leftrotbox{width=0.15\textwidth}{graphics/vanilla_effects/0_I1.png}{GMA}    & \fbox{\includegraphics[width=0.15\textwidth]{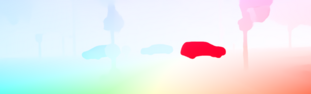}}            & \includegraphics[width=0.15\textwidth]{graphics/vanilla_effects/0_I1_None_LGS.png} & \fbox{\includegraphics[width=0.15\textwidth]{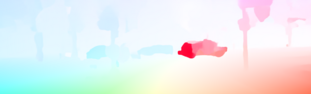}}            & \includegraphics[width=0.15\textwidth]{graphics/vanilla_effects/0_I1_None_ILP.png} & \fbox{\includegraphics[width=0.15\textwidth]{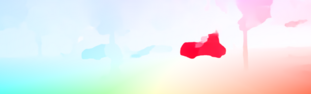}}            \\[-.3em]
            \leftrotbox{width=0.15\textwidth}{graphics/vanilla_effects/0_I1.png}{FF}     & \fbox{\includegraphics[width=0.15\textwidth]{graphics/vanilla_effects/0_FlowFormer_F_None_None.png}}     & \includegraphics[width=0.15\textwidth]{graphics/vanilla_effects/0_I1_None_LGS.png} & \fbox{\includegraphics[width=0.15\textwidth]{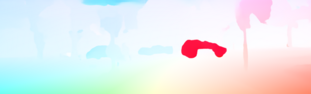}}     & \includegraphics[width=0.15\textwidth]{graphics/vanilla_effects/0_I1_None_ILP.png} & \fbox{\includegraphics[width=0.15\textwidth]{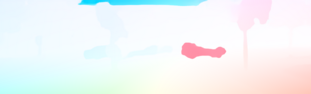}}     \\[-.3em]
            \bottomrule
        \end{tabular}
    \caption{Optical flow prediction on an unattacked frame of the KITTI dataset for optical flow methods with different defenses. Defenses from left to right: None, LGS and ILP. See \cref{fig:supp:defense_effects2} for more samples.}
    \label{fig:supp:defense_effects}
\end{figure*}

\begin{figure*}
    \centering
    \setlength{\tabcolsep}{0.3pt}
    \setlength{\fboxsep}{0pt}
    \setlength{\fboxrule}{.1pt}
        \begin{tabular}{cc@{\quad}cc@{\quad}cc}
            \toprule
            \multicolumn{2}{c}{No defense}                                              & \multicolumn{2}{c}{LGS defense}                                                                                 & \multicolumn{2}{c}{ILP defense}                                                                                                                                                                                                                                                                                                                                                               \\
            \midrule
            \leftrotbox{width=0.15\textwidth}{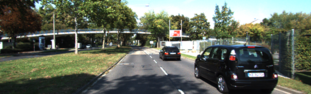}{FNC}    & \fbox{\includegraphics[width=0.15\textwidth]{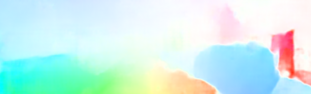}}       & \includegraphics[width=0.15\textwidth]{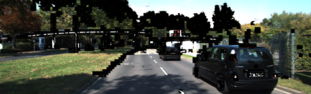} & \fbox{\includegraphics[width=0.15\textwidth]{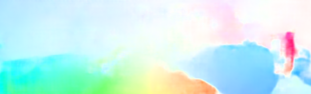}}       & \includegraphics[width=0.15\textwidth]{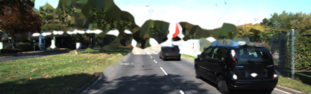} & \fbox{\includegraphics[width=0.15\textwidth]{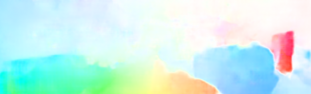}}       \\[-.3em]
            \leftrotbox{width=0.15\textwidth}{graphics/vanilla_effects/1_I1.png}{FNCF}   & \fbox{\includegraphics[width=0.15\textwidth]{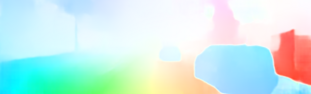}} & \includegraphics[width=0.15\textwidth]{graphics/vanilla_effects/1_I1_None_LGS.png} & \fbox{\includegraphics[width=0.15\textwidth]{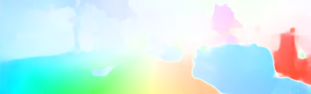}} & \includegraphics[width=0.15\textwidth]{graphics/vanilla_effects/1_I1_None_ILP.png} & \fbox{\includegraphics[width=0.15\textwidth]{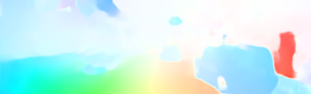}} \\[-.3em]
            \leftrotbox{width=0.15\textwidth}{graphics/vanilla_effects/1_I1.png}{PWC}    & \fbox{\includegraphics[width=0.15\textwidth]{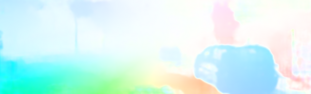}}         & \includegraphics[width=0.15\textwidth]{graphics/vanilla_effects/1_I1_None_LGS.png} & \fbox{\includegraphics[width=0.15\textwidth]{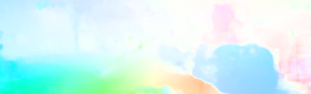}}         & \includegraphics[width=0.15\textwidth]{graphics/vanilla_effects/1_I1_None_ILP.png} & \fbox{\includegraphics[width=0.15\textwidth]{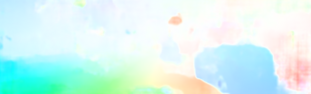}}         \\[-.3em]
            \leftrotbox{width=0.15\textwidth}{graphics/vanilla_effects/1_I1.png}{SpyNet} & \fbox{\includegraphics[width=0.15\textwidth]{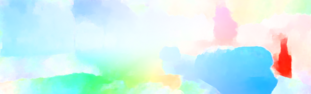}}         & \includegraphics[width=0.15\textwidth]{graphics/vanilla_effects/1_I1_None_LGS.png} & \fbox{\includegraphics[width=0.15\textwidth]{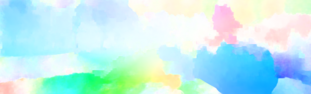}}         & \includegraphics[width=0.15\textwidth]{graphics/vanilla_effects/1_I1_None_ILP.png} & \fbox{\includegraphics[width=0.15\textwidth]{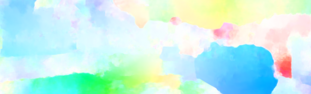}}         \\[-.3em]
            \leftrotbox{width=0.15\textwidth}{graphics/vanilla_effects/1_I1.png}{RAFT}   & \fbox{\includegraphics[width=0.15\textwidth]{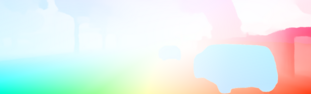}}           & \includegraphics[width=0.15\textwidth]{graphics/vanilla_effects/1_I1_None_LGS.png} & \fbox{\includegraphics[width=0.15\textwidth]{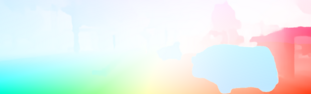}}           & \includegraphics[width=0.15\textwidth]{graphics/vanilla_effects/1_I1_None_ILP.png} & \fbox{\includegraphics[width=0.15\textwidth]{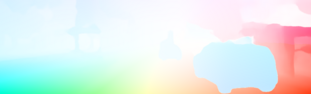}}           \\[-.3em]
            \leftrotbox{width=0.15\textwidth}{graphics/vanilla_effects/1_I1.png}{GMA}    & \fbox{\includegraphics[width=0.15\textwidth]{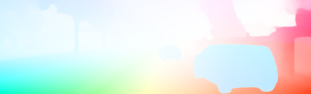}}            & \includegraphics[width=0.15\textwidth]{graphics/vanilla_effects/1_I1_None_LGS.png} & \fbox{\includegraphics[width=0.15\textwidth]{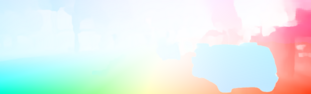}}            & \includegraphics[width=0.15\textwidth]{graphics/vanilla_effects/1_I1_None_ILP.png} & \fbox{\includegraphics[width=0.15\textwidth]{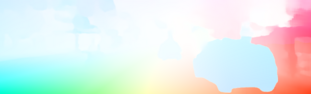}}            \\[-.3em]
            \leftrotbox{width=0.15\textwidth}{graphics/vanilla_effects/1_I1.png}{FF}     & \fbox{\includegraphics[width=0.15\textwidth]{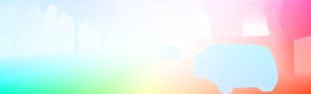}}     & \includegraphics[width=0.15\textwidth]{graphics/vanilla_effects/1_I1_None_LGS.png} & \fbox{\includegraphics[width=0.15\textwidth]{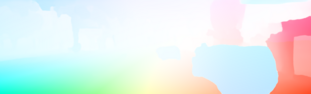}}     & \includegraphics[width=0.15\textwidth]{graphics/vanilla_effects/1_I1_None_ILP.png} & \fbox{\includegraphics[width=0.15\textwidth]{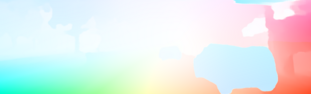}}     \\[-.3em]
            \bottomrule
        \end{tabular}
    \caption{Optical flow prediction on an unattacked frame of the KITTI dataset for optical flow methods with different defenses. Defenses from left to right: None, LGS and ILP. See \cref{fig:supp:defense_effects} for more samples.}
    \label{fig:supp:defense_effects2}
\end{figure*}

\cref{tab:defendedDistanceToGroundTruth} provides the quality scores for unattacked but defended networks (block 1, corresponds to values from Main \cref{tab:accur}), for full pipelines where the defended method is attacked with the corresponding defense-awareness (block 2) and for our manual attack on defended networks (block 3).
Compared to the original baseline $Q=\text{EPE}(\flow^*,\flow)$ (block 1, marked in gray), all defenses decrease the quality.
\cref{fig:supp:defense_effects} and \cref{fig:supp:defense_effects2} visualize the flow output for unattacked models on KITTI samples when no defense, LGS or ILP is applied.

Then, we begin by comparing the quality for defended methods in the first two blocks, \ie we exclude the manual patch attack in block 3, and mark the best quality \textbf{bold} in the table.
Note that we exclude the gray rows, as they contain the quality for \emph{undefended} methods.
For optical flow methods that have a good undefended quality $Q$, \ie FlowNetCRobust, RAFT, GMA and FlowFormer, we find that a defense-aware attack on a defended model actually yields a better quality than the defended but unattacked model: $Q_\text{D}^\text{D} > Q_\text{D}$.
For these methods, a noisy patch was revealed to be the most effective.
Hence, it is easier for an adaptive attack to exploit the changes introduced by the defense than to influence the flow estimation.

Now we also include the manual patch attack in the defense evaluation, again \underline{underlining} the highest-quality flow per method over all three blocks in \cref{tab:defendedDistanceToGroundTruth}.
Again we exclude the gray rows that contain the quality for \emph{undefended} methods in order to compare the influence of the defenses.
Now, for almost all methods the best defended quality is achieved for manual patch attacks.
When we compare the underlined numbers to the baseline quality $Q$, we find that our manual patch attack almost restores the undefended and unattacked quality for our defended methods outside the patch area.
While this underlines the finding from the Main paper that the low quality of defended but unattacked methods is the main reason for the low quality (and robustness) of defended methods, it also yields another point:
If the defenses did not deteriorate the unattacked quality, they could be effective in terms of quality and robustness because they restore high-quality optical flow fields in the presence of adversarial-like patches.

\subsection{Defense evaluation on additional datasets}

We evaluate the defenses and their effectiveness on more datasets besides KITTI~\cite{kitti2015}, and consider Sintel~\cite{Butler2012NaturalisticOpenSource}, Driving~\cite{Mayer2016LargeDatasetTrain}, HD1K~\cite{Kondermann2016HciBenchmarkSuite} and Spring~\cite{Mehl2023SpringHighResolution}.
Because evaluating the defended quality requires ground truth optical flow data, we use validation splits of the respective test sets for all datasets.
Dataset-specific patches are then trained on the remaining training data.
For Sintel, we use the validation set from~\cite{Yang2019VolumetricCorrespondenceNetworks} which splits Sintel-test such that the flow magnitudes of the validation set match the flow-magnitude distribution of the full training set~\cite{Sun2022DisentanglingArchitectureTraining}.
For HD1K and Spring, we are unaware of flow-magnitude matching validation splits in the literature, and create validation splits with matching flow-magnitude distributions as detailed in \cref{tab:validationsplits}.
For Driving, we use the scenes with focal length 15mm, forwards, fast speed and left camera as validation split.
Note that during our evaluations, we half the image resolution for HD1K and Spring, to keep the image sizes and hence results for patches with size 100 comparable across all datasets.

\begin{table}
   \caption{Validation split details for the evaluation datasets. \enquote{Frames} denotes frame pairs (for the optical flow calculation) rather than single frames, if \enquote{Half} is checked the frame size is halved. If all scenes except the validation scenes make up the set for training patches, the training scenes are marked with \enquote{EV: except validation}. \enquote{OFM-id} denotes \enquote{optical flow magnitude in-distribution}, meaning the validation set is in-distribution w.r.t.\ to the optical flow magnitude distribution of the original training set. \enquote{I3} means that only every third frame pair of the validation scenes is added to the validation split.}
   \label{tab:validationsplits}
   \small
   \centering
   \scalebox{0.79}{
   \begin{tabular}{@{}l@{\ }c@{\ }p{3.5cm}@{\ \ }c@{\ \ }c@{\ \ }p{2cm}@{}}
   \toprule
   Dataset & \rotatebox{60}{Val. frames} & \rotatebox{60}{Val. scenes} & \rotatebox{60}{Half} & \rotatebox{60}{Train. scenes} & \rotatebox{60}{Notes} \\
   \midrule
   KITTI~\cite{kitti2015} & 200 & KITTI-train~\cite{kitti2015} & -- & Raw~\cite{kittiRaw} & Split from~\cite{Ranjan2019}\\
   \midrule
   Sintel~\cite{Butler2012NaturalisticOpenSource} & \phantom{0}89 & ambush2, bamboo2, cave2, market2, shaman2, temple2 & -- & EV & Split from~\cite{Yang2019VolumetricCorrespondenceNetworks} OFM-id,  I3\\
   \midrule
   Driving~\cite{Mayer2016LargeDatasetTrain} & 299 & 15mm focal length, scene forwards, fast, into future, left & -- & EV & \\
   \midrule
   HD1K~\cite{Kondermann2016HciBenchmarkSuite} & \phantom{0}94 & 000009, 000013, 000018, 000019, 000032 & \checkmark & EV & OFM-id\\
   \midrule
   Spring~\cite{Mehl2023SpringHighResolution} & 658 & 0002, 0010, 0018, 0026, 0032, 0045 & \checkmark & EV & OFM-id\\
   \bottomrule
   \end{tabular}
   }
\end{table}

\begin{figure*}
\centering
        \begin{subfigure}[c]{0.32\textwidth}
            \includegraphics[width=\textwidth]{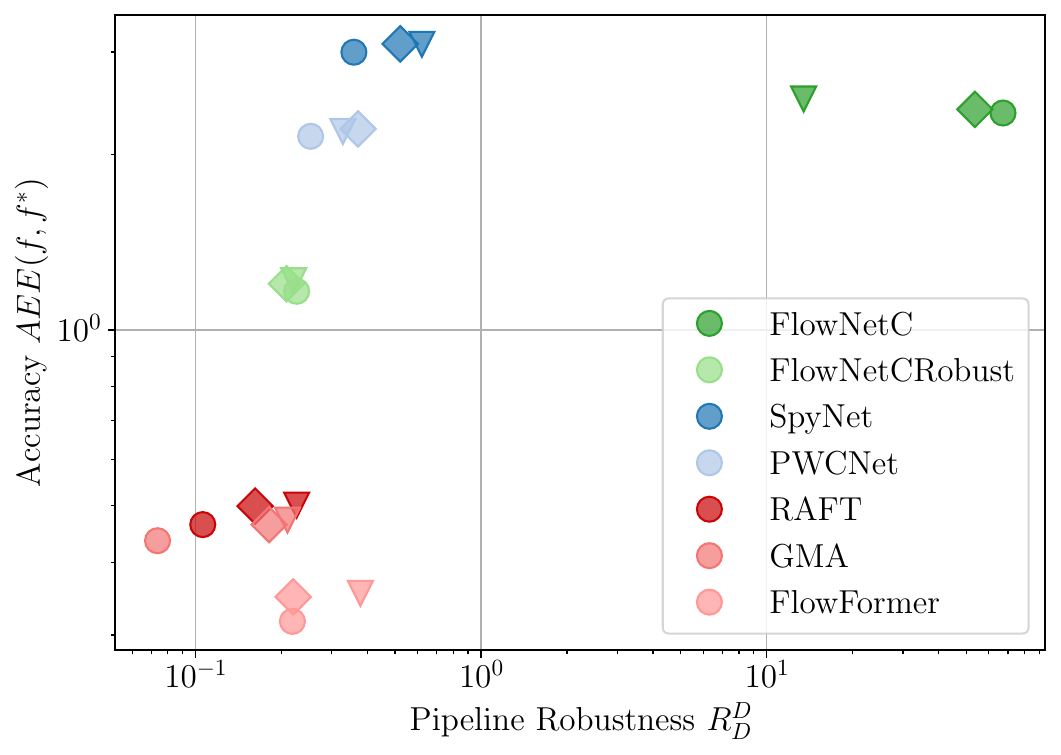}
            \caption{HD1K}
            \label{sfig:HD1K}
        \end{subfigure}
        \hfill
        \begin{subfigure}[c]{0.32\textwidth}
            \includegraphics[width=\textwidth]{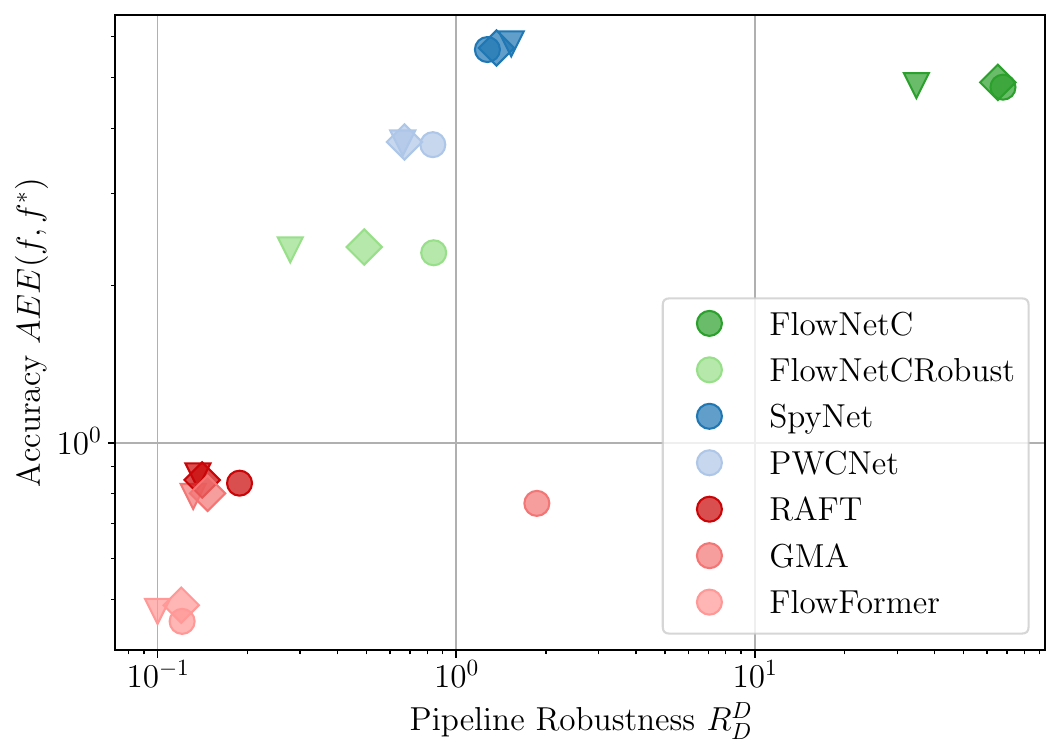}
            \caption{Sintel-clean}
            \label{sfig:Sintel-c}
        \end{subfigure}
        \hfill
        \begin{subfigure}[c]{0.32\textwidth}
            \includegraphics[width=\textwidth]{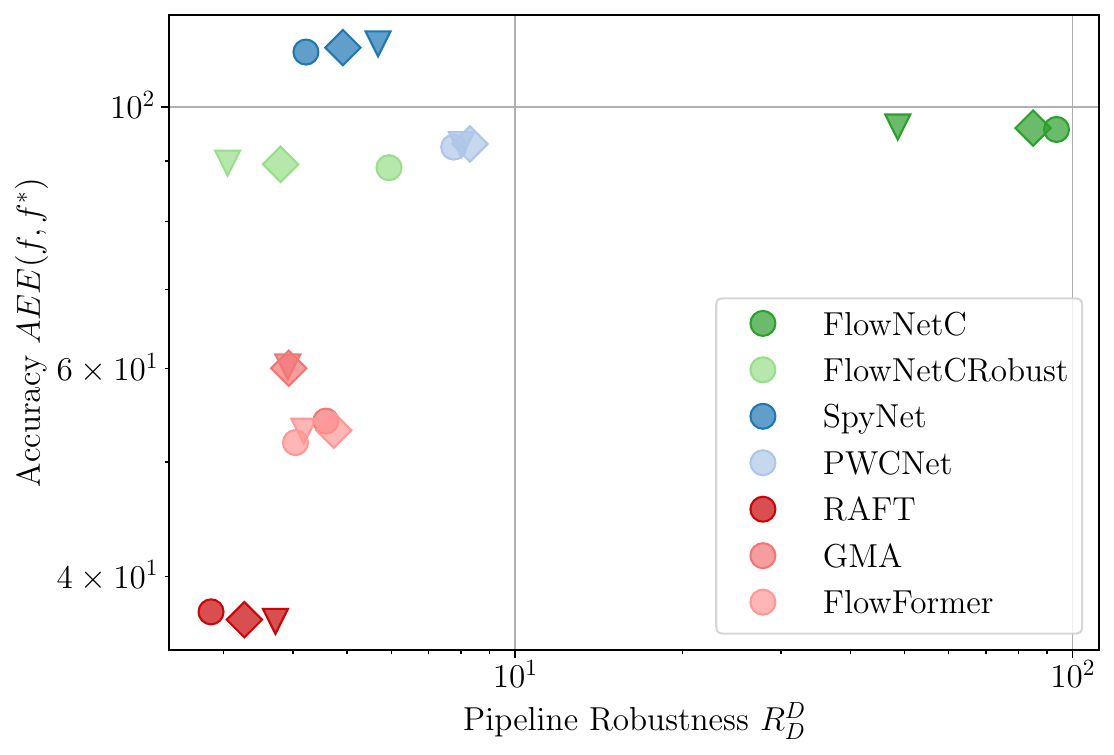}
            \caption{Driving-clean}
            \label{sfig:Driving-c}
        \end{subfigure}
        ~
        \begin{subfigure}[c]{0.32\textwidth}
            \includegraphics[width=\textwidth]{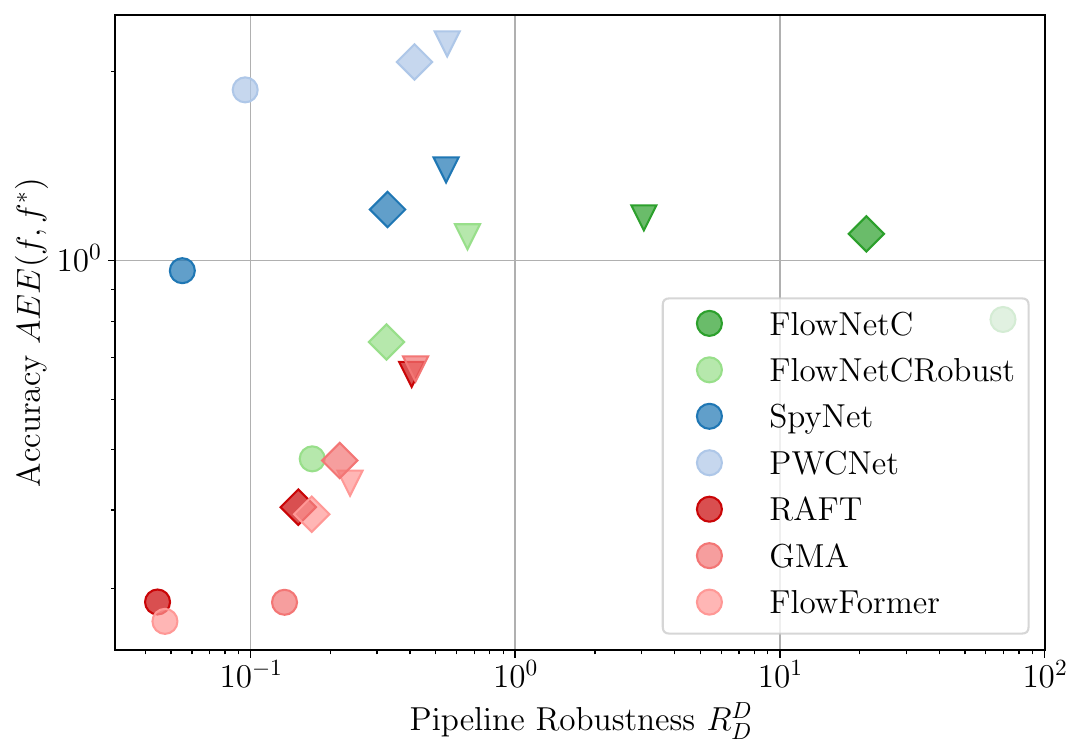}
            \caption{Spring}
            \label{sfig:Spring}
        \end{subfigure}
        \hfill
        \begin{subfigure}[c]{0.32\textwidth}
            \includegraphics[width=\textwidth]{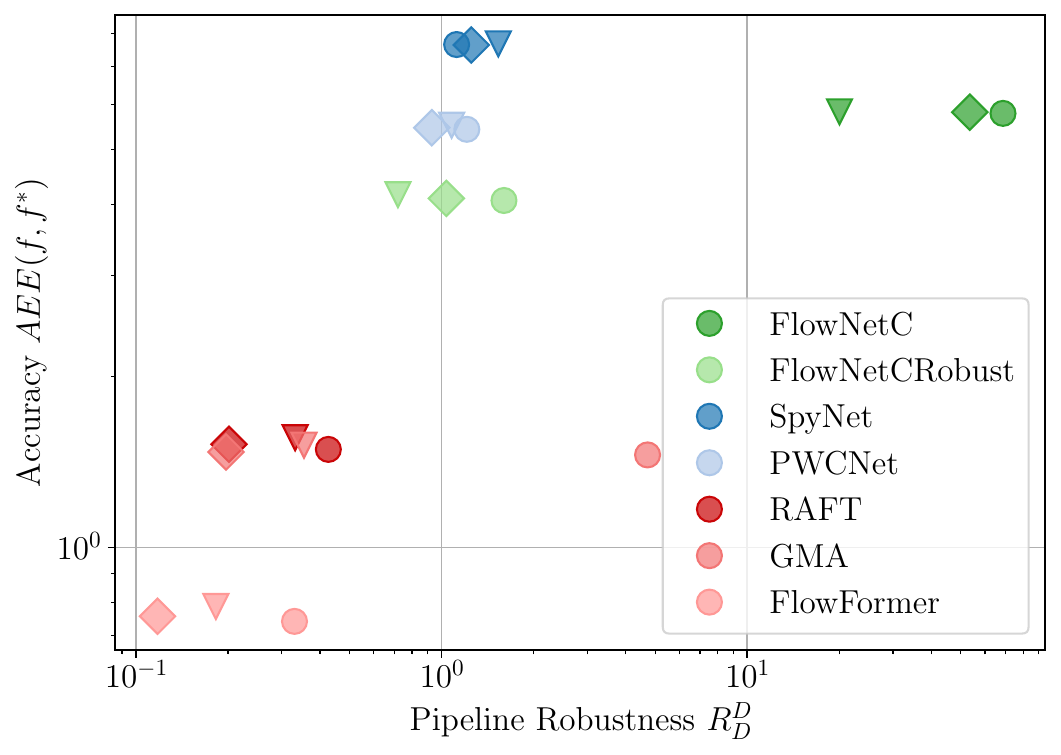}
            \caption{Sintel-final}
            \label{sfig:Sintel-f}
        \end{subfigure}
        \hfill
        \begin{subfigure}[c]{0.32\textwidth}
            \includegraphics[width=\textwidth]{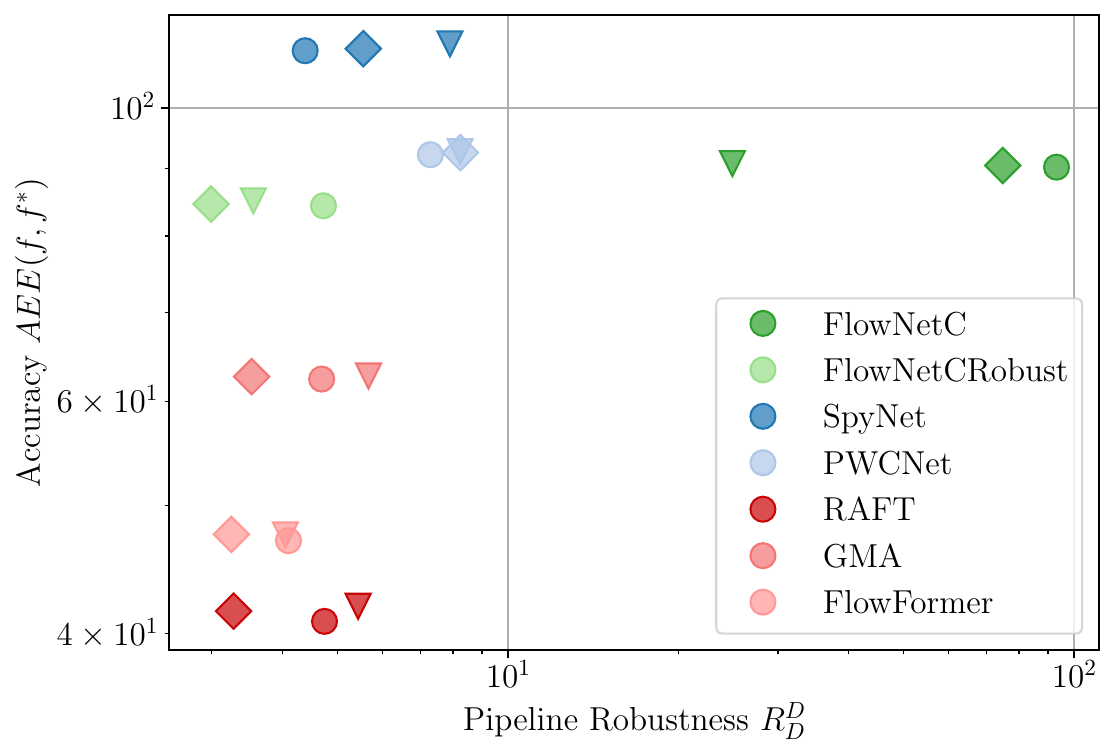}
            \caption{Driving-final}
            \label{sfig:Driving-f}
        \end{subfigure}
    \caption{Quality vs. robustness of flow networks on different datasets in a double logarithmic plot.
   An ideal method would be in the origin. Undefended networks are circles~$\bigcirc$, networks defended with LGS are triangles~$\triangledown$ and networks defended with ILP are diamonds~$\diamondsuit$.}
    \label{fig:qual-rob-datasets}
\end{figure*}

\begin{figure*}
\centering
        \begin{tabular}{cc}
        Gradient magitude (LGS) & Laplace (ILP)
        \\[-1mm]
            \includegraphics[width=0.48\textwidth]{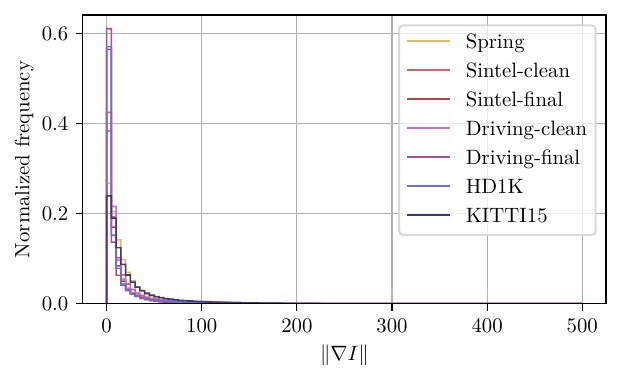}
            &
            \includegraphics[width=0.48\textwidth]{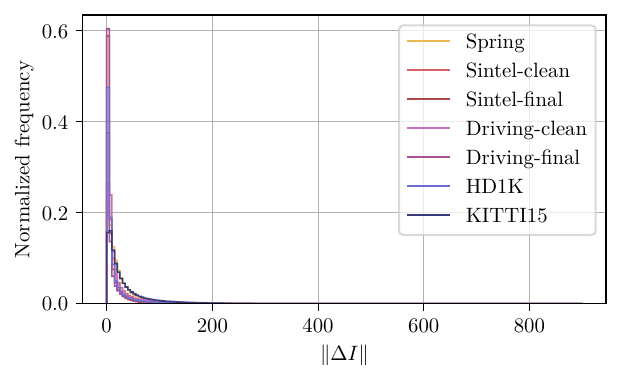}
            \\[-3mm]
            \includegraphics[width=0.48\textwidth]{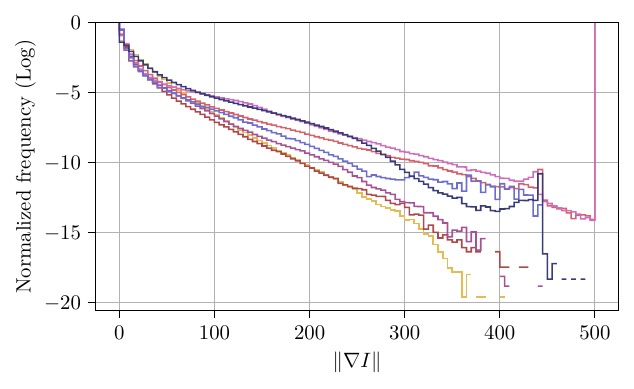}
            &
            \includegraphics[width=0.48\textwidth]{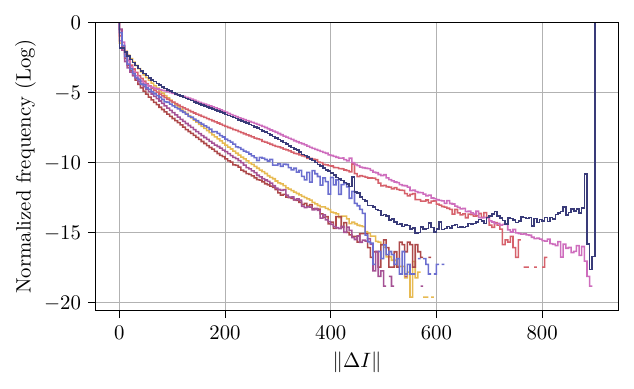}
        \end{tabular}
    \vspace*{-1.\baselineskip}
    \caption{Image statistics for optical flow datasets. The plots show the histograms over the magnitude of first and second image derivatives for different optical flow datasets, where the LGS defense considers first (left) and ILP considers second (right) derivatives. The histograms are normalized by the number of pixels in the respective dataset. The top row shows the pure histograms, while the bottom row shows the log-transformed frequency for better visualization of statistics for large gradient magnitudes, which are filtered by the defenses.}
    \label{fig:dataset-statistics}
\end{figure*}

For the datasets HD1K, Spring, Sintel (clean and final) and Driving (clean and final), we show the numerical results of the defended quality analysis in \cref{tab:accur-HD1K}, \cref{tab:accur-Spring}, \cref{tab:accur-Sintel} and \cref{tab:accur-Driving}, and the respective robustness analyses in \cref{tab:rob-HD1K}, \cref{tab:rob-Spring}, \cref{tab:rob-Sintel} and \cref{tab:rob-Driving}.
For a better overview, \cref{fig:qual-rob-datasets} shows the quality vs.\ robustness plots for all tested optical flow methods on all tested datasets, which can be compared to the results on KITTI in Main Fig.\ 5.

\begin{table}
   \caption[Qual HD1K caption]{Quality $\text{Q}_\text{D}=\text{EPE}(\flow^*, \flow_\text{D})$ for optical flow pipelines with defense D on the HD1K~\cite{Kondermann2016HciBenchmarkSuite} validation split\footnotemark; Best quality is \textbf{bold}. All defenses lead to a worse quality on unattacked frames.}
   \label{tab:accur-HD1K}
   \small
   \centering
   \begin{tabular}{l@{\ \ }l|r@{\ \ \ }r@{\ \ \ }r@{\ \ \ }r@{\ \ \ }r@{\ \ \ }r@{\ \ \ }r}
      \toprule
      \multicolumn{2}{l}{\rotatebox{0}{Defense}}
      & \rotatebox{60}{FNC} & \rotatebox{60}{FNCR} & \rotatebox{60}{SpyNet} & \rotatebox{60}{PWC} & \rotatebox{60}{RAFT} & \rotatebox{60}{GMA} & \rotatebox{60}{FF}  \\
      \midrule
      None   & $\text{Q}$             &  \textbf{2.36} &          \textbf{1.17} &  \textbf{3.00} &  \textbf{2.15} & \textbf{0.46} & \textbf{0.44} &      \textbf{0.32} \\
      LGS    & $\text{Q}_\text{LGS}$  &   2.49 &           1.22 &   3.09 &   2.19 &  0.50 &  0.47 &       0.35 \\
      ILP    & $\text{Q}_\text{ILP}$  &   2.39 &           1.20 &   3.10 &   2.21 &  0.50 &  0.46 &       0.35 \\
      \bottomrule
   \end{tabular}
\end{table}
\footnotetext{\label{footn:valsplits}See \cref{tab:validationsplits} for details on the used validation splits}

\begin{table}
   \caption[Qual Spring caption]{Quality $\text{Q}_\text{D}=\text{EPE}(\flow^*, \flow_\text{D})$ for optical flow pipelines with defense D on the Spring~\cite{Mehl2023SpringHighResolution} validation split\cref{footn:valsplits}; Best quality is \textbf{bold}. All defenses lead to a worse quality on unattacked frames.}
   \label{tab:accur-Spring}
   \small
   \centering
   \begin{tabular}{l@{\ \ }l|r@{\ \ \ }r@{\ \ \ }r@{\ \ \ }r@{\ \ \ }r@{\ \ \ }r@{\ \ \ }r}
      \toprule
      \multicolumn{2}{l}{\rotatebox{0}{Defense}}
      & \rotatebox{60}{FNC} & \rotatebox{60}{FNCR} & \rotatebox{60}{SpyNet} & \rotatebox{60}{PWC} & \rotatebox{60}{RAFT} & \rotatebox{60}{GMA} & \rotatebox{60}{FF}  \\
      \midrule
      None   & $\text{Q}$             &  \textbf{0.81} &           \textbf{0.48} &   \textbf{0.96} &   \textbf{1.87} &  \textbf{0.29} &  \textbf{0.29} &       \textbf{0.27} \\
      LGS    & $\text{Q}_\text{LGS}$  &  1.17 &           1.09 &   1.39 &   2.21 &  0.66 &  0.67 &       0.44 \\
      ILP    & $\text{Q}_\text{ILP}$  &  1.10 &           0.74 &   1.21 &   2.07 &  0.40 &  0.48 &       0.39 \\
      \bottomrule
   \end{tabular}
\end{table}

\begin{table}
   \caption{Quality $\text{Q}_\text{D}=\text{EPE}(\flow^*, \flow_\text{D})$ for optical flow pipelines with defense D on the Sintel~\cite{Butler2012NaturalisticOpenSource} clean and final validation splits\cref{footn:valsplits}~\cite{Yang2019VolumetricCorrespondenceNetworks}; Best quality is \textbf{bold}. All defenses lead to a worse quality on unattacked frames.}
   \label{tab:accur-Sintel}
   \small
   \centering
   \begin{tabular}{l@{\ \ }l|r@{\ \ \ }r@{\ \ \ }r@{\ \ \ }r@{\ \ \ }r@{\ \ \ }r@{\ \ \ }r}
      \toprule
      \multicolumn{2}{l}{\rotatebox{0}{Defense}}
      & \rotatebox{60}{FNC} & \rotatebox{60}{FNCR} & \rotatebox{60}{SpyNet} & \rotatebox{60}{PWC} & \rotatebox{60}{RAFT} & \rotatebox{60}{GMA} & \rotatebox{60}{FF}  \\
      \midrule
      \multicolumn{2}{l}{} & \multicolumn{7}{c}{\rotatebox{0}{clean}}\\
      \cmidrule(l{0em}r{0em}){3-9}
      None   & $\text{Q}$             & \textbf{4.80} &           \textbf{2.31} &   \textbf{5.66} &   \textbf{3.72} &  \textbf{0.84} &  \textbf{0.77} &       \textbf{0.45}\\
      LGS    & $\text{Q}_\text{LGS}$  & 4.83 &           2.34 &   5.81 &   3.75 &  0.86 &  0.79 &       0.48\\
      ILP    & $\text{Q}_\text{ILP}$  & 4.90 &           2.37 &   5.70 &   3.77 &  0.85 &  0.80 &       0.49\\
      \midrule
      \multicolumn{2}{l}{} & \multicolumn{7}{c}{\rotatebox{0}{final}}\\
      \cmidrule(l{0em}r{0em}){3-9}
      None   & $\text{Q}$             &   \textbf{5.79} &           \textbf{4.07} &   7.64 &   \textbf{5.42} &  \textbf{1.49} &  \textbf{1.45} &       \textbf{0.74} \\
      LGS    & $\text{Q}_\text{LGS}$  &   5.82 &           4.16 &   7.66 &   5.51 &  1.56 &  1.51 &       0.79 \\
      ILP    & $\text{Q}_\text{ILP}$  &   5.81 &           4.10 &   \textbf{7.63} &   5.46 &  1.52 &  1.47 &       0.76 \\
      \bottomrule
   \end{tabular}
\end{table}

\begin{table}
   \caption{Quality $\text{Q}_\text{D}=\text{EPE}(\flow^*, \flow_\text{D})$ for optical flow pipelines with defense D on the Driving~\cite{Mayer2016LargeDatasetTrain} clean and final validation splits\cref{footn:valsplits}; Best quality is \textbf{bold}. All defenses lead to a worse quality on unattacked frames.}
   \label{tab:accur-Driving}
   \small
   \centering
   \scalebox{0.93}{
   \begin{tabular}{l@{\ \ }l|r@{\ \ \ }r@{\ \ \ }r@{\ \ \ }r@{\ \ \ }r@{\ \ \ }r@{\ \ \ }r}
      \toprule
      \multicolumn{2}{l}{\rotatebox{0}{Defense}}
      & \rotatebox{60}{FNC} & \rotatebox{60}{FNCR} & \rotatebox{60}{SpyNet} & \rotatebox{60}{PWC} & \rotatebox{60}{RAFT} & \rotatebox{60}{GMA} & \rotatebox{60}{FF}  \\
      \midrule
      \multicolumn{2}{l}{} & \multicolumn{7}{c}{\rotatebox{0}{clean}}\\
      \cmidrule(l{0em}r{0em}){3-9}
      None   & $\text{Q}$             & \textbf{95.73} &          \textbf{88.85} &  \textbf{111.36} &  92.49 &  37.32 &  \textbf{54.17} &      \textbf{51.94}\\
      LGS    & $\text{Q}_\text{LGS}$  & 96.20 &          89.64 &  113.15 &   92.93 &  \textbf{36.63} &  60.21 &      53.11\\
      ILP    & $\text{Q}_\text{ILP}$  & 95.99 &          89.43 &  112.32 & \textbf{93.06} &  36.75 &  60.06 &      53.21\\
      \midrule
      \multicolumn{2}{l}{} & \multicolumn{7}{c}{\rotatebox{0}{final}}\\
      \cmidrule(l{0em}r{0em}){3-9}
      None   & $\text{Q}$             & \textbf{90.21} &          \textbf{84.34} &  \textbf{110.52} &  \textbf{92.19} &  \textbf{40.87} &  \textbf{62.35} &      \textbf{47.05} \\
      LGS    & $\text{Q}_\text{LGS}$  & 90.78 &          85.07 &  111.84 &  92.70 &  41.96 &  62.68 &      47.52 \\
      ILP    & $\text{Q}_\text{ILP}$  & 90.47 &          84.59 &  110.91 &  92.52 &  41.60 &  62.61 &      47.56 \\
      \bottomrule
   \end{tabular}
   }
\end{table}

\begin{table*}
   \caption{Robustness scores for all combinations of defended methods and defense-aware attacks on optical flow methods on the HD1K~\cite{Kondermann2016HciBenchmarkSuite} validation split\cref{footn:valsplits}. Per attack, the robustness values of the best defense are \textbf{bold}. Per defense, the robustness values for the attack it is most vulnerable to are \underline{underlined}.
   Full pipelines are highlighted in gray, and provide the corresponding robustness values to the quality scores from \cref{tab:accur-HD1K}.
   }
   \label{tab:rob-HD1K}
   \small
   \centering
   \begin{tabular}{lll@{\quad}|@{\quad}rrrrrrr}
      \toprule
      Attack type                    & \multicolumn{2}{l}{Defense}                        & \multicolumn{1}{c}{\rotatebox{0}{FNC}} & \multicolumn{1}{c}{\rotatebox{0}{FNCR}}  & \multicolumn{1}{c}{\rotatebox{0}{SpyNet}} & \multicolumn{1}{c}{\rotatebox{0}{PWC}} & \multicolumn{1}{c}{\rotatebox{0}{RAFT}} & \multicolumn{1}{c}{\rotatebox{0}{GMA}} & \multicolumn{1}{c}{\rotatebox{0}{FF}} \\
      \midrule
      \multirow{3}{*}{Vanilla}       & \cellcolor{gray!20}None & \cellcolor{gray!20}$R^\text{Van}$             &  \cellcolor{gray!20}\underline{67.24} &           \cellcolor{gray!20}\underline{0.23} &   \cellcolor{gray!20}\textbf{0.36} &   \cellcolor{gray!20}\underline{0.25} &  \cellcolor{gray!20}\textbf{\underline{0.11}} &  \cellcolor{gray!20}\textbf{\underline{0.07}} &       \cellcolor{gray!20}\textbf{\underline{0.22}} \\
                                     & LGS  & $R^\text{Van}_\text{LGS}$                                        &   \textbf{0.45} &           0.21 &   0.59 &   0.25 &  \underline{0.23} &  0.17 &       0.36 \\
                                     & ILP  & $R^\text{Van}_\text{ILP}$                                        &   0.60 &           \textbf{0.17} &   0.51 &   \textbf{0.24} &  0.17 &  0.13 &       \underline{0.22} \\
      \cmidrule(l{0em}r{0em}){2-10}
      \multirow{3}{*}{+LGS (LGS-aware)}     & None & $R^\text{LGS}$                                            &  51.97 &           \textbf{0.07} &   \textbf{0.35} &   \textbf{0.22} &  \textbf{0.09} &  \textbf{0.06} &       \textbf{0.17} \\
                                     & \cellcolor{gray!20}LGS  & \cellcolor{gray!20}$R^\text{LGS}_\text{LGS}$  &  \cellcolor{gray!20}\underline{13.47} &           \cellcolor{gray!20}\underline{0.22} &   \cellcolor{gray!20}\underline{0.62} &   \cellcolor{gray!20}\underline{0.33} & \cellcolor{gray!20}0.23 & \cellcolor{gray!20}0.21 &      \cellcolor{gray!20}0.38 \\
                                     & ILP  & $R^\text{LGS}_\text{ILP}$                                        &  \textbf{10.02} &           0.17 &   0.51 &   0.35 &  \underline{0.18} &  \underline{0.18} &       0.22 \\
      \cmidrule(l{0em}r{0em}){2-10}
      \multirow{3}{*}{+ILP (ILP-aware)}     & None & $R^\text{ILP}$                                            &  60.52 &           \textbf{0.14} &   \textbf{\underline{0.36}} &   \textbf{0.24} &  \textbf{0.06} &  \textbf{0.06} &       \textbf{0.18} \\
                                     & LGS  & $R^\text{ILP}_\text{LGS}$                                        &   \textbf{2.89} &           0.21 &   0.62 &   0.27 &  0.23 &  \underline{0.21} &       \underline{0.38} \\
                                     & \cellcolor{gray!20}ILP  & \cellcolor{gray!20}$R^\text{ILP}_\text{ILP}$  &  \cellcolor{gray!20}\underline{53.63} &           \cellcolor{gray!20}\underline{0.21} &   \cellcolor{gray!20}\underline{0.52} &   \cellcolor{gray!20}\underline{0.37} & \cellcolor{gray!20}0.16 & \cellcolor{gray!20}0.18 &      \cellcolor{gray!20}0.22 \\
      \bottomrule
   \end{tabular}
\end{table*}

\begin{table*}
   \caption{Robustness scores for all combinations of defended methods and defense-aware attacks on optical flow methods on the Spring~\cite{Mehl2023SpringHighResolution} validation split\cref{footn:valsplits}. Per attack, the robustness values of the best defense are \textbf{bold}. Per defense, the robustness values for the attack it is most vulnerable to are \underline{underlined}.
   Full pipelines are highlighted in gray, and provide the corresponding robustness values to the quality scores from \cref{tab:accur-Spring}.
   }
   \label{tab:rob-Spring}
   \small
   \centering
   \begin{tabular}{lll@{\quad}|@{\quad}rrrrrrr}
      \toprule
      Attack type                    & \multicolumn{2}{l}{Defense}                        & \multicolumn{1}{c}{\rotatebox{0}{FNC}} & \multicolumn{1}{c}{\rotatebox{0}{FNCR}}  & \multicolumn{1}{c}{\rotatebox{0}{SpyNet}} & \multicolumn{1}{c}{\rotatebox{0}{PWC}} & \multicolumn{1}{c}{\rotatebox{0}{RAFT}} & \multicolumn{1}{c}{\rotatebox{0}{GMA}} & \multicolumn{1}{c}{\rotatebox{0}{FF}} \\
      \midrule
      \multirow{3}{*}{Vanilla}       & \cellcolor{gray!20}None & \cellcolor{gray!20}$R^\text{Van}$             & \cellcolor{gray!20}\underline{69.64} &           \cellcolor{gray!20}\underline{\textbf{0.17}} &   \cellcolor{gray!20}\textbf{0.06} &   \cellcolor{gray!20}\textbf{0.10} &  \cellcolor{gray!20}\underline{\textbf{0.04}} &  \cellcolor{gray!20}\underline{\textbf{0.13}} &       \cellcolor{gray!20}\underline{\textbf{0.05}} \\
                                     & LGS  & $R^\text{Van}_\text{LGS}$                                        &  0.62 &           0.67 &   0.55 &   0.54 &  0.40 &  \underline{0.42} &       \underline{0.24} \\
                                     & ILP  & $R^\text{Van}_\text{ILP}$                                        &  \textbf{0.58} &           0.31 &   0.33 &   0.36 &  0.15 &  \underline{0.22} &       \underline{0.17} \\
      \cmidrule(l{0em}r{0em}){2-10}
      \multirow{3}{*}{+LGS (LGS-aware)}     & None & $R^\text{LGS}$                                            & 33.27 &           \textbf{0.02} &   \underline{\textbf{0.06}} &   \textbf{0.16} &  \textbf{0.02} &  \textbf{0.02} &       \textbf{0.03} \\
                                     & \cellcolor{gray!20}LGS  & \cellcolor{gray!20}$R^\text{LGS}_\text{LGS}$  &  \cellcolor{gray!20}\underline{\textbf{3.06}} &           \cellcolor{gray!20}0.66 &   \cellcolor{gray!20}\underline{0.55} &   \cellcolor{gray!20}\underline{0.55} &  \cellcolor{gray!20}\underline{0.41} &  \cellcolor{gray!20}0.42 &       \cellcolor{gray!20}0.24 \\
                                     & ILP  & $R^\text{LGS}_\text{ILP}$                                        & 11.87 &           0.32 &   \underline{0.33} &   \underline{0.42} &  \underline{0.15} &  0.22 &       0.17 \\
      \cmidrule(l{0em}r{0em}){2-10}
      \multirow{3}{*}{+ILP (ILP-aware)}     & None & $R^\text{ILP}$                                            & 30.01 &           \textbf{0.10} &   \textbf{0.06} &   \underline{\textbf{0.19}} &  \textbf{0.01} &  \textbf{0.02} &       \textbf{0.03} \\
                                     & LGS  & $R^\text{ILP}_\text{LGS}$                                        &  \textbf{0.74} &           \underline{0.67} &   0.55 &   0.55 &  0.41 &  0.42 &       0.24 \\
                                     & \cellcolor{gray!20}ILP  & \cellcolor{gray!20}$R^\text{ILP}_\text{ILP}$  & \cellcolor{gray!20}\underline{21.22} &           \cellcolor{gray!20}\underline{0.33} &   \cellcolor{gray!20}0.33 &   \cellcolor{gray!20}0.42 &  \cellcolor{gray!20}0.15 &  \cellcolor{gray!20}0.22 &       \cellcolor{gray!20}0.17 \\
      \bottomrule
   \end{tabular}
\end{table*}

\begin{table*}
   \caption{Robustness scores for all combinations of defended methods and defense-aware attacks on optical flow methods on the Sintel~\cite{Butler2012NaturalisticOpenSource} final (f) and clean (c) validation splits\cref{footn:valsplits}~\cite{Yang2019VolumetricCorrespondenceNetworks}. Per attack, the robustness values of the best defense are \textbf{bold}. Per defense, the robustness values for the attack it is most vulnerable to are \underline{underlined}.
   Full pipelines are highlighted in gray, and provide the corresponding robustness values to the quality scores from \cref{tab:accur-Sintel}.
   }
   \label{tab:rob-Sintel}
   \small
   \centering
   \scalebox{.98}{
   \begin{tabular}{l@{\ \ \ }l@{\ \ \ }l@{\quad}|@{\quad}r@{\ \ \ }rr@{\ \ \ }rr@{\ \ \ }rr@{\ \ \ }rr@{\ \ \ }rr@{\ \ \ }rr@{\ \ \ }r}
      \toprule
      Attack type                    & \multicolumn{2}{l}{Defense}                        & \multicolumn{2}{c}{\rotatebox{0}{FNC}} & \multicolumn{2}{c}{\rotatebox{0}{FNCR}}  & \multicolumn{2}{c}{\rotatebox{0}{SpyNet}} & \multicolumn{2}{c}{\rotatebox{0}{PWC}} & \multicolumn{2}{c}{\rotatebox{0}{RAFT}} & \multicolumn{2}{c}{\rotatebox{0}{GMA}} & \multicolumn{2}{c}{\rotatebox{0}{FF}} \\
      \cmidrule(l{0.5em}r{0.5em}){4-5} \cmidrule(l{0.5em}r{0.5em}){6-7} \cmidrule(l{0.5em}r{0.5em}){8-9} \cmidrule(l{0.5em}r{0.5em}){10-11} \cmidrule(l{0.5em}r{0.5em}){12-13} \cmidrule(l{0.5em}r{0.5em}){14-15} \cmidrule(l{0.5em}r{0.5em}){16-17}
      & & \multicolumn{1}{c}{} & \multicolumn{1}{c}{f} & \multicolumn{1}{c}{c} & \multicolumn{1}{c}{f} & \multicolumn{1}{c}{c} & \multicolumn{1}{c}{f} & \multicolumn{1}{c}{c} & \multicolumn{1}{c}{f} & \multicolumn{1}{c}{c} & \multicolumn{1}{c}{f} & \multicolumn{1}{c}{c} & \multicolumn{1}{c}{f} & \multicolumn{1}{c}{c} & \multicolumn{1}{c}{f} & \multicolumn{1}{c}{c} \\
      \midrule
      \multirow{3}{*}{Vanilla}       & \cellcolor{gray!20}None & \cellcolor{gray!20}$R^\text{Van}$             &  \cellcolor{gray!20}\underline{68.71}  & \cellcolor{gray!20}67.84  &           \cellcolor{gray!20}\underline{1.60} &  \cellcolor{gray!20}\underline{0.84} &                 \cellcolor{gray!20}\textbf{1.12} &  \cellcolor{gray!20}\textbf{\underline{1.27}} &         \cellcolor{gray!20}\underline{1.21} &  \cellcolor{gray!20}\underline{0.84} &            \cellcolor{gray!20}\underline{0.43} & \cellcolor{gray!20}\underline{0.19} &             \cellcolor{gray!20}\underline{4.72} & \cellcolor{gray!20}\underline{1.86} &              \cellcolor{gray!20}\underline{0.33} & \cellcolor{gray!20}\underline{0.12} \\
                                     & LGS  & $R^\text{Van}_\text{LGS}$                                        &   \textbf{1.06}  &   \textbf{0.90}  &           0.73 &  0.25 &                 1.45 &  1.41 &         0.86 &  0.52 &            \underline{0.40} & \underline{0.17} &             0.34 & \underline{0.13} &              \underline{0.19} & \textbf{0.09} \\
                                     & ILP  & $R^\text{Van}_\text{ILP}$                                        &   1.49  &   1.76  &           \textbf{0.63} &  \textbf{0.29} &                 1.16 &  1.34 &         \textbf{0.74} &  \textbf{0.51} &            \textbf{0.19} & \textbf{0.14} &             \textbf{\underline{0.24}} & \textbf{0.13} &              \textbf{\underline{0.13}} & \underline{0.12} \\
      \cmidrule(l{0em}r{0em}){2-17}
      \multirow{3}{*}{+LGS (LGS-aware)}     & None & $R^\text{LGS}$                                            &  57.21  &  57.11  &           0.44 &  \textbf{0.20} &                 \textbf{1.15} &  \textbf{1.23} &         0.90 &  \textbf{0.61} &            0.23 & \textbf{0.10} &             0.20 & \textbf{0.09} &              0.16 & \textbf{0.07} \\
                                     & \cellcolor{gray!20}LGS  & \cellcolor{gray!20}$R^\text{LGS}_\text{LGS}$  &  \cellcolor{gray!20}\textbf{\underline{20.04}}  &  \cellcolor{gray!20}\textbf{\underline{34.78}}  &           \cellcolor{gray!20}\textbf{0.72} &  \cellcolor{gray!20}\underline{0.28} &                 \cellcolor{gray!20}\underline{1.53} & \cellcolor{gray!20}1.53 &         \cellcolor{gray!20}\underline{1.08} &  \cellcolor{gray!20}\underline{0.66} &           \cellcolor{gray!20}0.33 &\cellcolor{gray!20}0.14 &             \cellcolor{gray!20}\underline{0.35} &\cellcolor{gray!20}0.13 &             \cellcolor{gray!20}0.18 &\cellcolor{gray!20}0.10 \\
                                     & ILP  & $R^\text{LGS}_\text{ILP}$                                        &  31.42  &  40.56  &           0.63 &  0.35 &                 1.23 &  1.33 &         \textbf{0.88} &  0.64 &            \textbf{0.20} & 0.13 &             \textbf{0.19} & 0.15 &              \textbf{0.12} & 0.12 \\
      \cmidrule(l{0em}r{0em}){2-17}
      \multirow{3}{*}{+ILP (ILP-aware)}     & None & $R^\text{ILP}$                                            &  68.58  &  \underline{68.13}  &           1.13 &  0.55 &                 \textbf{\underline{1.17}} &  \textbf{1.26} &         0.96 &  0.65 &            \textbf{0.20} & \textbf{0.09} &             \textbf{0.19} & \textbf{0.09} &              0.19 & \textbf{0.08} \\
                                     & LGS  & $R^\text{ILP}_\text{LGS}$                                        &   \textbf{2.54}  &  \textbf{24.19}  &           \textbf{\underline{0.74}} &  \textbf{0.27} &                 1.53 &  \underline{1.53} &         0.98 &  \textbf{0.64} &            0.35 & 0.16 &             0.35 & 0.13 &              0.19 & \underline{0.10} \\
                                     & \cellcolor{gray!20}ILP  & \cellcolor{gray!20}$R^\text{ILP}_\text{ILP}$  &  \cellcolor{gray!20}\underline{53.52}  &  \cellcolor{gray!20}\underline{65.18}  &           \cellcolor{gray!20}\underline{1.04} &  \cellcolor{gray!20}\underline{0.49} &                 \cellcolor{gray!20}\underline{1.25} &  \cellcolor{gray!20}\underline{1.36} &         \cellcolor{gray!20}\textbf{\underline{0.93}} &  \cellcolor{gray!20}\underline{0.67} &            \cellcolor{gray!20}\underline{0.20} & \cellcolor{gray!20}\underline{0.14} &            \cellcolor{gray!20}0.20 & \cellcolor{gray!20}\underline{0.15} &              \cellcolor{gray!20}\textbf{0.12} &\cellcolor{gray!20}0.12 \\
      \bottomrule
   \end{tabular}
   }
\end{table*}

\begin{table*}
   \caption{Robustness scores for all combinations of defended methods and defense-aware attacks on optical flow methods on the Driving~\cite{Mayer2016LargeDatasetTrain} final (f) and clean (c) validation splits\cref{footn:valsplits}. Per attack, the robustness values of the best defense are \textbf{bold}. Per defense, the robustness values for the attack it is most vulnerable to are \underline{underlined}.
   Full pipelines are highlighted in gray, and provide the corresponding robustness values to the quality scores from \cref{tab:accur-Driving}.
   }
   \label{tab:rob-Driving}
   \small
   \centering
   \scalebox{.99}{
   \begin{tabular}{l@{\ \ \ }l@{\ \ \ }l@{\quad}|@{\quad}r@{\ \ \ }rr@{\ \ \ }rr@{\ \ \ }rr@{\ \ \ }rr@{\ \ \ }rr@{\ \ \ }rr@{\ \ \ }r}
      \toprule
      Attack type                    & \multicolumn{2}{l}{Defense}                        & \multicolumn{2}{c}{\rotatebox{0}{FNC}} & \multicolumn{2}{c}{\rotatebox{0}{FNCR}}  & \multicolumn{2}{c}{\rotatebox{0}{SpyNet}} & \multicolumn{2}{c}{\rotatebox{0}{PWC}} & \multicolumn{2}{c}{\rotatebox{0}{RAFT}} & \multicolumn{2}{c}{\rotatebox{0}{GMA}} & \multicolumn{2}{c}{\rotatebox{0}{FF}} \\
      \cmidrule(l{0.5em}r{0.5em}){4-5} \cmidrule(l{0.5em}r{0.5em}){6-7} \cmidrule(l{0.5em}r{0.5em}){8-9} \cmidrule(l{0.5em}r{0.5em}){10-11} \cmidrule(l{0.5em}r{0.5em}){12-13} \cmidrule(l{0.5em}r{0.5em}){14-15} \cmidrule(l{0.5em}r{0.5em}){16-17}
      & & \multicolumn{1}{c}{} & \multicolumn{1}{c}{f} & \multicolumn{1}{c}{c} & \multicolumn{1}{c}{f} & \multicolumn{1}{c}{c} & \multicolumn{1}{c}{f} & \multicolumn{1}{c}{c} & \multicolumn{1}{c}{f} & \multicolumn{1}{c}{c} & \multicolumn{1}{c}{f} & \multicolumn{1}{c}{c} & \multicolumn{1}{c}{f} & \multicolumn{1}{c}{c} & \multicolumn{1}{c}{f} & \multicolumn{1}{c}{c} \\
      \midrule
      \multirow{3}{*}{Vanilla}       & \cellcolor{gray!20}None & \cellcolor{gray!20}$R^\text{Van}$             &    \cellcolor{gray!20}\underline{93.15} & \cellcolor{gray!20}3.64 &            \cellcolor{gray!20}\underline{4.72} &  \cellcolor{gray!20}\underline{5.94} &              \cellcolor{gray!20}\textbf{\underline{4.39}} &  \cellcolor{gray!20}\textbf{\underline{4.22}} &              \cellcolor{gray!20}\underline{7.30} & \cellcolor{gray!20}\underline{7.77} &               \cellcolor{gray!20}\underline{4.74} &\cellcolor{gray!20}2.85 &                   \cellcolor{gray!20}\underline{4.69} & \cellcolor{gray!20}\underline{4.58} &                 \cellcolor{gray!20}\underline{4.10} &  \cellcolor{gray!20}\underline{4.04}  \\
                                     & LGS  & $R^\text{Van}_\text{LGS}$                                        &     \textbf{4.43} &  \textbf{3.25} &            3.39 &  \textbf{2.12} &              7.72 & 5.33 &              6.83 &\textbf{6.69} &               \underline{5.49} & 3.41 &                   5.18 & 3.40 &                 3.91 &  3.44  \\
                                     & ILP  & $R^\text{Van}_\text{ILP}$                                        &     4.88 &  \underline{6.05} &            \textbf{2.54} &  2.23 &              5.54 & 4.90 &              \textbf{5.73} &7.33 &               \textbf{3.31} & \textbf{2.83} &                   \textbf{3.43} & \textbf{3.19} &                 \textbf{\underline{3.26}} &  \textbf{3.38}  \\
      \cmidrule(l{0em}r{0em}){2-17}
      \multirow{3}{*}{+LGS (LGS-aware)}     & None & $R^\text{LGS}$                                            &    74.11 &  \textbf{4.11} &            \textbf{1.30} &  \textbf{1.64} &              \textbf{4.25} & \textbf{4.07} &              \textbf{6.63} & \textbf{7.00} &              4.19 &  \textbf{\underline{2.90}} &                  \textbf{3.14} &  \textbf{3.01} &                3.53 &   \textbf{3.71} \\
                                     & \cellcolor{gray!20}LGS  & \cellcolor{gray!20}$R^\text{LGS}_\text{LGS}$  &    \cellcolor{gray!20}\underline{24.93} &  \cellcolor{gray!20}\underline{8.61} &            \cellcolor{gray!20}\underline{3.55} &  \cellcolor{gray!20}\underline{3.05} &              \cellcolor{gray!20}\underline{7.90} &\cellcolor{gray!20}5.68 &              \cellcolor{gray!20}\underline{8.24} & \cellcolor{gray!20}\underline{8.02} &             \cellcolor{gray!20}5.44 &  \cellcolor{gray!20}\underline{3.72} &                  \cellcolor{gray!20}\underline{5.67} & \cellcolor{gray!20}3.91 &               \cellcolor{gray!20}4.05 &  \cellcolor{gray!20}4.18 \\
                                     & ILP  & $R^\text{LGS}_\text{ILP}$                                        &    \textbf{24.21} &  5.93 &            2.48 &  \underline{4.07} &              5.49 & 5.18 &              7.96 & 7.90 &              \textbf{\underline{3.98}} &  \underline{3.77} &                  3.51 &  3.87 &                \textbf{3.16} &   4.54 \\
      \cmidrule(l{0em}r{0em}){2-17}
      \multirow{3}{*}{+ILP (ILP-aware)}     & None & $R^\text{ILP}$                                            &    84.98 &  \textbf{\underline{5.34}} &            \textbf{2.70} &  3.49 &              \textbf{4.34} & \textbf{4.17} &              \textbf{6.82} & \textbf{7.44} &              \textbf{3.05} &  \textbf{2.56} &                  \textbf{3.15} &  \textbf{2.96} &                3.74 &   \textbf{3.93} \\
                                     & LGS  & $R^\text{ILP}_\text{LGS}$                                        &     \textbf{7.92} &  8.20 &            3.32 &  \textbf{2.38} &              7.86 & \underline{5.68} &              7.25 & 7.58 &              5.46 &  3.69 &                  5.52 &  \underline{4.04} &                \underline{4.16} &   \underline{4.42} \\
                                     & \cellcolor{gray!20}ILP  & \cellcolor{gray!20}$R^\text{ILP}_\text{ILP}$  &    \cellcolor{gray!20}\underline{74.86} & \cellcolor{gray!20}5.00 &            \cellcolor{gray!20}\underline{2.99} & \cellcolor{gray!20}3.80 &              \cellcolor{gray!20}\underline{5.55} & \cellcolor{gray!20}\underline{4.91} &              \cellcolor{gray!20}\underline{8.24} & \cellcolor{gray!20}\underline{8.30} &             \cellcolor{gray!20}3.28 & \cellcolor{gray!20}3.27 &                  \cellcolor{gray!20}\underline{3.53} &  \cellcolor{gray!20}\underline{3.93} &                \cellcolor{gray!20}\textbf{3.25} &   \cellcolor{gray!20}\underline{4.73} \\
      \bottomrule
   \end{tabular}
   }
\end{table*}

Focusing on the quality-robustness plots in \cref{fig:qual-rob-datasets}, we observe that defenses worsen quality and robustness for all optical flow methods (except those of FlowNetC) on HD1K and Spring, \cf \cref{sfig:HD1K} and \cref{sfig:Spring}.
On Sintel and Driving, the results are more differentiated:
For high-quality methods like RAFT, GMA and FlowFormer (red markers), defending them with ILP improves the robustness for the final versions of the datasets in \cref{sfig:Sintel-f} and \cref{sfig:Driving-f} -- on the clean dataset versions in \cref{sfig:Sintel-c} and \cref{sfig:Driving-c}, however, both defenses deteriorate either quality, or robustness, or both.
Defending the lower-quality methods SpyNet and PWCNet (blue markers) also deteriorates at least quality or robustness on both datasets, with the exception of PWCNet, where defending leads to minor robustness improvements on Sintel.
For FlowNetC and FlowNetCRobust (green markers), defenses do indeed improve the robustness on Sintel and Driving, but here it is LGS that leads to the best robustness scores.
Overall, this clearly supports that defenses should not be used in a \enquote{plug'n'play} manner without extensive application-specific testing, as they either do not improve the optical flow methods at all, or -- when they do improve the robustness -- their effect is small and does not apply to more than a few selected optical flow methods.
Hence, current detect-and-remove defenses cannot be recommended for general use.

To better understand the effectiveness differences of defenses on the tested datasets, we analyze the results in relation to the datasets in more detail.
When we consider the datasets KITTI, HD1K and Spring and their quality-robustness plots in Main Fig.\ 5, \cref{sfig:HD1K} and \cref{sfig:Spring}, we find that applying defenses to optical flow methods worsens quality \emph{and} robustness, which leads to a slanted line of markers per optical flow network.
This indicates that for these datasets, the defenses affect the unattacked defended flow $\flow_\text{D}$ as described in Sec.\ 6.4, Main paper, because worsening this flow enters into both, the quality calculation with $\text{EPE}(\flow^*, \flow_\text{D})$ and the robustness calculation with $\text{EPE}_{P}(\flow_\text{D}, \flow_\text{D}^\text{A})$.
For the datasets Sintel and Driving in \cref{sfig:Sintel-f}, \cref{sfig:Sintel-c}, \cref{sfig:Driving-f} and \cref{sfig:Driving-c}, applying defenses almost exclusively changes the robustness, leading to a horizontal line of markers per optical flow network.
This indicates that the defenses work \enquote{as intended}, affecting only the attacked defended flow $\flow_\text{D}^\text{A}$ and hence the robustness, but are on average not very effective under attack with defense-aware patches.
In summary, defenses have the worst side effects on the image quality for natural or naturalistic data: KITTI and HD1K contain camera-captured real-world images and Spring is a recently rendered dataset that focuses on high-detail images.
Even though defenses work partially on the synthetic datasets Sintel and Driving, which were rendered and created before 2016, they still fail to demonstrate consistent advantages over undefended networks on these datasets.
These differences in the datasets are also visible in terms of the dataset image statistics that are considered by the LGS and ILP defenses.
In \cref{fig:dataset-statistics} we show the histograms over first- and second-order image derivatives for all datasets.
There, the synthetic Sintel and Driving datasets have a very \enquote{even} gradient magnitude decay for large gradients on the log scale (both for clean and final rendering passes), while the realistic KITTI, HD1K and Spring datasets do not show such a clear exponential gradient decay.
All in all, defenses fail most severely on safety-critical real-world datasets, where reliable predictions are needed most.

\end{document}

%% file: graphics/patches-overview/patches-overview_None_ranking.tex
\begin{table}
    \centering
    \caption{
    Robustness $\text{EPE}(\flow,\flow^\text{Van})$~\cite{Schmalfuss2022} for undefended networks under vanilla patch attacks with different optimization parameter combinations. Non-evaluated settings are marked by \enquote{n.e.}.}
    \label{tab:patches-overview_None_ranking}
    \resizebox{\columnwidth}{!}{
        \begin{tabular}{@{}c@{\ \ }r@{\quad}c|c@{\ \ \ }c@{\ \ \ }c@{\ \ \ }c@{\ \ \ }c@{\ \ \ }c@{\ \ \ }c@{}}
            \toprule
            \rotatebox{60}{Optim.}   & \rotatebox{60}{LR}    & \multicolumn{1}{c}{\rotatebox{60}{Box}}   &    \rotatebox{60}{FNC} & \rotatebox{60}{FNCR} & \rotatebox{60}{PWC} & \rotatebox{60}{SpyNet} & \rotatebox{60}{RAFT} & \rotatebox{60}{GMA} & \rotatebox{60}{FF}    \\
            \midrule
            SGD   & 10.00  & CoV & 61.86          & 0.73          & 1.34          & 1.10           & 0.27          & 0.29          & 0.42          \\
            SGD   & 10.00  & Clip & 52.41          & 0.97          & 1.17          & 1.01           & 0.28          & 0.31          & 0.45          \\
            SGD   & 100.00 & CoV & \textbf{76.28} & 0.62          & 1.28          & 1.26           & 0.29          & 0.34          & n.e.       \\
            SGD   & 100.00 & Clip & 63.74          & 0.44          & 1.17          & 1.26           & 0.27          & 0.28          & 0.49          \\
            \midrule
            IFGSM & 0.01  & CoV & 58.56          & 1.28          & \textbf{1.84} & 1.30           & 0.29          & 0.61          & n.e.       \\
            IFGSM & 0.01  & Clip & 32.19          & \textbf{1.58} & 1.80          & 1.19           & 0.29          & \textbf{0.55} & \textbf{0.54} \\
            IFGSM & 0.10   & CoV & 57.55          & 1.47          & 1.84          & 1.33           & \textbf{0.34} & 0.46          & n.e.       \\
            IFGSM & 0.10   & Clip & 55.92          & 0.50          & 1.03          & 1.15           & 0.24          & 0.30          & 0.49          \\
            IFGSM & 1.00   & CoV & 60.62          & 1.23          & 1.60          & \textbf{1.33}  & 0.34          & 0.41          & n.e.       \\
            IFGSM & 1.00   & Clip & 8.22           & 0.45          & 0.88          & 1.11           & 0.26          & 0.32          & 0.44          \\
            \bottomrule
        \end{tabular}
    }
\end{table}

%% file: graphics/patches-overview/patches-overview_LGS_ranking.tex
\begin{table}
    \centering
    \caption{Robustness $\text{EPE}(\flow_\text{LGS},\flow^\text{LGS}_\text{LGS})$ for LGS-defended networks under LGS-aware patch attacks with different optimization parameter combinations. Non-evaluated settings are marked by \enquote{n.e.}, while diverging optimization runs are marked as \enquote{div}.}
    \label{tab:patches-overview_LGS_ranking}
    \resizebox{\columnwidth}{!}{
        \begin{tabular}{@{}c@{\ \ }r@{\quad}c|c@{\ \ \ }c@{\ \ \ }c@{\ \ \ }c@{\ \ \ }c@{\ \ \ }c@{\ \ \ }c@{}}
            \toprule
            \rotatebox{60}{Optim.}   & \rotatebox{60}{LR}    & \multicolumn{1}{c}{\rotatebox{60}{Box}}   &    \rotatebox{60}{FNC} & \rotatebox{60}{FNCR} & \rotatebox{60}{PWC} & \rotatebox{60}{SpyNet} & \rotatebox{60}{RAFT} & \rotatebox{60}{GMA} & \rotatebox{60}{FF}    \\
            \midrule
            SGD   & 10.00  & CoV & 3.98           & 3.07           & 3.28         & 3.62           & 1.45          & 1.57          & n.e.        \\
            SGD   & 10.00  & Clip & 3.02           & 2.98           & 3.03         & 3.50           & 1.31          & 1.49          & 1.59           \\
            SGD   & 100.00 & CoV & 3.44           & 3.17           & 3.29         & 3.74           & \textbf{1.47} & div.           & n.e.        \\
            SGD   & 100.00 & Clip & 3.27           & 3.17           & 3.15         & 3.59           & 1.33          & div.           & 1.56           \\
            \midrule
            IFGSM & 0.01  & CoV & \textbf{22.64} & 2.51           & 3.74         & 3.69           & 1.05          & 1.24          & n.e.        \\
            IFGSM & 0.01  & Clip & 19.03          & 2.71           & \textbf{3.90}& 3.67           & 1.17          & 1.34          & 1.31           \\
            IFGSM & 0.10  & CoV & 20.70          & 2.89           & 3.68         & \textbf{3.98}  & 1.33          & 1.49          & n.e.        \\
            IFGSM & 0.10  & Clip & 8.42           & 2.62           & 3.04         & 3.60           & 1.22          & 1.33          & 1.28           \\
            IFGSM & 1.00   & CoV & 4.61           & 3.10           & 3.28         & 3.62           & 1.42          & 1.53          & n.e.        \\
            IFGSM & 1.00   & Clip & 3.48           & \textbf{3.28}  & 3.37         & 3.86           & 1.45          & \textbf{1.62} & \textbf{1.71}  \\
            \bottomrule
        \end{tabular}
    }
\end{table}

%% file: graphics/patches-overview/patches-overview_ILP_ranking.tex
\begin{table}
    \centering
    \caption{Robustness $\text{EPE}(\flow_\text{ILP},\flow^\text{ILP}_\text{ILP})$ for ILP-defended networks under ILP-aware patch attacks with different optimization parameter combinations. Non-evaluated settings are marked by \enquote{n.e.}, while diverging optimization runs are marked as \enquote{div}.}
    \label{tab:patches-overview_ILP_ranking}
    \resizebox{\columnwidth}{!}{
        \begin{tabular}{@{}c@{\ \ }r@{\quad}c|c@{\ \ \ }c@{\ \ \ }c@{\ \ \ }c@{\ \ \ }c@{\ \ \ }c@{\ \ \ }c@{}}
            \toprule
            \rotatebox{60}{Optim.}   & \rotatebox{60}{LR}    & \multicolumn{1}{c}{\rotatebox{60}{Box}}   &    \rotatebox{60}{FNC} & \rotatebox{60}{FNCR} & \rotatebox{60}{PWC} & \rotatebox{60}{SpyNet} & \rotatebox{60}{RAFT} & \rotatebox{60}{GMA} & \rotatebox{60}{FF}    \\
            \midrule
            SGD   & 10.00  & CoV & 11.55          & 1.53           & 2.21         & 1.73           & 1.40          & 1.46          & n.e.        \\
            SGD   & 10.00  & Clip & 4.09           & 2.99           & 2.99         & 2.17           & 1.45          & 1.48          & 1.75           \\
            SGD   & 100.00 & CoV & 3.17           & 2.90           & 3.08         & 2.52           & 1.43          & div.           & n.e.        \\
            SGD   & 100.00 & Clip & 3.56           & 3.27           & 3.37         & 2.83           & 1.42          & 1.55          & 1.69           \\
            \midrule
            IFGSM & 0.01  & CoV & \textbf{57.46} & 2.95           & 3.84         & 2.77           & 1.12          & 1.25          & n.e.        \\
            IFGSM & 0.01  & Clip & 42.87          & 2.91           & \textbf{3.87}& 2.72           & 1.08          & 1.24          & 1.15           \\
            IFGSM & 0.10   & CoV & 54.74          & \textbf{3.30}  & 3.87         & \textbf{3.15}  & 1.36          & 1.46          & n.e.        \\
            IFGSM & 0.10   & Clip & 18.70          & 2.93           & 3.11         & 2.77           & 1.23          & 1.32          & 1.39           \\
            IFGSM & 1.00   & CoV & 3.84           & 3.22           & 3.34         & 2.98           & 1.37          & 1.43          & n.e.        \\
            IFGSM & 1.00   & Clip & 3.58           & 3.28           & 3.42         & 2.78           & \textbf{1.48} & \textbf{1.54} & \textbf{1.82}  \\
            \bottomrule
        \end{tabular}
    }
\end{table}

%% file: graphics/patches-overview/patches-overview_None.tex
\begin{figure*}
    \centering
    \resizebox{\textwidth}{!}{
        \begin{tabular}{crc@{\quad} c c c c c c c c c}
            \toprule\\[.1cm]
            Optim.                             & LR                              & Box                                 & FlowNetC                                                                                                              & FlowNetCRobust                                                                                                                & PWCNet                                                                                                               & SpyNet                                                                                                                 & RAFT                                                                                                              & GMA                                                                                                              & FlowFormer                                                                                                                \\[.2cm]
            \midrule\\[.1cm]
            \raisebox{.06\textwidth}{SGD}   & \raisebox{.06\textwidth}{10.00}  & \raisebox{.06\textwidth}{CoV} & \includegraphics[width=0.1\textwidth]{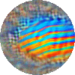}    & \includegraphics[width=0.1\textwidth]{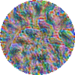}    & \includegraphics[width=0.1\textwidth]{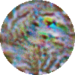}    & \includegraphics[width=0.1\textwidth]{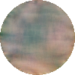}    & \includegraphics[width=0.1\textwidth]{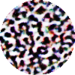}    & \includegraphics[width=0.1\textwidth]{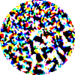}    & \includegraphics[width=0.1\textwidth]{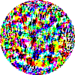}    & \\
            \raisebox{.06\textwidth}{SGD}   & \raisebox{.06\textwidth}{10.00}  & \raisebox{.06\textwidth}{Clip} & \includegraphics[width=0.1\textwidth]{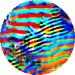}   & \includegraphics[width=0.1\textwidth]{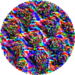}   & \includegraphics[width=0.1\textwidth]{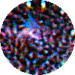}   & \includegraphics[width=0.1\textwidth]{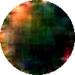}   & \includegraphics[width=0.1\textwidth]{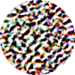}   & \includegraphics[width=0.1\textwidth]{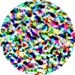}   & \includegraphics[width=0.1\textwidth]{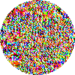}   & \\
            \raisebox{.06\textwidth}{SGD}   & \raisebox{.06\textwidth}{100.00} & \raisebox{.06\textwidth}{CoV} & \includegraphics[width=0.1\textwidth]{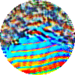}   & \includegraphics[width=0.1\textwidth]{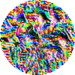}   & \includegraphics[width=0.1\textwidth]{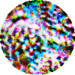}   & \includegraphics[width=0.1\textwidth]{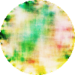}   & \includegraphics[width=0.1\textwidth]{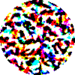}   & \includegraphics[width=0.1\textwidth]{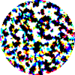}   & \raisebox{.05\textwidth}{n.e.}                                                                                        & \\
            \raisebox{.06\textwidth}{SGD}   & \raisebox{.06\textwidth}{100.00} & \raisebox{.06\textwidth}{Clip} & \includegraphics[width=0.1\textwidth]{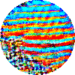}  & \includegraphics[width=0.1\textwidth]{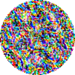}  & \includegraphics[width=0.1\textwidth]{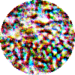}  & \includegraphics[width=0.1\textwidth]{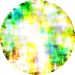}  & \includegraphics[width=0.1\textwidth]{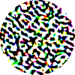}  & \includegraphics[width=0.1\textwidth]{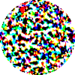}  & \includegraphics[width=0.1\textwidth]{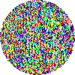}  & \\
            \raisebox{.06\textwidth}{IFGSM} & \raisebox{.06\textwidth}{0.01}  & \raisebox{.06\textwidth}{CoV} & \includegraphics[width=0.1\textwidth]{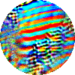}  & \includegraphics[width=0.1\textwidth]{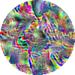}  & \includegraphics[width=0.1\textwidth]{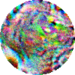}  & \includegraphics[width=0.1\textwidth]{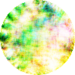}  & \includegraphics[width=0.1\textwidth]{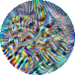}  & \includegraphics[width=0.1\textwidth]{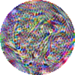}  & \raisebox{.05\textwidth}{n.e.}                                                                                        & \\
            \raisebox{.06\textwidth}{IFGSM} & \raisebox{.06\textwidth}{0.01}  & \raisebox{.06\textwidth}{Clip} & \includegraphics[width=0.1\textwidth]{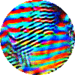} & \includegraphics[width=0.1\textwidth]{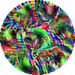} & \includegraphics[width=0.1\textwidth]{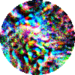} & \includegraphics[width=0.1\textwidth]{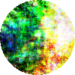} & \includegraphics[width=0.1\textwidth]{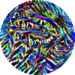} & \includegraphics[width=0.1\textwidth]{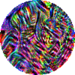} & \includegraphics[width=0.1\textwidth]{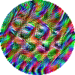} & \\
            \raisebox{.06\textwidth}{IFGSM} & \raisebox{.06\textwidth}{0.10}   & \raisebox{.06\textwidth}{CoV} & \includegraphics[width=0.1\textwidth]{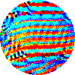}   & \includegraphics[width=0.1\textwidth]{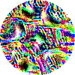}   & \includegraphics[width=0.1\textwidth]{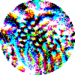}   & \includegraphics[width=0.1\textwidth]{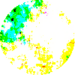}   & \includegraphics[width=0.1\textwidth]{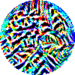}   & \includegraphics[width=0.1\textwidth]{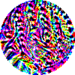}   & \raisebox{.05\textwidth}{n.e.}                                                                                        & \\
            \raisebox{.06\textwidth}{IFGSM} & \raisebox{.06\textwidth}{0.10}   & \raisebox{.06\textwidth}{Clip} & \includegraphics[width=0.1\textwidth]{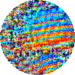}  & \includegraphics[width=0.1\textwidth]{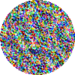}  & \includegraphics[width=0.1\textwidth]{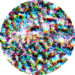}  & \includegraphics[width=0.1\textwidth]{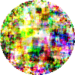}  & \includegraphics[width=0.1\textwidth]{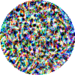}  & \includegraphics[width=0.1\textwidth]{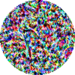}  & \includegraphics[width=0.1\textwidth]{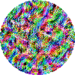}  & \\
            \raisebox{.06\textwidth}{IFGSM} & \raisebox{.06\textwidth}{1.00}   & \raisebox{.06\textwidth}{CoV} & \includegraphics[width=0.1\textwidth]{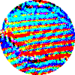}   & \includegraphics[width=0.1\textwidth]{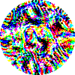}   & \includegraphics[width=0.1\textwidth]{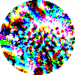}   & \includegraphics[width=0.1\textwidth]{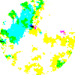}   & \includegraphics[width=0.1\textwidth]{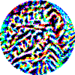}   & \includegraphics[width=0.1\textwidth]{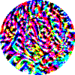}   & \raisebox{.05\textwidth}{n.e.}                                                                                        & \\
            \raisebox{.06\textwidth}{IFGSM} & \raisebox{.06\textwidth}{1.00}   & \raisebox{.06\textwidth}{Clip} & \includegraphics[width=0.1\textwidth]{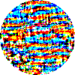}  & \includegraphics[width=0.1\textwidth]{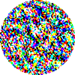}  & \includegraphics[width=0.1\textwidth]{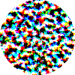}  & \includegraphics[width=0.1\textwidth]{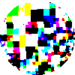}  & \includegraphics[width=0.1\textwidth]{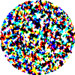}  & \includegraphics[width=0.1\textwidth]{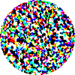}  & \includegraphics[width=0.1\textwidth]{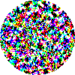}  & \\[0.5em]

            \bottomrule
        \end{tabular}
    }
    \caption{Best-performing vanilla patches for different networks and optimization parameter combinations. Non-evaluated settings are marked by \enquote{n.e.}. See \cref{tab:patches-overview_None_ranking} for the corresponding robustness values, averaged over four patches.}
    \label{tab:patches-overview_None}
\end{figure*}

%% file: graphics/patches-overview/patches-overview_LGS.tex
\begin{figure*}
    \centering
    \resizebox{\textwidth}{!}{
        \begin{tabular}{crc@{\quad} c c c c c c c c c}
            \toprule\\[.1cm]
            Optim.                             & LR                              & Box                                 & FlowNetC                                                                                                              & FlowNetCRobust                                                                                                              & PWCNet                                                                                                              & SpyNet                                                                                                              & RAFT                                                                                                              & GMA                                                                                                              & FlowFormer                                                                                                                \\[.2cm]
            \midrule\\[.1cm]
            \raisebox{.05\textwidth}{SGD}   & \raisebox{.05\textwidth}{10.00}  & \raisebox{.05\textwidth}{CoV} & \includegraphics[width=0.1\textwidth]{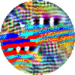}    & \includegraphics[width=0.1\textwidth]{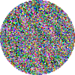}    & \includegraphics[width=0.1\textwidth]{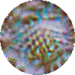}    & \includegraphics[width=0.1\textwidth]{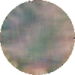}    & \includegraphics[width=0.1\textwidth]{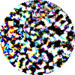}    & \includegraphics[width=0.1\textwidth]{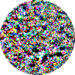}    & \raisebox{.05\textwidth}{n.e.}                                                                                       & \\
            \raisebox{.05\textwidth}{SGD}   & \raisebox{.05\textwidth}{10.00}  & \raisebox{.05\textwidth}{Clip} & \includegraphics[width=0.1\textwidth]{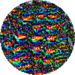}   & \includegraphics[width=0.1\textwidth]{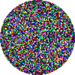}   & \includegraphics[width=0.1\textwidth]{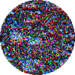}   & \includegraphics[width=0.1\textwidth]{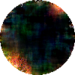}   & \includegraphics[width=0.1\textwidth]{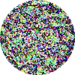}   & \includegraphics[width=0.1\textwidth]{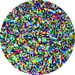}   & \includegraphics[width=0.1\textwidth]{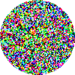}   & \\
            \raisebox{.05\textwidth}{SGD}   & \raisebox{.05\textwidth}{100.00} & \raisebox{.05\textwidth}{CoV} & \includegraphics[width=0.1\textwidth]{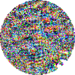}   & \includegraphics[width=0.1\textwidth]{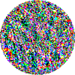}   & \includegraphics[width=0.1\textwidth]{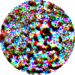}   & \includegraphics[width=0.1\textwidth]{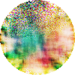}   & \includegraphics[width=0.1\textwidth]{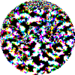}   & \raisebox{.05\textwidth}{div.}                                                                                    & \raisebox{.05\textwidth}{n.e.}                                                                                       & \\
            \raisebox{.05\textwidth}{SGD}   & \raisebox{.05\textwidth}{100.00} & \raisebox{.05\textwidth}{Clip} & \includegraphics[width=0.1\textwidth]{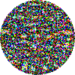}  & \includegraphics[width=0.1\textwidth]{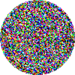}  & \includegraphics[width=0.1\textwidth]{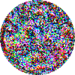}  & \includegraphics[width=0.1\textwidth]{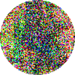}  & \includegraphics[width=0.1\textwidth]{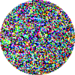}  & \raisebox{.05\textwidth}{div.}                                                                                    & \includegraphics[width=0.1\textwidth]{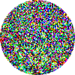}  & \\
            \raisebox{.05\textwidth}{IFGSM} & \raisebox{.05\textwidth}{0.01}  & \raisebox{.05\textwidth}{CoV} & \includegraphics[width=0.1\textwidth]{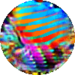}  & \includegraphics[width=0.1\textwidth]{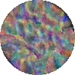}  & \includegraphics[width=0.1\textwidth]{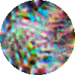}  & \includegraphics[width=0.1\textwidth]{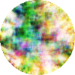}  & \includegraphics[width=0.1\textwidth]{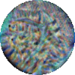}  & \includegraphics[width=0.1\textwidth]{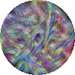}  & \raisebox{.05\textwidth}{n.e.}                                                                                       & \\
            \raisebox{.05\textwidth}{IFGSM} & \raisebox{.05\textwidth}{0.01}  & \raisebox{.05\textwidth}{Clip} & \includegraphics[width=0.1\textwidth]{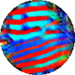} & \includegraphics[width=0.1\textwidth]{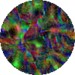} & \includegraphics[width=0.1\textwidth]{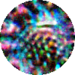} & \includegraphics[width=0.1\textwidth]{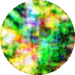} & \includegraphics[width=0.1\textwidth]{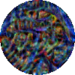} & \includegraphics[width=0.1\textwidth]{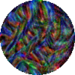} & \includegraphics[width=0.1\textwidth]{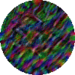} & \\
            \raisebox{.05\textwidth}{IFGSM} & \raisebox{.05\textwidth}{0.10}   & \raisebox{.05\textwidth}{CoV} & \includegraphics[width=0.1\textwidth]{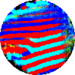}   & \includegraphics[width=0.1\textwidth]{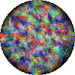}   & \includegraphics[width=0.1\textwidth]{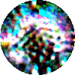}   & \includegraphics[width=0.1\textwidth]{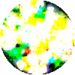}   & \includegraphics[width=0.1\textwidth]{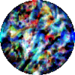}   & \includegraphics[width=0.1\textwidth]{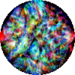}   & \raisebox{.05\textwidth}{n.e.}                                                                                       & \\
            \raisebox{.05\textwidth}{IFGSM} & \raisebox{.05\textwidth}{0.10}   & \raisebox{.05\textwidth}{Clip} & \includegraphics[width=0.1\textwidth]{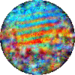}  & \includegraphics[width=0.1\textwidth]{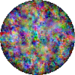}  & \includegraphics[width=0.1\textwidth]{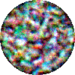}  & \includegraphics[width=0.1\textwidth]{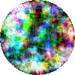}  & \includegraphics[width=0.1\textwidth]{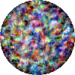}  & \includegraphics[width=0.1\textwidth]{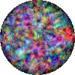}  & \includegraphics[width=0.1\textwidth]{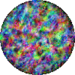}  & \\
            \raisebox{.05\textwidth}{IFGSM} & \raisebox{.05\textwidth}{1.00}   & \raisebox{.05\textwidth}{CoV} & \includegraphics[width=0.1\textwidth]{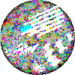}   & \includegraphics[width=0.1\textwidth]{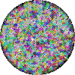}   & \includegraphics[width=0.1\textwidth]{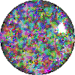}   & \includegraphics[width=0.1\textwidth]{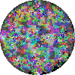}   & \includegraphics[width=0.1\textwidth]{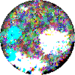}   & \includegraphics[width=0.1\textwidth]{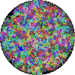}   & \raisebox{.05\textwidth}{n.e.}                                                                                       & \\
            \raisebox{.05\textwidth}{IFGSM} & \raisebox{.05\textwidth}{1.00}   & \raisebox{.05\textwidth}{Clip} & \includegraphics[width=0.1\textwidth]{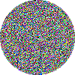}  & \includegraphics[width=0.1\textwidth]{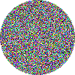}  & \includegraphics[width=0.1\textwidth]{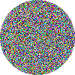}  & \includegraphics[width=0.1\textwidth]{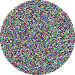}  & \includegraphics[width=0.1\textwidth]{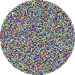}  & \includegraphics[width=0.1\textwidth]{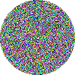}  & \includegraphics[width=0.1\textwidth]{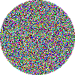}  & \\[0.5em]
            \bottomrule
        \end{tabular}
    }
    \caption{Best-performing LGS-aware patches for different networks and optimization parameter combinations. Non-evaluated settings are marked by \enquote{n.e.}, while diverging optimization runs are marked as \enquote{div}. See \cref{tab:patches-overview_LGS_ranking} for the corresponding robustness values, averaged over four patches.}
    \label{tab:patches-overview_LGS}
\end{figure*}

%% file: graphics/patches-overview/patches-overview_ILP.tex
\begin{figure*}
    \centering
    \resizebox{\textwidth}{!}{
        \begin{tabular}{crc@{\qquad} c c c c c c c c c}
            \toprule                                                                                                                                                                                                                                                                                                                                                                                                                                                                                                                                                                                                                                                                                                                                                                                                                                                                                                                                                                     \\[.1cm]
            Optim.                             & LR                              & Box                                 & FlowNetC                                                                                                              & FlowNetCRobust                                                                                                              & PWCNet                                                                                                             & SpyNet                                                                                                               & RAFT                                                                                                              & GMA                                                                                                              & FlowFormer                                                                                                                \\[.2cm]
            \midrule                                                                                                                                                                                                                                                                                                                                                                                                                                                                                                                                                                                                                                                                                                                                                                                                                                               \\[.1cm]
            \raisebox{.05\textwidth}{SGD}   & \raisebox{.05\textwidth}{10.00}  & \raisebox{.05\textwidth}{CoV} & \includegraphics[width=0.1\textwidth]{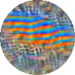}    & \includegraphics[width=0.1\textwidth]{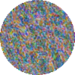}    & \includegraphics[width=0.1\textwidth]{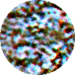}   & \includegraphics[width=0.1\textwidth]{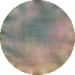}     & \includegraphics[width=0.1\textwidth]{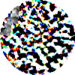}    & \includegraphics[width=0.1\textwidth]{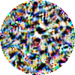}    & \raisebox{.05\textwidth}{n.e.}                                                                                       & \\
            \raisebox{.05\textwidth}{SGD}   & \raisebox{.05\textwidth}{10.00}  & \raisebox{.05\textwidth}{Clip} & \includegraphics[width=0.1\textwidth]{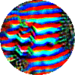}   & \includegraphics[width=0.1\textwidth]{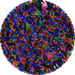}   & \includegraphics[width=0.1\textwidth]{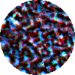}  & \includegraphics[width=0.1\textwidth]{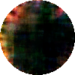}    & \includegraphics[width=0.1\textwidth]{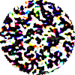}   & \includegraphics[width=0.1\textwidth]{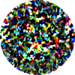}   & \includegraphics[width=0.1\textwidth]{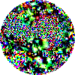}   & \\
            \raisebox{.05\textwidth}{SGD}   & \raisebox{.05\textwidth}{100.00} & \raisebox{.05\textwidth}{CoV} & \includegraphics[width=0.1\textwidth]{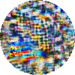}   & \includegraphics[width=0.1\textwidth]{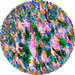}   & \includegraphics[width=0.1\textwidth]{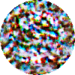}  & \includegraphics[width=0.1\textwidth]{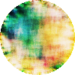}    & \includegraphics[width=0.1\textwidth]{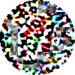}   & \raisebox{.05\textwidth}{div.}                                                                                    & \raisebox{.05\textwidth}{n.e.}                                                                                       & \\
            \raisebox{.05\textwidth}{SGD}   & \raisebox{.05\textwidth}{100.00} & \raisebox{.05\textwidth}{Clip} & \includegraphics[width=0.1\textwidth]{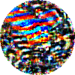}  & \includegraphics[width=0.1\textwidth]{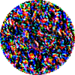}  & \includegraphics[width=0.1\textwidth]{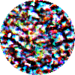} & \includegraphics[width=0.1\textwidth]{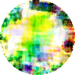}   & \includegraphics[width=0.1\textwidth]{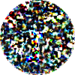}  & \includegraphics[width=0.1\textwidth]{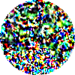}  & \includegraphics[width=0.1\textwidth]{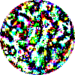}  & \\
            \raisebox{.05\textwidth}{IFGSM} & \raisebox{.05\textwidth}{0.01}  & \raisebox{.05\textwidth}{CoV} & \includegraphics[width=0.1\textwidth]{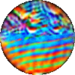}  & \includegraphics[width=0.1\textwidth]{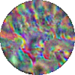}  & \includegraphics[width=0.1\textwidth]{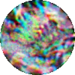} & \includegraphics[width=0.1\textwidth]{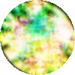}   & \includegraphics[width=0.1\textwidth]{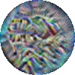}  & \includegraphics[width=0.1\textwidth]{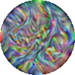}  & \raisebox{.05\textwidth}{n.e.}                                                                                       & \\
            \raisebox{.05\textwidth}{IFGSM} & \raisebox{.05\textwidth}{0.01}  & \raisebox{.05\textwidth}{Clip} & \includegraphics[width=0.1\textwidth]{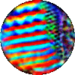} & \includegraphics[width=0.1\textwidth]{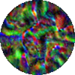} & \includegraphics[width=0.1\textwidth]{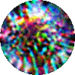}& \includegraphics[width=0.1\textwidth]{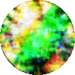}  & \includegraphics[width=0.1\textwidth]{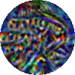} & \includegraphics[width=0.1\textwidth]{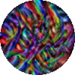} & \includegraphics[width=0.1\textwidth]{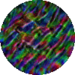} & \\
            \raisebox{.05\textwidth}{IFGSM} & \raisebox{.05\textwidth}{0.10}   & \raisebox{.05\textwidth}{CoV} & \includegraphics[width=0.1\textwidth]{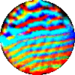}   & \includegraphics[width=0.1\textwidth]{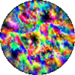}   & \includegraphics[width=0.1\textwidth]{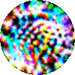}  & \includegraphics[width=0.1\textwidth]{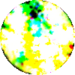}    & \includegraphics[width=0.1\textwidth]{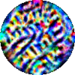}   & \includegraphics[width=0.1\textwidth]{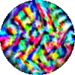}   & \raisebox{.05\textwidth}{n.e.}                                                                                       & \\
            \raisebox{.05\textwidth}{IFGSM} & \raisebox{.05\textwidth}{0.10}   & \raisebox{.05\textwidth}{Clip} & \includegraphics[width=0.1\textwidth]{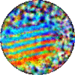}  & \includegraphics[width=0.1\textwidth]{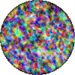}  & \includegraphics[width=0.1\textwidth]{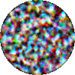} & \includegraphics[width=0.1\textwidth]{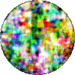}   & \includegraphics[width=0.1\textwidth]{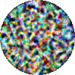}  & \includegraphics[width=0.1\textwidth]{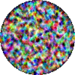}  & \includegraphics[width=0.1\textwidth]{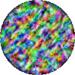}  & \\
            \raisebox{.05\textwidth}{IFGSM} & \raisebox{.05\textwidth}{1.00}   & \raisebox{.05\textwidth}{CoV} & \includegraphics[width=0.1\textwidth]{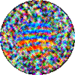}   & \includegraphics[width=0.1\textwidth]{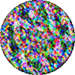}   & \includegraphics[width=0.1\textwidth]{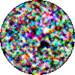}  & \includegraphics[width=0.1\textwidth]{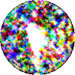}    & \includegraphics[width=0.1\textwidth]{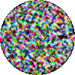}   & \includegraphics[width=0.1\textwidth]{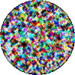}   & \raisebox{.05\textwidth}{n.e.}                                                                                       & \\
            \raisebox{.05\textwidth}{IFGSM} & \raisebox{.05\textwidth}{1.00}   & \raisebox{.05\textwidth}{Clip} & \includegraphics[width=0.1\textwidth]{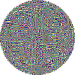}  & \includegraphics[width=0.1\textwidth]{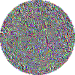}  & \includegraphics[width=0.1\textwidth]{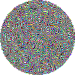} & \includegraphics[width=0.1\textwidth]{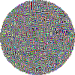}   & \includegraphics[width=0.1\textwidth]{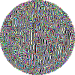}  & \includegraphics[width=0.1\textwidth]{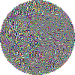}  & \includegraphics[width=0.1\textwidth]{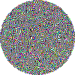}  & \\[0.5em]
            \bottomrule
        \end{tabular}
    }
    \caption{Best-performing vanilla patches for different networks and optimization parameter combinations. Non-evaluated settings are marked by \enquote{n.e.}, while diverging optimization runs are marked as \enquote{div}. See \cref{tab:patches-overview_ILP_ranking} for the corresponding robustness values, averaged over four patches.}
    \label{tab:patches-overview_ILP}
\end{figure*}

%% file: graphics/vanilla_effects/effects.tex
\begin{figure}
    \setlength{\tabcolsep}{0.3pt}
    \setlength{\fboxsep}{0pt}
    \setlength{\fboxrule}{.1pt}
    \centering
    \resizebox{\linewidth}{!}{
        \begin{tabular}{c c c}
            \leftrotboxwithfboxupper{height=.11\linewidth}{graphics/vanilla_effects/0_FlowNetC_F_None_None.png}{FNC}{$f$ unattacked}
            &
            \smallcirclebox{\leftrotboxwithfboxupperone{height=.11\linewidth}{graphics/vanilla_effects/0_FlowNetC_F_Std_None.png}{$f^\text{Van}_\text{None}$ vanilla attacked}}{.5}{.46}{.24}
            &
            \leftrotboxwithfboxupperone{height=.11\linewidth}{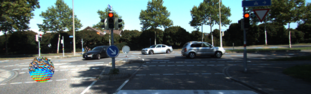}{$I^\text{Van}_\text{None}$ vanilla attacked}
            \\[-.3em]
            \leftrotboxwithfbox{height=.11\linewidth}{graphics/vanilla_effects/0_FlowNetCRobust_F_None_None.png}{FNCR}
            &
            \smallcirclebox{\fbox{\includegraphics[height=.11\linewidth]{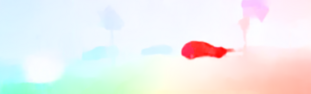}}}{.5}{.46}{.24}
            &
            \fbox{\includegraphics[height=.11\linewidth]{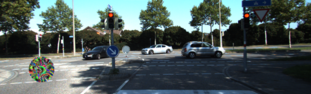}}
            \\[-.3em]
            \leftrotboxwithfbox{height=.11\linewidth}{graphics/vanilla_effects/0_PWCNet_F_None_None.png}{PWC}
            &
            \smallcirclebox{\fbox{\includegraphics[height=.11\linewidth]{graphics/vanilla_effects/0_PWCNet_F_Std_None.png}}}{.5}{.46}{.24}
            &
            \fbox{\includegraphics[height=.11\linewidth]{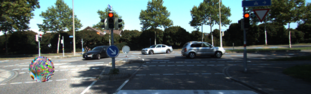}}
            \\[-.3em]
            \leftrotboxwithfbox{height=.11\linewidth}{graphics/vanilla_effects/0_SpyNet_F_None_None.png}{SpyNet}
            &
            \smallcirclebox{\fbox{\includegraphics[height=.11\linewidth]{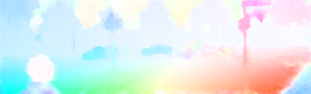}}}{.5}{.46}{.24}
            &
            \fbox{\includegraphics[height=.11\linewidth]{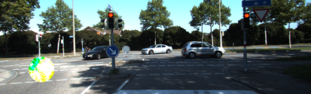}}
            \\[-.3em]
            \leftrotboxwithfbox{height=.11\linewidth}{graphics/vanilla_effects/0_RAFT_F_None_None.png}{RAFT}
            &
            \smallcirclebox{\fbox{\includegraphics[height=.11\linewidth]{graphics/vanilla_effects/0_RAFT_F_Std_None.png}}}{.5}{.46}{.24}
            &
            \fbox{\includegraphics[height=.11\linewidth]{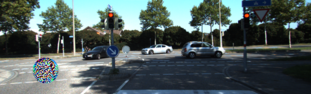}}
            \\[-.3em]
            \leftrotboxwithfbox{height=.11\linewidth}{graphics/vanilla_effects/0_GMA_F_None_None.png}{GMA}
            &
            \smallcirclebox{\fbox{\includegraphics[height=.11\linewidth]{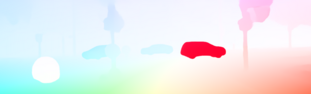}}}{.5}{.46}{.24}
            &
            \fbox{\includegraphics[height=.11\linewidth]{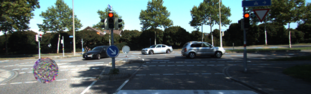}}
            \\[-.3em]
            \leftrotboxwithfbox{height=.11\linewidth}{graphics/vanilla_effects/0_FlowFormer_F_None_None.png}{FF}
            &
            \smallcirclebox{\fbox{\includegraphics[height=.11\linewidth]{graphics/vanilla_effects/0_FlowFormer_F_Std_None.png}}}{.5}{.46}{.24}
            &
            \fbox{\includegraphics[height=.11\linewidth]{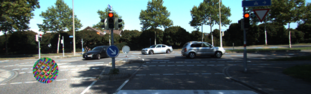}}
            \\
        \end{tabular}
    }
    \caption{Unattacked optical flow estimation (left) and corresponding vanilla-attacked optical flow (middle) for all tested methods on a KITTI sample (right). Complements Main \cref{fig:vanilla_attack}, see \cref{supp:fig:vanilla-rob2} for more samples.}
    \label{supp:fig:vanilla-rob1}
\end{figure}

\begin{figure}
    \setlength{\tabcolsep}{0.3pt}
    \setlength{\fboxsep}{0pt}
    \setlength{\fboxrule}{.1pt}
    \centering
    \resizebox{\linewidth}{!}{
        \begin{tabular}{ccc}
            \leftrotboxwithfboxupper{height=.11\linewidth}{graphics/vanilla_effects/1_FlowNetC_F_None_None.png}{FNC}{$f$ unattacked}
            &
            \smallcirclebox{\leftrotboxwithfboxupperone{height=.11\linewidth}{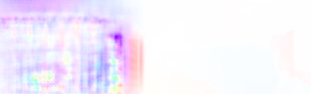}{$f^\text{Van}_\text{None}$ vanilla attacked}}{.5}{.46}{.24}
            &
            \leftrotboxwithfboxupperone{height=.11\linewidth}{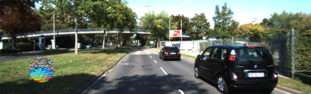}{$I^\text{Van}_\text{None}$ vanilla attacked}
            \\[-.3em]
            \leftrotboxwithfbox{height=.11\linewidth}{graphics/vanilla_effects/1_FlowNetCRobust_F_None_None.png}{FNCR}
            &
            \smallcirclebox{\fbox{\includegraphics[height=.11\linewidth]{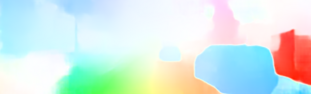}}}{.5}{.46}{.24}
            &
            \fbox{\includegraphics[height=.11\linewidth]{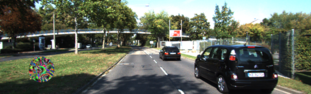}}
            \\[-.3em]
            \leftrotboxwithfbox{height=.11\linewidth}{graphics/vanilla_effects/1_PWCNet_F_None_None.png}{PWC}
            &
            \smallcirclebox{\fbox{\includegraphics[height=.11\linewidth]{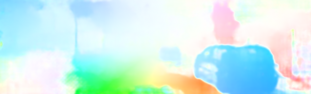}}}{.5}{.46}{.24}
            &
            \fbox{\includegraphics[height=.11\linewidth]{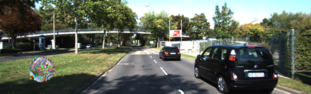}}
            \\[-.3em]
            \leftrotboxwithfbox{height=.11\linewidth}{graphics/vanilla_effects/1_SpyNet_F_None_None.png}{SpyNet}
            &
            \smallcirclebox{\fbox{\includegraphics[height=.11\linewidth]{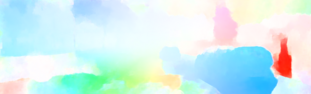}}}{.5}{.46}{.24}
            &
            \fbox{\includegraphics[height=.11\linewidth]{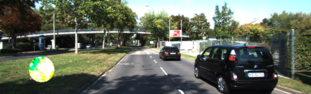}}
            \\[-.3em]
            \leftrotboxwithfbox{height=.11\linewidth}{graphics/vanilla_effects/1_RAFT_F_None_None.png}{RAFT}
            &
            \smallcirclebox{\fbox{\includegraphics[height=.11\linewidth]{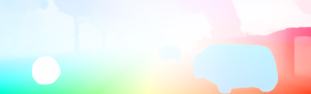}}}{.5}{.46}{.24}
            &
            \fbox{\includegraphics[height=.11\linewidth]{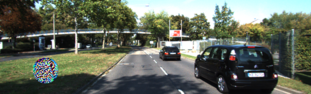}}
            \\[-.3em]
            \leftrotboxwithfbox{height=.11\linewidth}{graphics/vanilla_effects/1_GMA_F_None_None.png}{GMA}
            &
            \smallcirclebox{\fbox{\includegraphics[height=.11\linewidth]{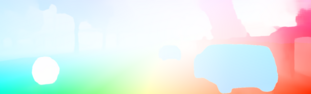}}}{.5}{.46}{.24}
            &
            \fbox{\includegraphics[height=.11\linewidth]{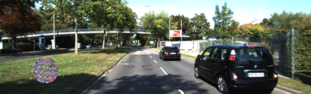}}
            \\[-.3em]
            \leftrotboxwithfbox{height=.11\linewidth}{graphics/vanilla_effects/1_FlowFormer_F_None_None.png}{FF}
            &
            \smallcirclebox{\fbox{\includegraphics[height=.11\linewidth]{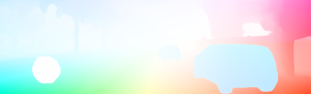}}}{.5}{.46}{.24}
            &
            \fbox{\includegraphics[height=.11\linewidth]{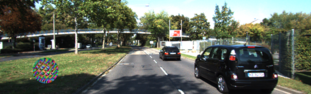}}
            \\
        \end{tabular}
    }
    \caption{Unattacked optical flow estimation (left) and corresponding vanilla-attacked optical flow (middle) for all tested methods on a KITTI sample (right). See \cref{supp:fig:vanilla-rob1} for more samples.}
    \label{supp:fig:vanilla-rob2}
\end{figure}